\documentclass{article}
\usepackage[a4paper, total={6in,9in}]{geometry}
\usepackage[utf8]{inputenc}
\usepackage[T1]{fontenc}
\usepackage{lmodern}
\usepackage{comment}

\usepackage[pdftex]{graphicx} 
\usepackage{subcaption}
\graphicspath{ {./images/} }
\usepackage{enumitem}
\usepackage[normalem]{ulem}
\usepackage[toc,page]{appendix}

\usepackage[bookmarks, bookmarksnumbered,
			colorlinks=true,
            linkcolor = blue,
            urlcolor  = blue,
            citecolor = teal]{hyperref}
\newcommand\fnsurl[1]{{\footnotesize\url{#1}}}

\newif\ifshownotes
\shownotestrue

\newif\ifarxiv
\arxivtrue

\newif\ifaistats
\aistatsfalse

\usepackage{myquicksetup}
\mathtoolsset{showonlyrefs} 

\usepackage{mylivemacros}

\usepackage{subfiles} 

\usepackage{mylivemacros}

\newcommand\blfootnote[1]{%
  \begingroup
  \renewcommand\thefootnote{}\footnote{#1}%
  \addtocounter{footnote}{-1}%
  \endgroup
}

\usepackage{authblk}
\newcommand\CoAuthorMark{\footnotemark[\arabic{footnote}]}
\author[1]{Stefan Stojanovic$^*$}
\author[1,2]{Konstantin Donhauser$^*$\protect\CoAuthorMark}
\author[1]{Fanny Yang}

\affil[1]{Department of Computer Science, ETH Z\"urich}
\affil[2]{ETH AI Center}

\title{Tight bounds for maximum $\ell_1$-margin classifiers}
\date{}

\hypersetup{
    pdftitle={Tight bounds for maximum l1-margin classifiers},
}

\usepackage{thm-restate}

\begin{document}

\maketitle

\vspace{-0.3in}
\blfootnote{*Equal contribution}
\begin{abstract}
Popular iterative algorithms such as boosting methods and coordinate descent on linear models converge to the maximum $\ell_1$-margin classifier, a.k.a. sparse hard-margin SVM, in high dimensional regimes where the data is linearly separable. Previous works consistently show that many estimators relying on the $\ell_1$-norm achieve improved statistical rates for hard sparse ground truths. We show that surprisingly, this adaptivity does not apply to the maximum $\ell_1$-margin classifier for a standard discriminative setting. In particular, for the noiseless setting, we prove tight upper and lower bounds for the prediction error that match existing rates of order $\frac{\|\wgt\|_1^{2/3}}{n^{1/3}}$  for general ground truths. To complete the picture, we show that when interpolating noisy observations, the error vanishes at a rate of order $\frac{1}{\sqrt{\log(d/n)}}$. We are therefore first to show benign overfitting for the maximum $\ell_1$-margin classifier. 




\end{abstract}


\ifaistats
\section{INTRODUCTION} 
\else
\section{Introduction} 
\fi
\label{sec:intro}

The ability to generalize in high-dimensional learning tasks crucially hinges on 
structural assumptions on the underlying ground truth. Probably the most commonly studied 
assumption is that the observations only depend on few input features, also called sparsity of the ground truth. Popular iterative algorithms widely used in practice to train models in such settings include coordinate descent (see \cite{wright2015coordinate} for a survey) and
boosting methods (e.g., Adaboost \cite{freund1997decision}). Numerous influential works \citesmart{bartlett98,rudin2004dynamics,zhang2005boosting,shalev2010equivalence,schapire2013boosting,telgarsky_13,gunasekar2018characterizing} make an important step towards mathematically
understanding these algorithms by showing that these solutions have the implicit bias of converging to the maximum $\ell_1$-margin
classifier (sparse hard-margin SVM).

So far, however, there exists relatively little analysis on  the generalization capabilities of the maximum $\ell_1$-margin classifier; existing non-asymptotic results only considers general (non-sparse) ground truths and adversarial corruptions \citesmart{chinot_2021}, while asymptotic
results consider regimes where the prediction error does not vanish \citesmart{liang2022precise}.  In this  paper, 
we derive tight matching upper and lower bounds for the
prediction error in a high-dimensional discriminative classification setting with (hard) sparse ground truths. 
Our theory holds for Gaussian covariate distributions
and the tightness of our bounds crucially rely on Gaussian comparison results \citesmart{gordon_1988,thrampoulidis_2015} (for comparison
with previous work see  Section~\ref{subsec:hyperplane}).
Our tight non-asymptotic bounds allow us to answer two open problems regarding maximum $\ell_1$-margin classifier related to its adaptivity to sparsity (Problem 1) and benign overfitting (Problem 2).

\paragraph{Problem 1: adaptivity to sparsity} Intuitively, linear estimators relying on the $\ell_1$-norm should \emph{adapt} to (hard) sparse ground truths by achieving faster rates than for ground truths where only the $\ell_1$-norm is bounded.
For instance, this gap has been proven for  $\ell_1$-norm penalized maximum average margin classifiers  \citesmart{zhang2014efficient}, as well as basis pursuit (which achieves exact recovery only under sparsity assumptions \citesmart{donoho2006compressed,candes2006near}) and the LASSO \citesmart{tibshirani_1996,vandegeer_2008} in linear regression  settings. 

However, so far there are no results in the literature that show adaptivity to sparsity of the (interpolating) maximum $\ell_1$-margin classifier in high-dimensional discriminative learning tasks. In fact, recent work \citesmart{chinot_2021} posed 
the following open problem:
\begin{center}
        \emph{ (Q1): Is the maximum $\ell_1$-margin classifier adaptive to sparsity for noiseless data?}
\end{center}

In Section~\ref{subsec:noiseless} we show that surprisingly, the answer is negative:
The tight rate $\frac{\|\wgt\|_1^{2/3}}{n^{1/3}}$ for (hard-) sparse normalized ground truths $\wgt$ in Theorem~\ref{thm:mainl1_class_noiseless} 
is of the same order as the upper bounds in \citesmart{chinot_2021} 
that hold for general ground truths.

\paragraph{Problem 2: benign overfitting}  Motivated by empirical observations for largely overparameterized models \citesmart{zhang_2021,belkin_2019}, a line of research recently emerged showing ``benign overfitting'' \citesmart{bartlett2020benign} for linear interpolating classifiers. More specifically, these papers show that the prediction error yields vanishing rates although the model interpolates noisy observations, where a constant fraction of labels are randomly corrupted \citesmart{muthukumar2021classification,donhauser2022fast,shamir22}.

So far, however, no such results exist for the  maximum $\ell_1$-margin classifier.
Existing upper bounds in \citesmart{chinot_2021} are tight for arbitrary (adversarial) corruptions but require the fraction of corrupted labels to go to zero to reach vanishing rates.   
It is unclear whether these rates can be improved for random (non-adversarial) corruptions:
\begin{center}
    \emph{(Q2): Does the prediction error for the maximum $\ell_1$-margin classifier yield vanishing rates when a constant fraction of the labels are randomly corrupted?}
\end{center}
In Section~\ref{subsec:main_results_noisy}, we show that this is indeed true: The maximum $\ell_1$-margin classifier achieves a logarithmic rate of order  $\frac{1}{\sqrt{\log(d/n)}}$ in Theorem~\ref{thm:mainl1_class_noisy}
--- which is much slower than for the noiseless case and far from being minimax optimal \citesmart{Wainwright2009InformationTheoreticLO, abramovich2018high}, but nonetheless vanishing in high-dimensional regimes when $d > n^{1+\epsilon}$.
We therefore complement the literature on benign overfitting for maximum $\ell_p$-margin classifiers with $p>1$, which can even achieve much faster polynomial rates \citesmart{donhauser2022fast}. 
\ifaistats
    \section{SETTING}
\else
    \ifarxiv
    \section{Setting}
    \else
    \section{Main Results}
    \label{sec:main_result}
    We begin this section by introducing the setting which we use in this thesis. Afterwards, in Section~\ref{subsec:noiseless}, we state our main result for the noiseless setting, Theorem~\ref{thm:mainl1_class_noiseless}, and in Section~\ref{subsec:noisy}, we present our main result for the noisy setting, Theorem~\ref{thm:mainl1_class_noisy}. Finally, we present a discussion comparing our main results with existing results based on hyperplane tessellation in Section~\ref{subsec:hyperplane}.
    \subsection{Setting}
    \fi
\fi
\label{sec:setting}

In this section we introduce the data model, prediction error and maximum $\ell_1$-margin classifier. We study a standard discriminative data model which is commonly studied  in the  1-bit compressed sensing literature (see e.g., \cite{boufounos2008,plan2012robust} and references therein). 

We assume that we observe $n$ pairs of i.i.d.~input features $x_i \overset{\iid}{\sim} \NNN(0,I_d)$ and associated labels 
$y_i =\sgn(\langle \xui, \wgt \rangle) \xi_i$ where $\wgt$ is the (normalized) ground truth (i.e.,  ${\norm{\wgt}_2 =1})$. Unlike previous works \citesmart{chinot_2021}, our proofs crucially rely on the Gaussianity of the input features (see Section~\ref{subsec:hyperplane} for a comparison with existing proof techniques). 
We say that the label $y_i$ is clean if  $\xi_i = 1$ and corrupted if $\xi_i = -1$.  We study the two cases where either all labels are clean (noiseless), i.e. $\forall i: ~\xi_i =1$, or where the corruptions $\xi_i \in \{-1,1\}$ are randomly drawn  from a distribution $\probsigma$ (noisy) only depending on the features in the direction of the ground truth:
\begin{equation}
\label{eq:labelscor}
  \xi_i | x_i \overset{\text{i.i.d.}}{\sim} \probsigma( \cdot; \langle x_i, \wgt \rangle).
\end{equation}
 As proposed in \citesmart{donhauser2022fast}, we make the following technical assumption on the noise distribution~$\probsigma$:

\begin{ass}[Noise model]
\label{ass:noise}
 The function ${z \mapsto \probsigma(\xi=1; z)}$ is a piece-wise continuous function such that the minimum ${\nubar := \underset{\nu}{\arg\min}~ \EE_{Z \sim \NNN(0,1)} \EE_{\xi \sim \probsigma(\cdot; Z)} \left( 1- \xi \nu \vert Z \vert\right)_+^2 }$ exists and is positive $\nubar>0$.  
\end{ass}

 This assumption is rather weak and satisfied by most noise models in the literature,  such as
\begin{itemize}
    \item\textit{Logistic regression} with  $\probsigma(\xi_i=1; z) = h(z \sigma)$ and $h(z) = \frac{e^{|z|}}{1+e^{|z|}}$ and $\sigma >0$. 
    \item \textit{Random label flips} with  $\probsigma(\xi=1; \langle \xui, \wstar \rangle) = 1-\sigma$ and  $\sigma \in (0,\frac{1}{2})$. 
    \item \textit{Random noise before quantization} where $\yui = \sgn(\langle \wgt, \xui \rangle + \tilde{\xi}_i)$ with $\tilde{\xi}_i\vert x_i \sim \NNN(0, \sigma^2)$ and  $\sigma^2>0$.
\end{itemize}

Given the data set $\{\left(\xui, \yui\right)\}_{i=1}^n$, the goal is to obtain an estimate $\what$ that
directionally aligns with the  normalized ground truth $\wgt$ and thus has a small prediction error:
\ifaistats
\begin{align}
  \RiskC(\what) &:=  \EE_{x\sim \NNN(0,I_d)} \idvec[\sgn(\langle x, \what\rangle) \neq \sgn(\langle x,\wgt \rangle)] \nonumber\\
    &= \frac{1}{\pi} \arccos\left(\left\langle \frac{\what}{\norm{\what}_2}, \wgt\right\rangle\right) \, \label{eq:prediction_error_cla}.
\end{align}
\else
\begin{align}
  \RiskC(\what) &:=  \EE_{x\sim \NNN(0,I_d)} \idvec[\sgn(\langle x, \what\rangle) \neq \sgn(\langle x,\wgt \rangle)] = \frac{1}{\pi} \arccos\left(\left\langle \frac{\what}{\norm{\what}_2}, \wgt\right\rangle\right) \, \label{eq:prediction_error_cla}.
\end{align}
\fi
By  the Taylor series approximation, one can directly see that a small prediction error corresponds to a small directional estimation error, which is  commonly studied in the $1$-bit compressed sensing literature \citesmart{boufounos2008} since
\begin{equation}
   \RiskC(\what) \approx \frac{1}{\pi}\norm{\frac{\what}{\norm{\what}_2} - \wgt}_2.
   \label{eq:risk_directional_estimation_error}
\end{equation}
We study the \emph{maximum $\ell_1$-margin interpolators}, or equivalently, the \emph{ sparse hard-margin SVM} solution defined by 
    \begin{equation}\label{eq:svm problem}
    \what = \argmin_{w} \norm{w}_1 \subjto \forall i:~ \yui \langle \xui, w \rangle \geq 1.
\end{equation}

\begin{remark}
While our two main results in Section~\ref{sec:main_result},  Theorem~\ref{thm:mainl1_class_noiseless} and \ref{thm:mainl1_class_noisy}, are stated for the maximum $\ell_1$-margin classifier, the bounds in the theorems hold uniformly for all interpolating classifiers with large (close to the optimal) $\ell_1$-margin (see Proposition~\ref{prop:BoundGamma0_class_noiseless} and \ref{prop:BoundGamma0_class_noisy}) 
\end{remark}


\ifaistats
    \section{MAIN RESULTS}
\else
    \section{Main Results}
\fi
    \label{sec:main_result}
    In this section we state our main result for the noiseless (Theorem~\ref{thm:mainl1_class_noiseless} in Section~\ref{subsec:noiseless})  and noisy setting (Theorem~\ref{thm:mainl1_class_noisy} in  Section~\ref{subsec:noisy}). For both results we assume that the data is distributed as described in Section~\ref{sec:setting}. 
    Furthermore, we present a discussion comparing our main results with existing results based on hyperplane tessellation in Section~\ref{subsec:hyperplane}.

\subsection{Main result for noiseless observations}
\label{subsec:noiseless}
Our first main result stated in the following theorem  provides tight upper and lower bounds in the noiseless setting:




\begin{theorem}[Noiseless classification] \label{thm:mainl1_class_noiseless}
    Assume that $\forall i, $ $\xi_i =1$ and $\wgt$ is a $s$-sparse vector with $s\leq  n^{\frac{2(1-7p)}{3}} $ and $p\in(0,\frac{1}{12})$. There exist universal constants $\kappa_1,\kappa_2,\kappa_3,c_1,c_2,c_3>0$ such that
    for any $n\geq \kappa_1$ and
    $\kappa_2 \sopt \leq d \leq \exp(\kappa_3 n^{p})$, the prediction error 
    is upper- and lower-bounded by
    \begin{equation}
    \label{eq:mainboundl1noiseless}
         \left\vert \RiskC(\what)  -  \left(\frac{ \kappa_0 \norm{\wgt}_1^{2} }{n \log^{1/2}(d/\sopt)} \right)^{1/3}\right\vert \lesssim \left(\frac{ \norm{\wgt}_1^{2} }{n\log(d/\sopt)}\right)^{1/3} \, ,
    \end{equation}
    with probability at least 
    $1- c_1 d^{-1} - c_2 \exp\left(-c_3 \frac{n^{1/3}}{\log^4(d/\sopt )} \right)$ over the draws of the data set
    where we define $\kappa_0 =\frac{8}{ \sqrt{3}\pi^{5/2}} $ and ${\sopt \asymp  (n \norm{\wgt}_1)^{2/3} \log^{1/3}(d/(n\norm{\wgt}_1)^{2/3})}$ (the exact expression is given in Equation \eqref{eq:definition_sopt} in Section~\ref{subsec:proof_sketch_noiseless_main}).
\end{theorem}
 The proof of the theorem is deferred to
 Appendix~\ref{sec:proof_noiseless_appendix} and an overview is given in Section~\ref{sec:proofsketch}. We now discuss the implications of the theorem in the following paragraphs.

\paragraph{Adaptivity to sparsity}

Existing upper bounds \citesmart{chinot_2021} for the maximum $\ell_1$-margin classifier  hold for any normalized ground truth $\wstar$ (with $\|\wstar\|_2 =1$) and
are of order 
 $\RiskC(\what) = O \left( \frac{\|\wgt\|_1^2}{n}\right)^{1/3}$ up to logarithmic factors. 
Our matching upper and lower bounds in  Theorem~\ref{thm:mainl1_class_noiseless} show that these rates can only be improved by logarithmic factors under the assumption that the ground truth is sparse. 
Maybe unexpectedly, we therefore conclude that the maximum $\ell_1$-norm classifier cannot (or only very mildly) adapt to sparsity of the ground truth!

\paragraph{Suboptimality of maximum $\ell_1$-margin}
This lack of adaptivity stands in stark contrast to other 
$\ell_1$-norm constrained classifiers from the one-bit CS
literature that can e.g.
achieve rates of order $\frac{\|\wgt\|_0 \log(d)}{\sqrt{n}}$  under sparsity assumptions (e.g., \cite{zhang2014efficient,awasthi2016learning}). 
We remark that even faster min-max optimal bounds of order $\frac{\|\wgt\|_0 \log(d)}{n}$ can be obtained by other specifying classifiers \citesmart{gopi13,jacques2013robust}.
Intuitively, the reason for the sub-optimality of the rates for the maximum $\ell_1$-margin classifier can be explained by the fact that the 
ground truth $\wgt$ has a small margin of order $\Theta(\frac{1}{\sqrt{n}})$
with high probability
whereas the 
maximum $\ell_1$-margin classifier has a larger margin at least of order
\footnote{  where we make use of  Proposition~\ref{prop:localization_noiseless_ssparse} and Lemma 4.1 in \cite{chinot_2021}}
 $\Omega(\frac{1}{(n\norm{\wgt}_1)^{1/3}})$. That is, the  max-$\ell_1$-margin classifier overfits to samples close to the decision boundary.

\subsection{Main result for noisy observations}
\label{subsec:noisy}
\label{subsec:main_results_noisy}
Our second main result considers the high noise regime where a constant fraction of the labels are (randomly) corrupted with high probability. 
We show in the following theorem that the prediction error vanishes for this setting at a logarithmic rate:



\begin{theorem}[Noisy classification] \label{thm:mainl1_class_noisy}
    Assume that the corruptions $\xi_i$ follow the law in Equation~\eqref{eq:labelscor} 
    with  $\probsigma$ independent of $n,d$ and satisfying Assumption~\ref{ass:noise}. Furthermore, assume that $\wgt$ is $s$-sparse with $s\lesssim n/\log^4(d/n)$. There exist universal constants $\kappa_1,\kappa_2,\kappa_3,c_1,\dots,c_4>0$ such that
    for any $n\geq \kappa_1$ and
    $\kappa_2 n \leq d \leq \exp(\kappa_3 n^{1/5})$, the prediction error is upper- and lower-bounded by
    \begin{equation}
    \label{eq:mainboundl1noisy}
         \left\vert \RiskC(\what)  - \sqrt{\frac{\kappanoise}
          {\log(d/n)}} \right\vert \lesssim \frac{ 1}{\log^{3/4}(d/n)} \, ,
    \end{equation}
    with probability at least 
    $1 - c_1\exp \left( -c_2  \frac{n}{\log^5(d/n)}  \right) -  c_3\exp \left( -c_4 \frac{ n }{\log n \log^{3/2}(d/n)} \right)$ over the draws of the data set and with $\kappanoise$  a constant only depending on $\probsigma$ (see Equation~\eqref{eq:kappanoise} in Appendix~\ref{subsec:proof_sketch_noisy_main} for the definition). 
\end{theorem}
 The proof of the theorem is deferred to
 Appendix~\ref{subsec:proof_sketch_noisy_main} and an overview is given in Section~\ref{sec:proofsketch}. We now discuss the implications of the theorem in the following paragraphs.
 

\paragraph{Benign overfitting:}
We are first to show that the prediction error of the max-$\ell_1$-margin classifier vanishes albeit interpolating a constant fraction of (randomly) corrupted labels, and thus exhibits benign overfitting \cite{bartlett2020benign}. Our work therefore complements recent works studying maximum $\ell_p$-margin classifiers with $p>1$
that can achieve polynomial rates \citesmart{donhauser2022fast}.


\paragraph{Comparison with optimal rates:}
Although vanishing, the rates in Theorem~\ref{thm:mainl1_class_noisy} are only of logarithmic orders and therefore far from being  min-max optimal. Indeed min-max optimal lower bounds for the noisy setting are of order 
$\frac{\|\wgt\|_0 \log(d)}{\sqrt{n}}$ \citesmart{Wainwright2009InformationTheoreticLO,abramovich2018high} and attained by regularized (non-interpolating) classifiers  maximizing the average margin under $\ell_1$-norm constraints (see e.g.,  \citesmart{zhang2014efficient}). Theorem~\ref{thm:mainl1_class_noisy} can therefore also be understood as a negative result showing that the maximum $\ell_1$-margin classifier suffers from overfitting the noise.

\subsection{Discussion of the assumptions in Theorem~\ref{thm:mainl1_class_noiseless} and \ref{thm:mainl1_class_noisy}}

In this section we discuss the assumptions in our main theorems on the sparsity of the ground truth and the data distribution and their limitations.

\paragraph{Sparsity of the ground truth $\wgt$}
While the upper bound  in Theorem~\ref{thm:mainl1_class_noiseless} can be generalized at the cost of a logarithmic factor (i.e. as in \cite{chinot_2021}), the lower bound requires a very tight analysis (proof of Proposition~\ref{prop:BoundGamma0_class_noiseless} in Appendix~\ref{sec:proof_unicon_noiseless}) and strongly relies on the sparsity of the ground truth. We would like to note at this place that only few high-probability lower bounds are known in the literature (beyond classifiers/regression estimators relying on the  $\ell_2$-norm) and leave lower bounds for non-sparse ground truths as an exciting and important future work. 

Further, we remark that the constraint on the degree of the sparsity of the ground truth in Theorem 2 cannot be relaxed without affecting the upper bound. However, it is an open question whether one can relax the constraint with a soft-sparsity constraint on the ground truth of the form $\|\wgt\|_1 \leq \sqrt{\frac{n}{\log(d/n)^4}}$. 
Finally, we note that the bounds in Theorem~\ref{thm:mainl1_class_noisy} do not depend on the ground truth, assuming that the degree of the sparsity is sufficiently small. Morally, this  is because  the effect of fitting the noise dominates the prediction error, similar to the rates for the prediction error of the minimum-$\ell_1$-norm interpolator (Basis pursuit) in \citesmart{wang2021tight}.

\paragraph{Data distribution beyond Gaussian} The assumption that the data is normally distributed is a
major limitation of the results presented in Theorem~\ref{thm:mainl1_class_noiseless} and \ref{thm:mainl1_class_noisy}. Attempts to generalize this assumption face the fundamental issue that the analysis needs to be tight including multiplicative constants. For instance, to obtain the logarithmic factors in the upper bounds as well as the matching lower bounds in Theorem~\ref{thm:mainl1_class_noiseless} and \ref{thm:mainl1_class_noisy}, we require the 
upper bounds for $\PhiCNuni \leq M$ (see Section~\ref{sec:proofsketch}), presented Proposition~\ref{prop:localization_noiseless_ssparse} and \ref{prop:Rbound_noisy} in Appendix~\ref{subsec:proof_sketch_noiseless_main} and \ref{subsec:proof_sketch_noisy_main}, to be exact including the multiplicative constants. 

However, generalizations of tools relying on Gaussian comparison to sub-Gaussian data  usually come at a price of a multiplicative constant. For instance, this is already the case for the famous Talagrand's majorizing measure theorem \citesmart{talagrand2005generic} which can be seen as a generalization of Slepian's lemma (see e.g., \cite{vershynin2018high}) to sub-Gaussian data. 
Nevertheless, deriving tight generalizations of the (C)GMT \citesmart{gordon_1988,thrampoulidis_2015} is a promising direction for extending our  main results (Theorem~\ref{thm:mainl1_class_noiseless} and \ref{thm:mainl1_class_noisy}) to non-Gaussian data, with first results in this direction in \citesmart{han2022universality}.

\paragraph{Non-isotropic features}
In this paper, we favor simplicity over generality and  only consider isotropic input features $x_i$.  Technically, our methodology relying on Gaussian comparison (Proposition~\ref{prop:CGMT_application_classification} in Subsection~\ref{subsec:proof_sketch_gmt}) also applies to non-isotropic features, as recently done in \citesmart{koehler2021uniform,zhou2021optimistic,zhou2022non}) for general estimators/classifiers. However, in order to obtain tight bounds as in Theorem~\ref{thm:mainl1_class_noiseless} and ~\ref{thm:mainl1_class_noisy}, we require a  fine grained analysis, and in particular, we need to be able to control the norms of the parametric curve $\gamma(\alpha)$  (see Section~\ref{subsec:proof_sketch}).  Generalizing this argument to non-isotropic features  poses many technical difficulties, and  is left as an interesting task for future work.

\subsection{Comparison with bounds relying on hyperplane tessellation}
\label{subsec:hyperplane}

We now discuss the limitations of proofs relying on hyperplane tessellation (see e.g. \cite{plan2014dimension}) -- a standard tool to bound the prediction error of linear classifier in high-dimensional settings, e.g. in \citesmart{chinot_2021}.

First, define the Hamming distance of two vectors $w_1, w_2$ to be the fraction of training samples where the corresponding classifiers differ:
\begin{equation}
    d_H(w_1, w_2) = \frac{1}{n} \sum_i  \indicator{\sign(\langle x_i, w_1 \rangle \neq \sign(\langle x_i, w_2 \rangle) }.
\end{equation}
Note that  $d_H(\what,\wstar)$ corresponds exactly to 
the fraction of corrupted labels  i.e.,  $d_H(\what, \wstar) = {\frac{1}{n} \sum_i \indicator{\xi_i = -1}}$. 
The high-level idea of hyperplane tessellation is to bound the directional estimation error~\eqref{eq:risk_directional_estimation_error} (which in turn gives a bound on the prediction error \eqref{eq:prediction_error_cla})  via the Hamming distance by uniformly bounding  the  difference between the Euclidean and scaled Hamming distance 
\begin{equation}
\label{eq:hyperplanetes}
    \sup_{w_1,w_2 \in T} \vert \lambda d_H(w_1, w_2) - \|w_1 - w_2\|_2\vert, 
\end{equation}
over some large enough set $T \subset S^{d-1}$ that contains the normalized classifier $\frac{\hat{w}}{\|\hat{w}\|_2 }$ with high probability. Here, $\lambda$ is some universal constant. 


Observe that this approach only leads to  tight bounds if the difference in Equation \eqref{eq:hyperplanetes} is small.
This, however, is not the case for the settings studied in our main results. Indeed, for noisy data (Theorem~\ref{thm:mainl1_class_noisy}), by definition of the 
interpolating classifier we have that
\begin{equation}
    \lambda d_H\left(\frac{\hat{w}}{\|\hat{w}\|_2} , \wgt\right) = \Theta(1)
\end{equation}
while $\| \frac{\hat{w}}{\|\hat{w}\|_2} - \wgt \|_2 $ vanishes at a logarithmic rate. Furthermore, in the noiseless case (Theorem~\ref{thm:mainl1_class_noiseless}), the Hamming distance $d_H\left(\frac{\hat{w}}{\|\hat{w}\|_2}, \wgt\right) $ is zero  --- meaning that we cannot obtain any lower bounds for the directional estimation error using a hyperplane tessellation argument.

This ``weakness'' of proofs relying on uniform hyperplane tessellation bounds is also not surprising since such approaches do not take the distributional assumptions of the noise into account --- in particular we cannot distinguish between adversarial and non-adversarial noise.
In contrast, the logarithmic rates in Theorem~\ref{thm:mainl1_class_noisy} crucially rely on Assumption~\ref{ass:noise} for the distribution of the corruptions. 
However, in defense of hyperplane tessellation bounds, we finally mention that unlike the proofs presented in this paper (see Section~\ref{sec:proofsketch}), results relying on hyperplane tessellation bounds give guarantees for arbitrary corruptions and can also be generalized to non-Gaussian features \citesmart{chinot_2021}.

\section{Proof overview}

\label{sec:proofsketch}




In this section, we give an overview of the proofs of the main results,  Theorem~\ref{thm:mainl1_class_noiseless} and Theorem~\ref{thm:mainl1_class_noisy}, and summarize the main tools used in the proof.
Both proofs rely on a standard localization/ uniform convergence argument (see e.g., \cite{koehler2021uniform,zhou2021optimistic,wang2021tight,donhauser2022fast}), where:
\begin{enumerate}
    \item \emph{(Localization)}
     we derive a  high-probability upper bound over the draws of $X$ and $\xi$ on the $\ell_1$-norm of the maximum $\ell_1$-margin interpolator $\what$,
    by finding $\loneboundclasg>0$ such that
    \begin{equation} 
        \min_{ \forall i:\:y_i \langle \xui, w \rangle \geq 1}  \norm{w}_1
        ~ =: \PhiCNuni
        ~ \leq \loneboundclasg.
    \end{equation}
    \item \emph{(Uniform convergence)}
   we  derive high-probability uniform bounds over $X$ and $\xi$ for all interpolators $w$ with  $\norm{w}_1 \leq \loneboundclasg$.
    Namely, we find a high-probability lower and upper bound, respectively, for
    \begin{align} 
        \PhiCmuni:&=\min_{\substack{
            \norm{w}_1 \leq \loneboundclasg \\
            \norm{w}_2 \geq \delta
        }}  \frac{\langle w, \wgt\rangle}{\norm{w}_2} \quad \text{s.t.}  \quad \forall i:\:y_i \langle \xui, w \rangle \geq 1
        ,\\
         \PhiCpuni:&=\max_{\substack{
            \norm{w}_1 \leq \loneboundclasg \\
            \norm{w}_2 \geq \delta
        }} \frac{\langle w, \wgt \rangle}{\norm{w}_2}\quad \text{s.t.}\quad 
             \forall i:\:y_i \langle \xui, w \rangle \geq 1
    \end{align}    
    with some $\delta >0$ arbitrarily small, 
    which in turn gives us high probability bounds for the prediction error using that \begin{equation}
        \RiskC(\what) =      \frac{1}{\pi} \arccos\left(\left\langle \frac{\what}{\norm{\what}_2}, \wgt\right\rangle\right).
    \end{equation} 
\end{enumerate}
\begin{remark}
The constraint $\norm{w}_2 \geq \delta$ in the definition of $\PhiCpuni, \PhiCmuni$ is only added to ensure the optimization problems are well defined. In particular, we can choose $\delta>0$ arbitrarily small and therefore neglect this constraint in the reminder of the analysis.
\end{remark}

The reminder of this section is structured as follows: We first present in Section~\ref{subsec:proof_sketch_gmt} an application of Gaussian comparison (Proposition~\ref{prop:CGMT_application_classification}), which allows us to reduce the optimization problems $\PhiCNuni, \PhiCmuni$ and $\PhiCpuni$ to simpler auxiliary optimisation problems $\PhiCdnuni,\PhiCdmuni$ and $\PhiCdpuni$. We then describe in Section~\ref{subsec:proof_path} how these auxiliary optimization problems can be further simplified using the localized Gaussian width (Proposition ~\ref{prop:parametrization_class_general}). 
Finally, in Section~\ref{subsec:proof_sketch}, we give a sketch for the remaining proofs of Theorem~\ref{thm:mainl1_class_noiseless} and Theorem~\ref{thm:mainl1_class_noisy}. 


\paragraph{Notation}
We define the function ${(\cdot)_+:\RR\to\RR_+}$, ${ (x)_+ = x\indicator{x\geq 0}}$. We denote with $s$ the sparsity ($\ell_0$-norm) of $\wgt$ and assume w.l.o.g. that the non-zero entries of $\wgt$ are exactly the first $s$-entries. 
Moreover, we use the following notation for components of the vector $w$: $\wpar\in\RR^d$ and $\wperp\in\RR^d$ for components parallel and perpendicular to $\wgt$, respectively. Furthermore, we use $\wperps\in\RR^s$ for the first $s$-entries of $\wperp$, and $\wperpsc\in\RR^{d-s}$ for the last $d-s$ entries of $\wperp$. 

We denote by $B_1,B_2$ unit balls with respect to the $\ell_1$ and $\ell_2$-norms, respectively. We use $\kappa_1,\kappa_2,...$ and $c_1,c_2,...$ for generic universal positive constants independent of $d$, $n$, whose value may change from display to display throughout the derivations. The standard notations $O(\cdot), o(\cdot), \Omega(\cdot), w(\cdot)$ and $\Theta(\cdot)$, as well as $\lesssim,\gtrsim$ and $\asymp$, are utilized to hide universal constants, without any hidden dependence on $d$ or $n$.
\vspace{-0.1in}

\subsection{Preliminary step 1: application of the (C)GMT}
\label{subsec:proof_sketch_gmt}
The proofs of both main results rely on the following application of the Gaussian Minmax Theorem (GMT) \citesmart{gordon_1988} and its convex variant ((C)GMT)  \citesmart{thrampoulidis_2015}, which is
also used in \citesmart{deng21,donhauser2022fast,koehler2021uniform,zhou2021optimistic,wang2021tight} for bounding the error of linear min-norm/max-margin interpolators. 
This tool gives us sharp control over $\PhiCNuni, \PhiCmuni, \PhiCpuni$ up to the exact constants, which is crucial for the subsequent steps in Section~\ref{subsec:proof_sketch}. 

Intuitively,  the (C)GMT allows us to replace the data dependent interpolation constraint $ \forall i:\:y_i \langle \xui, w \rangle \geq 1$ in $\PhiCNuni, \PhiCmuni, \PhiCpuni$ with a simpler constraint, only depending on four i.i.d.~isotropic zero mean unit variance Gaussian random vectors $\gausswone,\gausswtwo \in \mathbb R^n$, $\hone \in \mathbb R^{s},\htwo \in \mathbb R^{d-s}$.

Define the function $\fnclasuni : \RR \times \mathbb R_+ \to \RR_+$, 
\begin{align}\label{eq:fnclas_def_noisy_par}
     \fnclasg(\nu,\eta) = \frac{1}{n} \sum_{i=1}^n ( 1 - \xi_i \nu |\gausswone_i| - \gausswtwo_i \eta )_+^2
\end{align}

and the set $\tilde \Gamma\subset \mathbb R^d$,
\begin{align}
    \tilde \Gamma = \{ w\in \mathbb R^d \subjto \text{Eq.~\eqref{eq:constraint1gamma}~holds}  ~\text{and}~~ \normk{w}_1 \leq \loneboundclasuni\},
\end{align}
with
\begin{align}
      \frac{(\langle\wperps,\hone \rangle + \langle\wperpsc,\htwo\rangle)^2 }{n} 
        \geq  \fnclasuni(\langle \wpar, \wgt \rangle,\norm{\wperp}_2).  \\     \label{eq:constraint1gamma}
\end{align}
We  define the  auxiliary optimization problems by:\footnote{We define $\PhiCNuni,\PhiCmuni,\PhiCdnuni,\PhiCdmuni = \infty$ and $\PhiCpuni,\PhiCdpuni = -\infty$ if the corresponding optimization problems have no feasible solution.} 
        \ifaistats
        \begin{align}
            \PhiCdnuni &= \min_{w} \|\wpar+\wperps\|_1+ \|\wperpsc\|_1
             \subjto \text{Eq.~\eqref{eq:constraint1gamma}~holds}\\
             &\quad\quad\text{and}\quad \langle\wperps,\hone \rangle + \langle\wperpsc,\htwo \rangle  \geq 0
             \end{align}
             \begin{align}
             \PhiCdpuni &= \max_{\norm{w}_2 \geq \delta} \frac{\langle \wpar, \wgt \rangle }{\sqrt{\|\wpar\|_2^2+\|\wperps\|_2^2+\|\wperpsc\|_2^2}}
            \subjto w\in \tilde \Gamma \\
            \PhiCdmuni &= \min_{\norm{w}_2 \geq \delta} \frac{\langle \wpar, \wgt \rangle }{\sqrt{\|\wpar\|_2^2+\|\wperps\|_2^2+\|\wperpsc\|_2^2}}  \subjto w\in \tilde \Gamma
            \end{align}
        \else
        \label{eq:Gamma_def}
        \begin{gather}
            \PhiCdnuni = \min_{w} \|\wpar+\wperps\|_1+ \|\wperpsc\|_1
             \subjto \text{Eq.~\eqref{eq:constraint1gamma}~holds}
             ~~\andtxt~~ \langle\wperps,\hone \rangle + \langle\wperpsc,\htwo \rangle  \geq 0 \\
            \PhiCdpuni = \max_{\norm{w}_2 \geq \delta} \frac{\langle \wpar, \wgt \rangle }{\sqrt{\|\wpar\|_2^2+\|\wperps\|_2^2+\|\wperpsc\|_2^2}}
            \subjto w\in \tilde \Gamma \\
            \PhiCdmuni = \min_{\norm{w}_2 \geq \delta} \frac{\langle \wpar, \wgt \rangle }{\sqrt{\|\wpar\|_2^2+\|\wperps\|_2^2+\|\wperpsc\|_2^2}}  \subjto w\in \tilde \Gamma.
        \end{gather}
        \fi

Using the Lagrange multiplier, we can bring $\PhiCNuni, \PhiCmuni, \PhiCpuni$ in a suitable form to apply the (C)GMT, which after straightforward simplifications yields: 

\begin{proposition}
\label{prop:CGMT_application_classification}
For any $t \in \RR$ we have:
        \begin{align}
            \PP( \PhiCNuni > t \vert \xi) &\leq 2\PP( \PhiCdnuni \geq t \vert \xi)
             \\
            \PP( \PhiCpuni > t \vert \xi) &\leq 2\PP( \PhiCdpuni \geq t \vert \xi) 
             \\
            \PP( \PhiCmuni < t \vert \xi) &\leq 2\PP( \PhiCdmuni \leq t \vert \xi) ,
        \end{align}
        where the probabilities on the LHS and RHS are over the draws of $X$ and of $\gausswone$, $\gausswtwo,\hone,\htwo$, respectively.
\end{proposition}   
The proof of the proposition is deferred to Section~\ref{sub:proof_CGMT_technical}.

\subsection{Preliminary step 2: simplification of the auxiliary optimization problems}
\label{subsec:proof_path}
\label{subsec:path_param}
In a second step, we reduce the auxiliary optimization problems $\PhiCdnuni,\PhiCdmuni$ and $\PhiCdpuni$ to low-dimensional optimization problems. We first discuss $\PhiCdmuni$ and $\PhiCdpuni$. 

In order to reduce the two stochastic optimization problems to low dimensional optimization problems, we  relax the constraint in Equation~\eqref{eq:constraint1gamma} by bounding the term $\langle\wperps,\hone \rangle + \langle\wperpsc,\htwo \rangle $ only using the $\ell_1$ and $\ell_2$-norms of $\wperps$ and $\wperpsc$.
The first term $\langle\wperps,\hone \rangle$ can be simply upper bounded using Cauchy Schwartz: ${\langle\wperps,\hone \rangle \leq \|\hone \|_{2} \|\wperps\|_2}$ where we recall that $\hone \in \mathbb R^s$. 
However, doing the same for the second term $\langle\wperpsc,\htwo\rangle$ would result in loose bounds since $\htwo \in \mathbb R^{d-s}$. 
In fact, similar to the proof in \cite{wang2021tight} for the related min $\ell_1$-norm interpolator (basis pursuit), using Hoelders inequality to bound 
$\langle\wperpsc,\htwo \rangle \leq \|\wperpsc\|_1 \|\htwo \|_\infty$ would still result in loose bounds (see the discussion in Section~3.2 in \cite{wang2021tight}).

Instead, we make use of a more refined (tight) upper bound using the (conditional) localized Gaussian width:
\begin{equation} 
\label{eq:defolw}
    \langle\wperpsc,\htwo \rangle \leq  \|\wperpsc\|_1  \ell^*_{\htwo}\left(\frac{\normk{\wperpsc}_2}{\normk{\wperpsc}_1} B_2 \cap  B_1\right)
\end{equation}
with $\ell^*_{\htwo}: [\frac{1}{\sqrt{d}},1] \to \mathbb R_+$, 
\begin{align}
 \ell^*_{\htwo}(\beta B_2 \cap B_1):= \max_{\substack{ \norm{w}_2 \leq \beta \\
   \| w\|_1 \leq 1}} \innerprod{w}{ \htwo}.
\end{align}
In summary, we can therefore relax the constraint in Equation~\eqref{eq:constraint1gamma} occurring in $\tilde \Gamma$  to:
\ifaistats
\begin{align}
     \frac{\left(\|\wperpsc\|_1  \ell^*_{\htwo}\left(\frac{\normk{\wperpsc}_2}{\normk{\wperpsc}_1} B_2 \cap  B_1\right) + \|\hone\|_{2} \|\wperps\|_2\right)^2 }{n} 
        \\
        \geq  \fnclasuni\left(\langle \wpar, \wgtp \rangle, \sqrt{\normk{\wperpsc}_2^2 + \normk{\wperps}_2^2}\right).
        \label{eq:constraint1gammarelaxed}
\end{align}
\else
\begin{align}
    \frac{\left(\|\wperpsc\|_1  \ell^*_{\htwo}\left(\frac{\normk{\wperpsc}_2}{\normk{\wperpsc}_1} B_2 \cap  B_1\right) + \|\hone\|_{2} \|\wperps\|_2\right)^2 }{n} 
    \geq  \fnclasuni\left(\langle \wpar, \wgtp \rangle, \sqrt{\normk{\wperpsc}_2^2 + \normk{\wperps}_2^2}\right).
    \label{eq:constraint1gammarelaxed}
\end{align}
\fi
In particular, we note that the resulting relaxed optimization problems for $\PhiCdmuni$ and $\PhiCdpuni$ only depend on the $\ell_1$ and $\ell_2$-norms of $\wperps$ and $\wperpsc$ and are therefore low-dimensional. 

A similar argument can also be used to convert $\PhiCdnuni$ into a low-dimensional optimization problem. However, instead of relaxing the constraint in Equation~\eqref{eq:constraint1gamma}, we now need to tighten it.  We can do this by setting $\wperps =0$ (which is negligible assuming that $s \ll n$) and choosing $\wperpsc$ as a function of $\beta$ to be optimizer of the optimization problem defining $\ell^*_{\htwo}\left(\beta B_2 \cap  B_1\right)$, which satisfies the constraints with equality.




Similar to  the related analysis in \citesmart{wang2021tight} for the estimation error of the  minimum $\ell_1$-norm interpolator,  we parameterize the resulting low-dimensional optimization problems via the curve
 $\gamma: \alpha\in[1,\alpham] \mapsto \gamma(\alpha) \in \RR^{d-s}$,
\begin{equation}
\label{eq:def_gamma_alpha}
    \gamma(\alpha) = \argmin_{w} \norm{w}_2^2
    \subjto \begin{cases}
                \innerprodk{w}{  \vert \htwo \vert  } \geq \normk{\htwo }_{\infty}\\
                w\geq 0\\
                \transp{\onevec} w = \norm{w}_1 = \alpha
            \end{cases},
\end{equation} 
with $\alpham = (d-s)\frac{\norm{\htwo}_{\infty}}{\norm{\htwo}_1}$. We refer to Section~\ref{subsec:proof_sketch} step 3 for an explanation why this parameterization is important. Via Lagrange duality, it is straightforward to show  that for any $\beta \in [\frac{1}{\sqrt{d}},1]$, there exists $\alpha \in [1, \alpham]$ such that   $ \frac{\gamma(\alpha)}{\alpha}$ is an optimal solution for the optimization problem defining $\ell^*_{\htwo}(\beta B_2 \cap B_1)$ (see \cite{wang2021tight}).

We can now summarize the resulting low-dimensional optimization problems in the following proposition. For brevity of
notation, define $\nu:= \innerprodk{\wpar}{\wgt}$, $\etasc := \normk{\wperpsc}_2$, $\etas := \normk{\wperps}_2$,  $\eta :=\normk{\wperp}_2 = \sqrt{\etasc^2+\etas^2}$ and $b = \frac{\|\wperpsc\|_1}{\alpha}$. We have:
\begin{proposition}
\label{prop:parametrization_class_general}

Let $\smax \in \mathbb R_+$ and let $\wgt$ be any $s$-sparse vector with $s\leq \smax$. Then, the optimization problems $\PhiCdnuni,\PhiCdpuni$ and $\PhiCdmuni$ can be bounded by:
    \ifaistats
    \begin{align}
            &\PhiCdnuni \leq 
            \Big[ \min_{\nu, b\geq 0, \alpha\in[1,\alpham]} \vert \nu \vert \|\wgt\|_1  + b \norm{\gamma(\alpha)}_1 \subjto\\
            &~~~~~~~~~~~~~~~~~\frac{1}{n}   b^2\normk{\htwo}_{\infty}^2  \geq  \fnclasg(\nu, b\|\gamma(\alpha)\|_2) \Big]
            \end{align}
            \begin{align}
            &\PhiCdpuni   \leq \max_{(\nu, b, \alpha, \etas) \in \setZeroUni}  \frac{\nu}{\sqrt{\nu^2+b^2\norm{\gamma(\alpha)}_2^2 + \etas^2}}
            \\
            &\PhiCdmuni   \geq \min_{(\nu, b, \alpha, \etas) \in \setZeroUni} \frac{\nu}{\sqrt{\nu^2+b^2\norm{\gamma(\alpha)}_2^2 + \etas^2}}
    \end{align}
    \else
    \begin{gather}
            \PhiCdnuni \leq 
            \Big[ \min_{\nu, b\geq 0, \alpha\in[1,\alpham]} \vert \nu \vert \|\wgt\|_1  + b \norm{\gamma(\alpha)}_1 \subjto\frac{1}{n}   b^2\normk{\htwo}_{\infty}^2  \geq  \fnclasg(\nu, b\|\gamma(\alpha)\|_2) \Big]\\
            \PhiCdpuni   \leq \max_{(\nu, b, \alpha, \etas) \in \setZeroUni}  \frac{\nu}{\sqrt{\nu^2+b^2\norm{\gamma(\alpha)}_2^2 + \etas^2}}
            \\
            \PhiCdmuni   \geq \min_{(\nu, b, \alpha, \etas) \in \setZeroUni} \frac{\nu}{\sqrt{\nu^2+b^2\norm{\gamma(\alpha)}_2^2 + \etas^2}}
    \end{gather}
    \fi
where the last two inequalities hold with probability at least $1-2\exp(-c_1 \smax)$, with universal constant $c_1$, and constraint set $\setZeroUni$ defined by:
\ifaistats
\begin{gather}
    \setZeroUni 
    =
    \Big\{ (\nu, b, \alpha, \etas) 
    \subjto  \etas\geq 0, b\geq0, \alpha\in[1,\alpham] ~\andtxt\\
    \frac{  (2\sqrt{\smax}\etas + b \norm{\htwo}_{\infty})^2}{n}  \geq  \fnclasg(\nu,\sqrt{b^2\|\gamma(\alpha)\|_2^2+\etas^2}) 
    \\
   \andtxt~~~ \max\left\{ |\nu| \|\wgt\|_1 - \sqrt{s}\etas , 0 \right\} 
    + b\alpha\leq \loneboundclasg 
    \Big\}.
    \label{eq:Gamma0_def_class_uni}
\end{gather}
\else
\begin{align}
    \setZeroUni 
    =
    \Big\{ (\nu, b, \alpha, \etas) 
    \subjto  &\etas\geq 0, b\geq 0, \alpha\in[1,\alpham]\\
    ~~\andtxt~~ &
    \frac{  (2\sqrt{\smax}\etas + b\normk{\htwo}_{\infty})^2}{n}  \geq  \fnclasg(\nu,\sqrt{b^2\|\gamma(\alpha)\|_2^2+\etas^2}) \\
   ~~\andtxt~~ &\max\left\{ |\nu| \|\wgt\|_1 - \sqrt{s}\etas , 0 \right\} 
    + b\alpha\leq \loneboundclasg 
    \Big\}.
    \label{eq:Gamma0_def_class_uni}
\end{align}
\fi
\end{proposition} 

The proof follows straight forwardly from the above discussion and when applying standard Gaussian concentration to control the tail of the term $\|\hone\|_{2}$. 

\subsection{Proof sketch  for bounding the auxiliary optimization problems}
\label{subsec:proof_sketch}
To obtain the desired results in Theorem~\ref{thm:mainl1_class_noiseless} and \ref{thm:mainl1_class_noisy}, we recall that by Proposition~\ref{prop:CGMT_application_classification},  it suffices to find high probability bounds for $\PhiCdnuni, \PhiCdmuni, \PhiCdpuni$  using the low-dimensional relaxations in Proposition~\ref{prop:parametrization_class_general}. 
We now present the main idea for the reminder of the proof which is rigorously presented in Appendix~\ref{subsec:proof_sketch_noiseless_main} and \ref{subsec:proof_sketch_noisy_main}. 
 We only discuss lower bounding  $\PhiCdmuni$, as the proof for upper bounding $\PhiCdpuni$ follows from exactly the same idea. Furthermore, upper bounding  $\PhiCdnuni$ is conceptually easier since we only need to find a feasible point satisfying the constraints. For simplicity, we present a slightly simplified structure of the proof (compared to the one presented in Appendix~\ref{subsec:proof_sketch_noiseless_main} and \ref{subsec:proof_sketch_noisy_main}), which consists of the following steps:

 
 

\paragraph{Step 1: reducing the problem to bounding the set $\Gamma$}
As done in the proof for upper bounds on the prediction error of maximum $\ell_p$-margin classifiers with $p>1$ in \citesmart{donhauser2022fast}, we can reduce the problem of bounding $\PhiCdmuni$ to one bounding $\Gamma$ in Equation~\eqref{eq:Gamma0_def_class_uni} (where we use Proposition~\ref{prop:parametrization_class_general}):
\begin{equation}
  \PhiCdmuni \geq  \left[1 + \frac{\max\limits_{(b,\alpha) \in \setZeroUni} ~b^2\norm{\gamma(\alpha)}_2^2 + \max\limits_{\etas \in \setZeroUni} ~\etas^2}{\min\limits_{\nu \in \setZeroUni}~ \nu^2}  \right]^{-1/2}, \label{eq:proofsketchmaintextphim}
  \end{equation}
  assuming that $\min\limits_{\nu \in \setZeroUni}~ \nu^2 >0$ and where we slightly abuse the notation by writing $\nu \in \Gamma$. 
   Maybe surprisingly, this seemingly loose lower bound turns out to be tight. 
  
 \paragraph{Step 2: controlling $\fnclasuni$} 
 In order to bound the set $\Gamma$, we need to first uniformly lower bound $\fnclasuni$.
 However, to do so by applying standard uniform convergence bounds (see Theorem~\ref{thm:unifconv_adamczak} which is taken from \cite{adamczak2008tail}),
 we need to restrict $(\nu, \eta)$  with high probability to a sufficiently small set, where we recall that $\eta := \sqrt{b^2\|\gamma(\alpha)\|_2^2+\etas^2}$. 
 
More precisely, in the noisy case, we show that $(\nu, \eta)$ is contained in a neighborhood around the global minima of $\EE \fnclasg(\nu,\eta)$, attained at $(\nubar, 0)$
 with $\nubar$ some constant only depending on the noise distribution $\probsigma$. Moreover, in the noiseless case, the global minima of the mean is attained at infinity. Instead, it suffices for the following analysis to show that $\nu$ is contained in some interval  and $\eta \geq 0$ is upper bounded by a constant. 
 We omit the proof sketch for this intermediate result and only remark that we use a similar argument as used for the derivation of the tight bounds in the following paragraph but with looser approximations for~$\fnclasuni$.

 As a consequence, we can apply standard uniform convergence arguments (Theorem~\ref{thm:unifconv_adamczak}) to show concentration of $\fnclasg$ around its mean
    $\fnclasg(\nu,\eta)
    \approx \EE \fnclasg(\nu,\eta)$.
In particular, we make heavily use of the key observation that 
 $ \EE \fnclasg(\nu,\eta)$ locally behaves like a quadratic function in $\eta$: 
\begin{align}
    \text{noiseless}\quad &\mathbb E \fnclasg(\nu,\eta)\approx
    \frac{\sqrt{2}}{3\sqrt{\pi}} \frac{1}{\nu} + \sqrt{\frac{2}{\pi}} \frac{\eta^2}{\nu} \label{eq:noislessinvnudependency}\\
    \text{noisy} \quad &\mathbb E \fnclasg(\nu,\eta) \approx \fstar + \frac{1}{2}\feestar\eta^2 + \frac{1}{2}\fnnstar \deltanu^2
    \end{align}
    with $\feestar, \fnnstar$ some constants only depending on $\probsigma$ and $\deltanu =  \nu - \nubar$. 
    \vspace{-0.15in}
\paragraph{Step 3: bounding the set $\Gamma$}
The local quadratic  behavior in $\eta$ of $\mathbb E \fnclas(\nu, \eta)$ allows us to rely on ideas developed in \citesmart{wang2021tight}, where the authors present error bounds for the related minimum $\ell_1$-norm interpolator. Similar to this proof, there the proof can essentially be reduced to studying 
a corresponding set $\Gamma_R$ which consists
of the two constraints $ b \alpha \leq B$ and ${\frac{b^2 \|\hsample \|_{\infty}^2}{n} \geq \sigma^2 + b^2 \|\gamma(\alpha)\|_2^2}$ with  variance of the noise~$\sigma^2$. \footnote{the definition for $b$ in  \cite{wang2021tight} is slightly different from the one used in this paper.}

We here only discuss the ``simpler'' noisy case which follows from an extension of the argument in \citesmart{wang2021tight}. Furthermore,  the noiseless relies on a similar argument, however, requires a much more careful analysis (described in Appendix~\ref{subsec:proof_sketch_noiseless_main}) due to the inverse proportional dependency of  ${\mathbb E \fn(\nu, \eta) \approx 
    \frac{\sqrt{2}}{3\sqrt{\pi}} \frac{1}{\nu} + \sqrt{\frac{2}{\pi}} \frac{\eta^2}{\nu}}$  on $\nu$. 

By the previous step 2 and when neglecting concentration terms, we can relax the constraints in $\Gamma$ in Equation~\eqref{eq:Gamma0_def_class_uni}
to 
\ifaistats
\begin{align}
\Gamma &\subset \{  (\nu, b, \alpha, \etas) \subjto   b \alpha \leq M \quad \andtxt\\
    &\frac{  b^2\normk{ \htwo }_{\infty}^2}{n}  \geq \fstar + \frac{1}{2}\feestar( \etas^2 + b^2 \|\gamma(\alpha)\|_2^2) + \frac{1}{2}\fnnstar \deltanu^2
    \}
\end{align} 
\else
\begin{align}
\Gamma \subset \{  (\nu, b, \alpha, \etas) \subjto   b \alpha \leq M ~~~\andtxt~~~
    \frac{  b^2\normk{\htwo }_{\infty}^2}{n}  \geq \fstar + \frac{1}{2}\feestar( \etasc^2 + b^2 \|\gamma(\alpha)\|_2^2) + \frac{1}{2}\fnnstar \deltanu^2
    \}
\end{align}
\fi
where we recall that $\deltanu = \nu - \nubar$. Furthermore, we dropped  the term $2\sqrt{\smax}\etas$ here since it is dominated by the much larger term $b\norm{\htwo}_{\infty}$. 

Using this relaxed constraint, we can upper bound $\max_{ (b,\alpha) \in \Gamma } b^2 \norm{\gamma(\alpha)}_2^2 $ by
\ifaistats
\begin{align}
    &\max_{b,\alpha}  b^2 \norm { \gamma ( \alpha ) }_2^2 \\ 
    &\subjto
    \frac{  b^2\normk{\htwo}_{\infty}^2}{n}  \geq \fstar + \frac{1}{2}\feestar b^2 \|\gamma(\alpha)\|_2^2~~\andtxt~~ b\alpha\leq \loneboundclasuni \\
    \leq
    &\max_{\alpha}   \frac{ \loneboundclasuni^2 \norm { \gamma ( \alpha ) }_2^2 }{ \alpha^2 } 
    \\
    &\subjto
    \frac{  \loneboundclasuni^2 \normk{\htwo}_{\infty}^2}{n}  \geq \fstar\alpha^2 + \frac{1}{2}\feestar \loneboundclasuni^2 \|\gamma(\alpha)\|_2^2 \label{eq:proof_sketch_sec3_conv}
\end{align}
\else
\begin{align}
    &\max_{b,\alpha}  b^2 \norm { \gamma ( \alpha ) }_2^2  
    \subjto
    \frac{  b^2\normk{\htwo}_{\infty}^2}{n} \geq \fstar + \frac{1}{2}\feestar b^2 \|\gamma(\alpha)\|_2^2 ~~\andtxt~~ b\alpha\leq \loneboundclasuni 
    \\
    &\leq \max_{\alpha}   \frac{ \loneboundclasuni^2 \norm { \gamma ( \alpha ) }_2^2 }{ \alpha^2 } 
    \subjto
    \frac{  \loneboundclasuni^2 \normk{\htwo}_{\infty}^2}{n}  \geq \fstar\alpha^2 + \frac{1}{2}\feestar \loneboundclasuni^2 \|\gamma(\alpha)\|_2^2 \label{eq:proof_sketch_sec3_conv}
\end{align}
\fi
The key idea here is that we can leverage two properties of $\gamma(\alpha)$ (see Section~4 in \cite{wang2021tight}): the function $\alpha\mapsto \frac{ \norm { \gamma ( \alpha ) }_2^2 }{ \alpha^2 } $ is monotonically decreasing and the constraint in Equation~\eqref{eq:proof_sketch_sec3_conv} is convex in $\alpha$.  Thus, we obtain the desired bounds when making use of tight concentration bounds for $\|\gamma(\alpha)\|_2$ in \citesmart{wang2021tight},  summarized in Appendix~\ref{sec:preliminary_tools_appendix}. 
Finally, the bounds for $\etas^2$ (and $\deltanu^2$) follow from a similar argument. 


In summary, we can then show that  $\nu \approx \nubar = \Theta(1)$, $b^2 \|\gamma(\alpha)\|_2^2 = O\left(\frac{1}{\log(d/n)}\right)$ and $\etas^2 = O\left(\frac{1}{\log^{3/2}(d/n)}\right)$, which gives the desired results when inserting into Equation~\eqref{eq:proofsketchmaintextphim}. \\\\\\

\paragraph{Novelty of the proofs compared to the analysis in \cite{wang2021tight}}
We now discuss the major technical contributions of this paper compared to the recent work \cite{wang2021tight}, which are two-folds:

\emph{Noisy case, Theorem~\ref{thm:mainl1_class_noisy}:} In the noisy case we reduce the analysis of the maximum-$\ell_1$-margin classifier to the one of the  minimum-$\ell_1$-norm interpolators for  regression introduced in  \citesmart{wang2021tight}. 
The key observation here lies in the quadratic behavior of the function  $f$ around its global minima (summarized in Step 2 in Section~\ref{subsec:proof_sketch}). To be able to use the tools in \citesmart{wang2021tight}, we first need to show that $\Gamma$ is contained in a sufficiently small set (Step 1-2 in the proof of Proposition~\ref{prop:BoundGamma0_class_noisy}), and then uniformly control the difference $\vert f_n -f\vert$ (Appendix~\ref{subsec:appendix_lower_bounds_fnclas}). 

\emph{Noiseless case, Theorem~\ref{thm:mainl1_class_noiseless}:} The proof of Theorem 1 is more involved than the proof of Theorem ~\ref{thm:mainl1_class_noisy}, but builds upon the same framework (Section \ref{subsec:proof_sketch_gmt}, \ref{subsec:path_param}).  The major  challenge arises from the inverse dependency of the approximation of the function $f$ on $\nu$ (see Equation~\eqref{eq:noislessinvnudependency}), which needs to be carefully controlled in the Steps 1-3 in the proof of Proposition~\ref{prop:BoundGamma0_class_noiseless}. Furthermore,  we can no longer relax the constraint $b\alpha + \vert |\nu| \|w^*\|_1 - \sqrt{s} \eta_{\mathcal S} \vert \leq M$ (as in Step 3 in Section~\ref{subsec:proof_sketch}) to $b \alpha \leq M$, but instead have to simultaneously control all quantities $b, \alpha, \nu$ and $\eta_{\mathcal S}$. 
\ifarxiv
\section{Related Work} 
\else
\section{RELATED WORK} 
\fi
\label{sec:related_work}
In this section, we discuss related work on existing  bounds for the prediction error of linear maximum margin classifiers, as well as tools that have so far been used to bound it. 

\paragraph{Related work on error bounds for maximum-margin classifiers}
Existing non-asymptotic upper bounds for the  maximum $\ell_1$-margin classifier in high-dimensional settings  hold for arbitrary (adversarial) corruptions and are discussed in detail in Section~\ref{subsec:hyperplane}. Moreover, complementary work \citesmart{liang2022precise} studies asymptotic  proportional regimes ($n,d \to \infty$ and $\frac{d}{n}\to c$) where the prediction error does not vanish.

Beyond the $\ell_1$ norm, several works present non-asymptotic bounds for the related maximum $\ell_p$-margin classifiers for $p>1$. The paper \citesmart{donhauser2022fast} studies the case where $p\in (1,2)$ for  $1$-sparse ground truths and shows that the prediction error can even vanish at polynomial rates close to the min-max lower bounds when trained on noisy dataset. 
Furthermore, the papers \citesmart{muthukumar2021classification,wang2021benign,shamir22} present bounds for the case where $p=2$ based on specific proof techniques relying on the geometry of the Euclidean $\ell_2$-norm. However, they only obtain vanishing rates, i.e. achieve benign overfitting, when assuming that the covariance matrix is spiked (i.e., for non-isotropic features).


\paragraph{Related work on proof techniques}
The proofs in this paper rely on Gaussian comparison results \citesmart{gordon_1988,thrampoulidis_2015} described in detail in Section~\ref{sec:proofsketch} and popularized for non-asymptotic bounds for linear interpolators in \citesmart{koehler2021uniform}. This technique has also recently  been used in the paper \citesmart{donhauser2022fast} to bound the prediction error of the maximum $\ell_p$-margin classifier when $p\in (1,2)$. However, the presented analysis in the mentioned paper would  yield loose bounds when $p = 1$ and is limited to noisy regimes and $1$-sparse ground truths.

Other common proof techniques for bounding the prediction error of interpolating linear classifiers include hyperplane tessellation bounds \citesmart{plan2014dimension,chinot_2021}, discussed in detail in Section~\ref{subsec:hyperplane}, and proliferation of support vector results \citesmart{muthukumar2021classification,hsu2021proliferation,wang2021benign,ardeshir2021support}. 
The idea of the latter approach is essentially to reduces the maximum margin classifier to an (approximately) equivalent minimum-norm interpolating classifier. The resulting ``simpler'' classifier can then be analyzed using tools from regression \citesmart{muthukumar2021classification,bartlett2020benign}.  However, so far such an approach only exists for the  maximum $\ell_2$-margin classifiers and it is an open conjecture to prove that proliferation of support vector results also apply to the maximum $\ell_1$-margin classifier \citesmart{ardeshir2021support}. \\\\\\






\ifaistats
\section{FUTURE WORK}
\else
\section{Future work} 
\fi
In this section, we discuss potentially interesting avenues for future work.



\paragraph{Early stopped coordinate descent} The bounds presented in this paper imply that the maximum $\ell_1$-margin classifier are not only only sub-optimal in noisy settings (Theorem~\ref{thm:mainl1_class_noisy}), but also for noiseless data (Theorem~\ref{thm:mainl1_class_noiseless}). As discussed in Section~\ref{subsec:noiseless}, this is because the classifier overfits on samples close to the decision boundary. In contrast, $\ell_1$-norm penalized classifiers which maximize the average margin \citesmart{zhang2014efficient} achieve much faster rates than $\|\wgt\|_1^{2/3} n^{-1/3}$. An interesting question for future work is whether these faster rates can be obtained for early stopped coordinate descent on  exponential losses, where we recall that the solutions of these algorithms converge (after infinite steps) to the maximum $\ell_1$-margin classifier \citesmart{telgarsky_13}. 

\paragraph{Future work on ``better'' implicit biases}
When samples in the training data have small margin  to the ground truth (see discussion in Section~\ref{subsec:noiseless}), our
 results in this paper suggest that the implicit bias of boosting methods with exponential loss functions and coordinate descent is  sub-optimal. Indeed, the maximum $\ell_1$-margin classifier which is obtained at convergence \citesmart{telgarsky_13}  only achieves sub-optimal rates even in the noiseless setting (see Theorem~\ref{thm:mainl1_class_noiseless} and subsequent discussion). An interesting direction for future work is therefore to investigate whether the implicit bias of the mentioned iterative training algorithms with  other loss functions such as polynomial losses would yield faster rates.




\ifaistats
\section{CONCLUSION} 
\else
\section{Conclusion} 
\fi

In our main results, Theorem~\ref{thm:mainl1_class_noiseless} and \ref{thm:mainl1_class_noisy}, we present tight matching non-asymptotic upper and lower bounds for the prediction error of the  maximum $\ell_1$-margin classifier, both in noiseless and noisy regimes. We thereby answer two open problems in the literature: maybe surprisingly, as a first result (Theorem~\ref{thm:mainl1_class_noiseless}), we show that the classifier is not adaptive to sparsity in a standard (noiseless) discriminate data model.
Further, as a second result (Theorem~\ref{thm:mainl1_class_noisy}), we show that the prediction error vanishes at a logarithmic rate despite interpolating a constant fraction of (randomly) corrupted labels, and thus that the classifier attains benign overfitting. 




\section*{Acknowledgements}
K.D.~is supported by the ETH AI Center and the ETH Foundations of Data Science.


\bibliographystyle{alpha}

\bibliography{references}

\newcommand{\etalchar}[1]{$^{#1}$}
\begin{thebibliography}{CKLvdG21}

\bibitem[ABHZ16]{awasthi2016learning}
Pranjal Awasthi, Maria-Florina Balcan, Nika Haghtalab, and Hongyang Zhang.
\newblock Learning and 1-bit compressed sensing under asymmetric noise.
\newblock In {\em Proceedings of the Conference on Learning Theory (COLT)},
  pages 152--192, 2016.

\bibitem[Ada08]{adamczak2008tail}
Radoslaw Adamczak.
\newblock A tail inequality for suprema of unbounded empirical processes with
  applications to markov chains.
\newblock {\em Electronic Journal of Probability}, 13:1000--1034, 2008.

\bibitem[AG18]{abramovich2018high}
Felix Abramovich and Vadim Grinshtein.
\newblock High-dimensional classification by sparse logistic regression.
\newblock {\em IEEE Transactions on Information Theory}, 65(5):3068--3079,
  2018.

\bibitem[ASH21]{ardeshir2021support}
Navid Ardeshir, Clayton Sanford, and Daniel Hsu.
\newblock Support vector machines and linear regression coincide with very
  high-dimensional features.
\newblock {\em Advances in Neural Information Processing Systems (NeurIPS)},
  2021.

\bibitem[BB08]{boufounos2008}
Petros~T Boufounos and Richard~G Baraniuk.
\newblock 1-bit compressive sensing.
\newblock {\em Proceedings of the Conference on Information Sciences and
  Systems}, pages 16--21, 2008.

\bibitem[BFLS98]{bartlett98}
Peter Bartlett, Yoav Freund, Wee~Sun Lee, and Robert~E. Schapire.
\newblock {Boosting the margin: a new explanation for the effectiveness of
  voting methods}.
\newblock {\em Annals of Statistics}, 26(5):1651 -- 1686, 1998.

\bibitem[BHMM19]{belkin_2019}
Mikhail Belkin, Daniel Hsu, Siyuan Ma, and Soumik Mandal.
\newblock Reconciling modern machine-learning practice and the classical
  bias–variance trade-off.
\newblock {\em Proceedings of the National Academy of Sciences},
  116(32):15849--15854, 2019.

\bibitem[BLLT20]{bartlett2020benign}
Peter~L Bartlett, Philip~M Long, G{\'a}bor Lugosi, and Alexander Tsigler.
\newblock Benign overfitting in linear regression.
\newblock {\em Proceedings of the National Academy of Sciences},
  117(48):30063--30070, 2020.

\bibitem[CKLvdG21]{chinot_2021}
Geoffrey Chinot, Felix Kuchelmeister, Matthias L{\"o}ffler, and Sara van~de
  Geer.
\newblock Adaboost and robust one-bit compressed sensing.
\newblock {\em arXiv preprint arXiv:2105.02083}, 2021.

\bibitem[CT06]{candes2006near}
Emmanuel~J Candes and Terence Tao.
\newblock Near-optimal signal recovery from random projections: Universal
  encoding strategies?
\newblock {\em IEEE transactions on information theory}, 52(12):5406--5425,
  2006.

\bibitem[DKT21]{deng21}
Zeyu Deng, Abla Kammoun, and Christos Thrampoulidis.
\newblock {A model of double descent for high-dimensional binary linear
  classification}.
\newblock {\em Information and Inference: A Journal of the IMA}, 2021.

\bibitem[DKW56]{dvoretzky1956asymptotic}
Aryeh Dvoretzky, Jack Kiefer, and Jacob Wolfowitz.
\newblock Asymptotic minimax character of the sample distribution function and
  of the classical multinomial estimator.
\newblock {\em The Annals of Mathematical Statistics}, pages 642--669, 1956.

\bibitem[Don06]{donoho2006compressed}
David~L Donoho.
\newblock Compressed sensing.
\newblock {\em IEEE transactions on information theory}, 52(4):1289--1306,
  2006.

\bibitem[DRSY22]{donhauser2022fast}
Konstantin Donhauser, Nicol{\`o} Ruggeri, Stefan Stojanovic, and Fanny Yang.
\newblock Fast rates for noisy interpolation require rethinking the effect of
  inductive bias.
\newblock In {\em Proceedings of the 39th International Conference on Machine
  Learning}, volume 162, pages 5397--5428, 17--23 Jul 2022.

\bibitem[FS97]{freund1997decision}
Yoav Freund and Robert~E Schapire.
\newblock A decision-theoretic generalization of on-line learning and an
  application to boosting.
\newblock {\em Journal of computer and system sciences}, 55(1):119--139, 1997.

\bibitem[GLSS18]{gunasekar2018characterizing}
Suriya Gunasekar, Jason Lee, Daniel Soudry, and Nathan Srebro.
\newblock Characterizing implicit bias in terms of optimization geometry.
\newblock In {\em Proceedings of the International Conference on Machine
  Learning (ICML)}, pages 1832--1841, 2018.

\bibitem[GNJN13]{gopi13}
Sivakant Gopi, Praneeth Netrapalli, Prateek Jain, and Aditya~V. Nori.
\newblock One-bit compressed sensing: Provable support and vector recovery.
\newblock In {\em Proceedings of the International Conference on Machine
  Learning (ICML)}, pages 154--162, 2013.

\bibitem[Gor88]{gordon_1988}
Yehoram Gordon.
\newblock On {Milman}'s inequality and random subspaces which escape through a
  mesh in rn.
\newblock In {\em Geometric aspects of functional analysis}, pages 84--106.
  Springer, 1988.

\bibitem[HMX21]{hsu2021proliferation}
Daniel Hsu, Vidya Muthukumar, and Ji~Xu.
\newblock On the proliferation of support vectors in high dimensions.
\newblock {\em Proceedings of the International Conference on Artificial
  Intelligence and Statistics (AISTATS)}, pages 91--99, 2021.

\bibitem[HS22]{han2022universality}
Qiyang Han and Yandi Shen.
\newblock Universality of regularized regression estimators in high dimensions.
\newblock {\em arXiv preprint arXiv:2206.07936}, 2022.

\bibitem[JLBB13]{jacques2013robust}
Laurent Jacques, Jason~N Laska, Petros~T Boufounos, and Richard~G Baraniuk.
\newblock Robust 1-bit compressive sensing via binary stable embeddings of
  sparse vectors.
\newblock {\em IEEE transactions on information theory}, 59(4):2082--2102,
  2013.

\bibitem[KZSS21]{koehler2021uniform}
Frederic Koehler, Lijia Zhou, Danica~J Sutherland, and Nathan Srebro.
\newblock Uniform convergence of interpolators: Gaussian width, norm bounds and
  benign overfitting.
\newblock {\em Advances in Neural Information Processing Systems (NeurIPS)},
  34, 2021.

\bibitem[Led92]{Ledoux1992AHS}
Michel Ledoux.
\newblock A heat semigroup approach to concentration on the sphere and on a
  compact riemannian manifold.
\newblock {\em Geometric \& Functional Analysis GAFA}, 2:221--224, 1992.

\bibitem[LS22]{liang2022precise}
Tengyuan Liang and Pragya Sur.
\newblock A precise high-dimensional asymptotic theory for boosting and
  minimum-l1-norm interpolated classifiers.
\newblock {\em Annals of Statistics}, 50:1669--1695, 2022.

\bibitem[Mas90]{massart}
P.~Massart.
\newblock {The Tight Constant in the Dvoretzky-Kiefer-Wolfowitz Inequality}.
\newblock {\em The Annals of Probability}, 18(3):1269 -- 1283, 1990.

\bibitem[MNS{\etalchar{+}}21]{muthukumar2021classification}
Vidya Muthukumar, Adhyyan Narang, Vignesh Subramanian, Mikhail Belkin, Daniel
  Hsu, and Anant Sahai.
\newblock Classification vs regression in overparameterized regimes: Does the
  loss function matter?
\newblock {\em Journal of Machine Learning Research (JMLR)},
  22(1):10104--10172, 2021.

\bibitem[PV12]{plan2012robust}
Yaniv Plan and Roman Vershynin.
\newblock Robust 1-bit compressed sensing and sparse logistic regression: A
  convex programming approach.
\newblock {\em IEEE Transactions on Information Theory}, 59(1):482--494, 2012.

\bibitem[PV14]{plan2014dimension}
Yaniv Plan and Roman Vershynin.
\newblock Dimension reduction by random hyperplane tessellations.
\newblock {\em Discrete \& Computational Geometry}, 51(2):438--461, 2014.

\bibitem[RDSR04]{rudin2004dynamics}
Cynthia Rudin, Ingrid Daubechies, Robert~E Schapire, and Dana Ron.
\newblock The dynamics of adaboost: cyclic behavior and convergence of margins.
\newblock {\em Journal of Machine Learning Research (JMLR)}, 5(10), 2004.

\bibitem[SF13]{schapire2013boosting}
Robert~E Schapire and Yoav Freund.
\newblock Boosting: Foundations and algorithms.
\newblock {\em Kybernetes}, 2013.

\bibitem[Sha22]{shamir22}
Ohad Shamir.
\newblock The implicit bias of benign overfitting.
\newblock In Po-Ling Loh and Maxim Raginsky, editors, {\em Proceedings of the
  Conference on Learning Theory (COLT)}, volume 178, pages 448--478, 02--05 Jul
  2022.

\bibitem[SSS10]{shalev2010equivalence}
Shai Shalev-Shwartz and Yoram Singer.
\newblock On the equivalence of weak learnability and linear separability: New
  relaxations and efficient boosting algorithms.
\newblock {\em Machine learning}, 80(2):141--163, 2010.

\bibitem[Tal05]{talagrand2005generic}
Michel Talagrand.
\newblock {\em The generic chaining: upper and lower bounds of stochastic
  processes}.
\newblock Springer Science \& Business Media, 2005.

\bibitem[Tel13]{telgarsky_13}
Matus Telgarsky.
\newblock Margins, shrinkage, and boosting.
\newblock {\em Proceedings of the International Conference on Machine Learning
  (ICML)}, 28(2):307--315, 2013.

\bibitem[Tib96]{tibshirani_1996}
Robert Tibshirani.
\newblock Regression shrinkage and selection via the lasso.
\newblock {\em Journal of the Royal Statistical Society}, 58(1):267--288, 1996.

\bibitem[TOH15]{thrampoulidis_2015}
Christos Thrampoulidis, Samet Oymak, and Babak Hassibi.
\newblock Regularized linear regression: {A} precise analysis of the estimation
  error.
\newblock {\em Conference on Learning Theory}, pages 1683--1709, 2015.

\bibitem[VdG08]{vandegeer_2008}
Sara~A. Van~de Geer.
\newblock High-dimensional generalized linear models and the lasso.
\newblock {\em Annals of Statistics}, 36(2):614--645, 2008.

\bibitem[Ver18]{vershynin2018high}
Roman Vershynin.
\newblock {\em High-dimensional probability: An introduction with applications
  in data science}, volume~47.
\newblock Cambridge university press, 2018.

\bibitem[Wai09]{Wainwright2009InformationTheoreticLO}
Martin~J. Wainwright.
\newblock Information-theoretic limits on sparsity recovery in the
  high-dimensional and noisy setting.
\newblock {\em IEEE Transactions on Information Theory}, 55:5728--5741, 2009.

\bibitem[Wai19]{wainwright2019high}
Martin~J Wainwright.
\newblock {\em High-dimensional statistics: A non-asymptotic viewpoint},
  volume~48.
\newblock Cambridge University Press, 2019.

\bibitem[WDY22]{wang2021tight}
Guillaume Wang, Konstantin Donhauser, and Fanny Yang.
\newblock Tight bounds for minimum l1-norm interpolation of noisy data.
\newblock {\em Proceedings of the International Conference on Artificial
  Intelligence and Statistics (AISTATS)}, 2022.

\bibitem[WMT21]{wang2021benign}
Ke~Wang, Vidya Muthukumar, and Christos Thrampoulidis.
\newblock Benign overfitting in multiclass classification: All roads lead to
  interpolation.
\newblock {\em Advances in Neural Information Processing Systems (NeurIPS)},
  34:24164--24179, 2021.

\bibitem[Wri15]{wright2015coordinate}
Stephen~J Wright.
\newblock Coordinate descent algorithms.
\newblock {\em Mathematical Programming}, 151(1):3--34, 2015.

\bibitem[ZBH{\etalchar{+}}21]{zhang_2021}
Chiyuan Zhang, Samy Bengio, Moritz Hardt, Benjamin Recht, and Oriol Vinyals.
\newblock Understanding deep learning (still) requires rethinking
  generalization.
\newblock {\em Communications of the ACM}, 64(3):107--115, 2021.

\bibitem[ZKS{\etalchar{+}}22]{zhou2022non}
Lijia Zhou, Frederic Koehler, Pragya Sur, Danica~J Sutherland, and Nathan
  Srebro.
\newblock A non-asymptotic moreau envelope theory for high-dimensional
  generalized linear models.
\newblock {\em arXiv preprint arXiv:2210.12082}, 2022.

\bibitem[ZKSS21]{zhou2021optimistic}
Lijia Zhou, Frederic Koehler, Danica~J Sutherland, and Nathan Srebro.
\newblock Optimistic rates: A unifying theory for interpolation learning and
  regularization in linear regression.
\newblock {\em arXiv preprint arXiv:2112.04470}, 2021.

\bibitem[ZY05]{zhang2005boosting}
Tong Zhang and Bin Yu.
\newblock Boosting with early stopping: Convergence and consistency.
\newblock {\em Annals of Statistics}, 33(4):1538--1579, 2005.

\bibitem[ZYJ14]{zhang2014efficient}
Lijun Zhang, Jinfeng Yi, and Rong Jin.
\newblock Efficient algorithms for robust one-bit compressive sensing.
\newblock In {\em International Conference on Machine Learning}, pages
  820--828, 2014.

\end{thebibliography}
\newpage


\appendix


%

%

\ifaistats
\onecolumn
\aistatstitle{Supplementary Material:\\ Tight bounds for maximum $\ell_1$-margin classifiers}
\fi

\ifaistats
\section{Preliminary technical tools}
\else
    \ifarxiv
    \section{Preliminary technical tools}
    \else
    \subsection{Preliminary technical tools}
    \fi
\fi
\label{sec:preliminary_tools_appendix}
The purpose of this section is to cite existing technical tools and simple corollaries of these results. In subsection~\ref{subsubsec:prop_gamma_alpha} we give some properties of the parametric path $\gamma(\alpha)$ introduced in \cite{wang2021tight}, which we used for reparametrization of optimization problems in preliminary step 2 in Section~\ref{subsec:proof_path}. Afterwards, in subsection~\ref{subsec:uniform_conv_tools} we recall a few concentration results, which we make use of in proving localization and uniform convergence propositions of Theorems~\ref{thm:mainl1_class_noiseless} and ~\ref{thm:mainl1_class_noisy}.

\subsection{A few good properties of \texorpdfstring{$\gamma(\alpha)$}{gammaalpha}}
\label{subsubsec:prop_gamma_alpha}

First, recall from Section~\ref{subsec:proof_sketch_gmt} that $h_2\in\RR^{d-s}$ contains samples of i.i.d. standard Gaussian random variables, and for the sake of brevity of notation we define $\hgaussian:=\abs{h_2}$. Moreover, recall the definition of the function $\gamma(\alpha):\RR\to\RR^{d-s}$ from Equation~\eqref{eq:def_gamma_alpha}:
\begin{equation}
    \gamma(\alpha) = \argmin_{w} \norm{w}_2^2
    \subjto \begin{cases}
                \innerprodk{w}{\hgaussian} \geq \normk{\hgaussian}_{\infty}\\
                w\geq 0\\
                \transp{\onevec} w = \norm{w}_1 = \alpha
            \end{cases}
\end{equation}
for some scalar variables $b\geq 0$, $\alpha\in[1,(d-s)\frac{\norm{\hgaussian}_{\infty}}{\norm{\hgaussian}_1}]$. Without loss of generality we can assume that $\hgaussian_i > \hgaussian_j$ for all $i>j$ (see also \cite{wang2021tight}). Furthermore, results of the main theorems do not change by considering $\gamma(\alpha):\RR\to\RR^d$ since by our assumptions on the sparsity $s$, we have that $d-s \approx d$. Therefore, in all discussion that follows we will assume that $\gamma(\alpha):\RR\to\RR^d$.

In order to study the optimization problem in Proposition~\ref{prop:parametrization_class_general}, we make use of the following three properties of the path $\gamma(\alpha)$:

\paragraph{Concentration of $\norm{\gamma(\alpha)}_1$ and $\norm{\gamma(\alpha)}_2$.} 
As proven in Section 3.4 in \cite{wang2021tight} the path $ \gamma ( \alpha ) $ is a piecewise linear with breakpoints at $ \alpha_m $ for integers $ m = 2, \dots, d$, with
\begin{equation}
    \alpha_m = \frac{ \left( \norm{\Hvecs}_1-m \hgaussian_m \right) \norm{\hgaussian}_{\infty} } {\norm{\Hvecs}_2^2 - \norm{\Hvecs}_1 \hgaussian_m}
\end{equation}
where $\Hvecs \in \RR^{d}$ denotes vector which is equal to $\hgaussian \in \RR^{d}$ on first $m$ components and zero elsewhere. Moreover, the following concentration result holds as shown in Proposition 4 in \cite{wang2021tight}.

\begin{proposition}
\label{prop:concentration_gammas}
Let $\tm$ be given by $2\Phic(\tm) = m/d$. There exists universal positive constants $c_1,c_2,c_3,c_4>0$ such that for any $m,d$ with $m\geq c_1$ and $c_2 m\leq d\leq \exp(c_3 m^{1/5})$ we have that:
    \begin{align}
        \left|
            \frac{\norm{\gamma(\alpha_m)}_1}{\norm{\hgaussian}_\infty} - \left(\frac{1}{\tm} - \frac{2}{\tm^3}\right)
        \right|
        &\leq \frac{c_4}{\tm^5}
        \qquad \andtxt \qquad
        \left|
            \frac{\norm{\gamma(\alpha_m)}_2^2}{\norm{\hgaussian}_\infty^2} - \frac{2}{m \tm^2}
        \right|
        \leq \frac{c_4}{m \tm^4}, \label{eq:concentration}
    \end{align}
with probability at least $1-6\exp\left(-\frac{2m}{\log^5(d/m)}\right)$ over the draws of $\hgaussian$.
\end{proposition}

\paragraph{Convexity and monotonicity of  $\gamma(\alpha)$. }
According to Lemma 4 in \cite{wang2021tight} the mapping $ \alpha \mapsto \norm{ \gamma(\alpha) }_2^2 $ is convex over $ [ 1, \alpham ]$, decreasing over $[1,\alpha_{d+1/2}]$ and increasing over $[\alpha_{d+1/2},\alphamax]$ where $\alpha_{d+1/2} := \frac{\norm{\hgaussian}_1\norm{\hgaussian}_{\infty}}{\norm{\hgaussian}_2^2}$ satisfies $\alpha_d<\alpha_{d+1/2}<\alpha_{d+1}$. Furthermore the map $ \alpha \mapsto \frac{ \norm{\gamma(\alpha)}_2^2} {\norm{\gamma(\alpha)}_1^2 }  =\frac{ \norm{\gamma(\alpha)}_2^2} {\alpha^2 }$ is monotonically decreasing.

\paragraph{Inequality constraint at optimal point.} According to Claim 3 in \cite{wang2021tight} the inequality constraint in the definition of $\gamma(\alpha)$ is tight for the optimal solution, i.e.,  $\innerprod{\gamma(\alpha)}{\hgaussian} = \norm{\hgaussian}_{\infty}$.

Furthermore we define $\tm$ as solution to equation
\begin{equation}
    \label{eq:deftm}
    2\Phic(\tm) = m/d
\end{equation} for some integer $m\in[2,d]$ where $\Phic(.) = \mathbb P(Z \geq .)$ with $Z\sim \mathcal N(0,1)$ is the complementary cumulative distribution function. 
We use the following two characterizations of $\tm$:
\paragraph{Approximation of $\tm$.}
From Remark 2 in \cite{wang2021tight} there exists universal constant $\kappa$ such that, for all $m \leq d/\kappa$ holds that 
\begin{equation}
   \tm^2 = 2 \log(d/m) - \log \log (d/m) -\log(\pi) +\frac{\log\log(d/m)}{2\log(d/m)} + O\left(\frac{1}{\log(d/m)}\right).
\end{equation}

\paragraph{Upper and lower bounds of $\tm$.}
Following the same argument as in Claim 7 and Claim 9 in \cite{wang2021tight}, we can prove the following lemma:

\begin{lemma} \label{lemma_uls_ols_exp}
    Let $\mstar$ be fixed and assume $\cl \mstar \leq d$.
    Let any fixed constant $\kappa>0$ and assume that parameter $\lambda$ satisfies $0<\lambda\leq
    (\log(\cl))^{\kappa/2}$, and
    let $\ulmstar$ be the largest integer $\widehat{m}$ such that $t_{\widehat{m}}^2 \geq \tmstar^2 + \frac{\lambda}{\tmstar^{\kappa}}$.
    Then,
    \begin{align}
        \ulmstar = \mstar \exp\left( -\frac{\lambda}{2\tmstar^\kappa} \right) \left( 1+O\left(\frac{1}{\tmstar^2}\right) \right) 
        ~~~\andtxt~~~
        \left|\tulmstar^2 - \left( \tmstar^2 + \frac{\lambda}{\tmstar^\kappa} \right)\right|
        \leq O \left(
            \frac{1}{\mstar}
        \right).
    \end{align}
    Moreover, let $\olmstar$ be the smallest integer $\widehat{m}$ such that $t_{\widehat{m}}^2 \leq \tmstar^2 - \frac{\lambda}{\tmstar^\kappa}$.
    Then,
    \begin{align}
        \olmstar = \mstar \exp\left( \frac{\lambda}{2\tmstar^\kappa} \right) \left( 1+O\left(\frac{1}{\tmstar^2}\right) \right)
        ~~~\andtxt ~~~
        \hphantom{{} - }
        \left|\tolmstar^2 - \left( \tmstar^2 - \frac{\lambda}{\tmstar^\kappa} \right)\right|
        \leq O \left(
            \frac{1}{\mstar}
        \right).
    \end{align}
\end{lemma}

Furthermore, analogously as in proof of Claim 8 in \cite{wang2021tight} we get:
\begin{gather}
    \frac{\tmstar^2}{\tulmstar^2} 
    = \frac{1}{1+ \frac{\lambda}{\tmstar^{2+\kappa}} + O\left(\frac{1}{\tmstar^2 \mstar}\right)} 
    = 1 - \frac{\lambda}{\tmstar^{2+\kappa}} + O\left(\frac{1}{\tmstar^2 \mstar}\right) + O\left(\frac{\lambda^2}{\tmstar^{4+2\kappa}}\right).
\end{gather}
Similar result holds for $\tolm$.

\subsection{Concentration results}
\label{subsec:uniform_conv_tools}
\subsubsection*{Pointwise convergence}
Lemmas in this section are used in the proofs of  Propositions~\ref{prop:localization_noiseless_ssparse} and ~\ref{prop:Rbound_noisy} (localization step). We recall two standard lemmas for pointwise convergence of functions of random variables to their expectation:
\begin{lemma}[Concentration of Lipschitz functions, \cite{Ledoux1992AHS,wainwright2019high}]
\label{lemma:conc_lipschitz}
Let $X=(X_1,\dots,X_n)$ be a vector of i.i.d. $\NNN(0,1)$ random variables and let $f:\RR^n\to \RR$ be Lipschitz continous with Lipschitz constant $L$. Then
\begin{align}
    \PP \left( \abs{f(X) - \EE f(X)} \geq \epsilon \right) \leq 2\exp \left( -\frac{\epsilon^2}{2L^2} \right)
\end{align}
for any $\epsilon \geq 0$.
\end{lemma}
\begin{lemma}[Bernstein's inequality for sub-exponentials, \cite{vershynin2018high}] 
\label{lemma:bernstein_subexponential}
Let $X_1,\dots,X_n$ be i.i.d. mean-zero random variables with sub-exponential norm $\kappa = \norm{X}_{\psi_1}$. Then for any $\epsilon\geq 0$
\begin{align}
    \PP \left( \left| \frac{1}{n} \sum_{i=1}^n X_i \right| \geq \epsilon \right) \leq 2\exp \left( -c n \min\left\{ \frac{\epsilon}{\kappa}, \frac{\epsilon^2}{\kappa^2} \right\} \right)
\end{align}
for some universal constant $c>0$.
\end{lemma}

\subsubsection*{Uniform convergence}
Results from this section are used in the proofs of  Propositions~\ref{prop:BoundGamma0_class_noiseless} and ~\ref{prop:BoundGamma0_class_noisy} (uniform convergence), and more specifically, for proving Propositions~\ref{prop:unifConv_noiseless_whole} and ~\ref{prop:unifConv}. 

Let $X_1,\dots,X_n$ be real i.i.d. random variables with continuous distribution function $F$ and let $F_n$ be empirical distribution function defined by $F_n(x) = \frac{1}{n}\sum_{i=1}^n \indicator{X_i \leq x}$. Then we have:
\begin{lemma}[Dvoretzky-Kiefer-Wolfowitz inequality, \cite{dvoretzky1956asymptotic,massart}]
\label{lemma:DKW_ineq}
For any $\epsilon>0$ holds:
\begin{align}
    \PP \left( \sup_{x} |F_n(x)-F(x)| > \frac{\epsilon}{\sqrt{n}} \right)
    \leq 
    2\exp(-2\epsilon^2)
\end{align}
\end{lemma}

Before we recall a result about uniform convergence of functions from a parametrized set, let us introduce additional notation. Let $\GGG$ be a countable class of functions $g:\RR\to\RR$. For a function $g\in\GGG$ we write $Pg = \EE g(X)$, and $P_n g = \frac{1}{n}\sum_{i=1}^n g(x_i)$. Moreover, define $\norm{ P_n - P }_{\GGG} := \sup_{g \in \GGG} \abs{(P_n - P)g}$.

Let $\epsilon_1,\dots,\epsilon_n$ be independent Rademacher random variables. Define $P_n^{\epsilon} g = \frac{1}{n}\sum_{i=1}^n \epsilon_i g (x_i)$ and $ \norm{P_n^{\epsilon}}_{\GGG} = \sup_{g\in \GGG} \abs{P_n^{\epsilon}g}$.
We further recall the definition of the Orlicz norm $\orlicznorm{\cdot}$. Let $\alpha>0$ and define Orlicz function $\psi_\alpha:\RR_+\to\RR_+$ by $\psi_\alpha(x) = \exp(x^{\alpha})-1$. The Orlicz norm of the random variable $X$ is given by:
\begin{align}
\label{eq:def_orlicz_norm}
    \orlicznorm{X} := \inf \{\lambda>0: \EE \psi_\alpha(|X|/\lambda)\leq 1\}
\end{align}

For the setting defined in this section we have:
\begin{theorem}[Corollary of Theorem 4 in \cite{adamczak2008tail}]
\label{thm:unifconv_adamczak}
For any $0<t<1$, $\delta>0$, $\alpha \in (0,1] $ there exists a constant $C = C ( \alpha, t, \delta )$ such that
\begin{align}
\label{eq:tailuniformconv}
    \prob \left( \norm{ P_n - P }_{\GGG} \geq (1+t) \EE \norm{ P_n - P }_{\GGG} + \epsilon \right)
    \leq
    \exp & \left( -\frac{n\epsilon^2}{2(1+\delta)\sigma_{\GGG}^2} \right) + 3 \exp \left( - \left( \frac{\epsilon}{C \psi_{\GGG}} \right)^{\alpha} \right)
\end{align}
with 
\begin{align}
    \sigma_{\GGG}^2 =  \sup_{ g \in \GGG } \Var[g(X)]
    \quad ~~\andtxt~~ \quad
    \psi_{\GGG} = \orlicznorm{\max_{1\leq i\leq n} \sup_{g \in \GGG} \frac{1}{n} \Big| g(x_i) - \EE_X [g(X)] \Big|}
\end{align}
\end{theorem}

\section{Proof of Theorem~\ref{thm:mainl1_class_noiseless}}
\label{sec:proof_noiseless_appendix}
\label{subsec:proof_sketch_noiseless_main}

In this section we present the proof of Theorem~\ref{thm:mainl1_class_noiseless}.
By Proposition~\ref{prop:CGMT_application_classification}, in order to give bounds for prediction error, it suffices to bound $\PhiCdnuni,\PhiCdpuni$ and $\PhiCdmuni$ (defined in Section~\ref{subsec:proof_sketch_gmt}).
Furthermore, we make use of the simplifications in Proposition~\ref{prop:parametrization_class_general}, which allow us to study low dimensional stochastic optimization problems. 
In a first step (localization), we derive an upper bound for $\PhiCdnuni$:
\begin{proposition}
\label{prop:localization_noiseless_ssparse}
Let the assumptions of Theorem~\ref{thm:mainl1_class_noiseless} hold, and let $\kappam = 3(72\pi)^{-1/6}$ and $\sopt$ be the solution of equation 
\begin{align}
    \sopt = \sqrt{\frac{2}{\pi}}(72\pi)^{1/6}{(n\tsopt\norm{\wgt}_1)^{2/3}},
    \label{eq:definition_sopt}
\end{align}
where $t_{\sopt}$ is defined as in Equation~\eqref{eq:deftm} in Appendix~\ref{subsubsec:prop_gamma_alpha}.
There exists universal positive constants $c_1,c_2,c_3$ such that
\begin{align}
\label{eq_loneboundclasn_def}
    \PhiCndn 
    \leq
    \kappam
    \Big( \frac{n}{\tsopt^2} \norm{\wgt}_1 \Big)^{1/3}
    \left( 1 - \frac{2}{3}\frac{1}{\tsopt^2} + \frac{c_1}{\tsopt^4} \right)
    =:
    \loneboundclasn 
\end{align}
holds with probability at least $1-c_2\exp\left(-c_3 \frac{n^{1/3}}{\log^{10/3}(d/\sopt)} \right)$ over the draws of $\hgaussian_1,\hgaussian_2$, $\gausswone,\gausswtwo$. 
\end{proposition}
The proof of the proposition is deferred to Appendix~\ref{sec:proof_loc_noiseless}. The second step (uniform convergence) gives the following bounds on the elements of the set $\Gamma$ from Proposition~\ref{prop:parametrization_class_general}:

\begin{proposition}
\label{prop:BoundGamma0_class_noiseless}
Let the assumptions of Theorem~\ref{thm:mainl1_class_noiseless} hold. Let $\tsetNoiselessZero$ be a set of all $(\nu,b,\alpha,\etas)$ that satisfy:
\begin{gather}
    \left| \nu^2-   \frac{(288\pi)^{-1/3} n^{2/3}  } {\norm{\wgt}_1^{4/3} \log^{2/3}(d/\sopt)}  \right| 
    \lesssim \frac{n^{2/3}}{ \norm{\wgt}_1^{4/3} \log(d/\sopt)}  
    ~~\andtxt~~  
    \etas^2 \lesssim  \frac{1}{\log^{7/6}(d/\sopt)}  \\
    ~~\andtxt~~ \left| b^2\norm{\gamma(\alpha)}_2^2 - \frac{1}{3}\frac{1}{\log(d/\sopt)} \right| \lesssim \frac{1}{\log^{7/6}(d/\sopt)} 
\end{gather}
where $\sopt$ is the solution of Equation~\eqref{eq:definition_sopt}. Then there exist positive universal constants $c_1,c_2,c_3$ such that $\setNoiselessZero\subset \tsetNoiselessZero$ with probability at least $ 1 - c_1 d^{-1} - c_2 \exp\left(-c_3 \frac{n^{1/3}}{\log^4(d/\sopt )} \right)$ over over the draws of $\hgaussian_1,\hgaussian_2$, $\gausswone,\gausswtwo$. 
\end{proposition}  
The proof is deferred to Appendix~\ref{sec:proof_unicon_noiseless}. 
From the Propositions~\ref{prop:parametrization_class_general} and \ref{prop:BoundGamma0_class_noiseless} and using that $\etas^2 \geq 0$, we get the following bounds on $\PhiCndp$ and $\PhiCndm$:
\begin{align}
    \PhiCndp
    \leq
    \left[1 + \frac{\min\limits_{(b,\alpha) \in \tsetNoiselessZero} ~b^2\norm{\gamma(\alpha)}_2^2 + \min\limits_{\etas \in \tsetNoiselessZero} ~\etas^2}{\max\limits_{\nu \in \tsetNoiselessZero}~ \nu^2}  \right]^{-1/2}
    \leq
    1 - \frac{4\norm{\wgt}_1^2}{\sopt}\left( 1 - \frac{c}{\log^{1/6}(d/\sopt)} \right) 
\end{align}

\begin{align}
    \PhiCndm 
    \geq
    \left[1 + \frac{\max\limits_{(b,\alpha) \in \tsetNoiselessZero} ~b^2\norm{\gamma(\alpha)}_2^2 + \max\limits_{\etas \in \tsetNoiselessZero} ~\etas^2}{\min\limits_{\nu \in \tsetNoiselessZero}~ \nu^2}  \right]^{-1/2}
    \geq
    1 - \frac{4\norm{\wgt}_1^2}{\sopt}\left( 1 + \frac{c}{\log^{1/6}(d/\sopt)} \right),
\end{align}

and the statement of Theorem~\ref{thm:mainl1_class_noiseless} follows straightforwardly  when applying Proposition~\ref{prop:CGMT_application_classification}  and 
using that \begin{align}
        \RiskC(\what) &=\frac{1}{\pi} \arccos\left(\left\langle \frac{\what}{\norm{\what}_2}, \wgt\right\rangle\right)
        =
        \frac{1}{\pi}\sqrt{2 \left( 1-\left\langle \frac{\what}{\norm{\what}_2}, \wgt\right\rangle \right)} + O  \left( 1-\left\langle \frac{\what}{\norm{\what}_2}, \wgt\right\rangle \right)^{3/2}  \\
        \label{eq:risk_equivalence_appendix_noiseless}
    \end{align}

\subsection{Proof of Localization Proposition~\ref{prop:localization_noiseless_ssparse}}
\label{sec:proof_loc_noiseless}
Recall the upper bound of $\PhiCndn$ from Proposition~\ref{prop:parametrization_class_general}, and note that to upper bound $\PhiCndn$ it is sufficient to find a feasible point $(\tilde{\nu},\tilde{b},\tilde{\alpha})$ which satisfies the constraint, i.e. we have:
\begin{align}
    \PhiCndn \leq  \tilde{\nu}\norm{\wgt}_1  + \tilde{b} \tilde{\alpha}
    ~~~\text{if}~~~ 
    \frac{1}{n} \tilde{b}^2\norm{\hgaussian}_{\infty}^2  \geq  \fnclasn(\tilde{\nu},\tilde{b}\|\gamma(\tilde{\alpha})\|_2)
    \label{eq:loc_phi_upp_bound_one_point}
\end{align}
holds with high probability for some $\tnu>0$. We further recall that in the noiseless setting we have 
\begin{equation}
\fclasn(\nu,\eta) = \EE  \fnclasn \left( \nu, \eta \right)  = 
\EE \left( 1-  \nu \vert \Gausswone \vert -  \Gausswtwo \eta \right)_+^2.
\end{equation} 
with $f_n$ from Equation~\eqref{eq:fnclas_def_noisy_par}. Next, note that random variable $(1-\nu \abs{\Gausswone} - \eta\Gausswtwo)_+^2$ for fixed $(\nu,\eta)$ is a sub-exponential random variable. Furthermore, since $(1-\nu \abs{\Gausswone} - \eta\Gausswtwo)_+^2 \lesssim 1+\eta^2(\Gausswtwo)^2$ we obtain that the sub-exponential norm of this random variable is bounded by a constant for $\eta \leq c$. We can therefore apply Lemma~\ref{lemma:bernstein_subexponential} to show that for fixed $\nu$, $\eta \leq c$ and $\sopt$ given in Equation~\eqref{eq:definition_sopt} we have
\begin{align}
    \PP \left( |\fnclasn(\nu,\eta)-\EE\fnclasn(\nu,\eta)|\lesssim \frac{1}{\nu\tsoptn^4} \right) 
    \geq 
    1-2\exp \left( -c_1 \frac{n}{\nu^2\tsoptn^8} \right).
\end{align}
Since $\fclas$ is an infinitely differentiable function we can use the  Taylor expansion of function $\fclasn = \EE \fnclasn$ around $\eta=0$ from Equation~\eqref{eq:taylor_fclasn} which holds for $\nu$ large. Combining the last two results we obtain that with probability $1-2\exp \left( -c_1 \frac{n}{\nu^2\tsoptn^8} \right)$ holds:
\begin{align}
    \fnclasn(\nu,b\norm{\gamma(\alpha)}_2)     
    \leq
    \frac{\sqrt{2}}{3\sqrt{\pi}} \frac{1}{\nu} + \sqrt{\frac{2}{\pi}} \frac{b^2\norm{\gamma(\alpha)}_2^2}{\nu} 
    +\OOO_t + \OOO_c
    \label{eq:fn_taylor_approx_localization_proof2}
\end{align}
where $\OOO_t:=O \left( \frac{1}{\nu^3},\frac{b^4\norm{\gamma(\alpha)}_2^4}{\nu^3} \right)$ and $\OOO_c:= O \left( \frac{1}{\nu \tsopt^4} \right)$.

We claim that for our choice of the point $(\tnu,\tb,\talpha)$ we get $\OOO_{t} +\OOO_c =  \frac{1}{\tnu} O \left( \frac{1}{\tsopt^4} \right)$. Once we have established inequality~\eqref{eq:fn_taylor_approx_localization_proof2}, the claim that the point $(\tnu,\tb,\talpha)$ satisfies constraint from \eqref{eq:loc_phi_upp_bound_one_point} is implied by proving the following inequality:
\begin{align}
    \frac{1}{n} \tb^2\norm{\hgaussian}_{\infty}^2  
    \geq
    \frac{\sqrt{2}}{3\sqrt{\pi}} \frac{1}{\tnu} + \sqrt{\frac{2}{\pi}} \frac{\tb^2\norm{\gamma(\talpha)}_2^2}{\tnu} + \frac{1}{\tnu}O \left( \frac{1}{\tsopt^4} \right)
    \label{eq:loc_proof_btilde_ineq_approx}
\end{align}
Defining $\tbealpha = \tb\talpha$ and rearranging the terms in Equation~\eqref{eq:loc_proof_btilde_ineq_approx} we obtain the following lower bound for $\tbealpha$:
\begin{align}
    \tbealpha^2 
    \geq 
    \frac{n\talpha^2}{\norm{\hgaussian}_{\infty}^2} \frac{ \frac{\sqrt{2}}{3\sqrt{\pi}}\frac{1}{\tnu} \left( 1  + O \left(\frac{1}{\tsopt^4} \right)  \right) }{ 1-\sqrt{\frac{2}{\pi}} \frac{n}{\tnu} \frac{\norm{\gamma(\talpha)}_2^2}{\norm{H}_{\infty}^2}}
\end{align}
From Section~\ref{subsubsec:prop_gamma_alpha} we have that $ \gamma ( \alpha ) $ is a piecewise linear function with breakpoints at $ \alpha_m $ for $ m = 2, \dots, d $, and thus, we can optimize over integers $m$ instead of $\alpha$. Using concentration results from Proposition~\ref{prop:concentration_gammas} we get the following result:
\begin{align}
    \tbealpha^2
    \geq
    n\frac{1}{\tm^2}\left( 1-\frac{4}{\tm^2} + O \left( \frac{1}{\tm^4} \right) \right)
    \frac{\frac{\sqrt{2}}{3\sqrt{\pi}}\frac{1}{\tnu}  \left( 1 + O \left( \frac{1}{\tsopt^4} \right) \right)}{ 1 - \sqrt{\frac{2}{\pi}}\frac{n}{\tnu}\frac{2}{m\tm^2}\left( 1+O\left( \frac{1}{\tm^2} \right)\right)}  
    \label{eq:loc_noiseless_balpha_midd}
\end{align}
with probability at least $1-6\exp\left(-\frac{2m}{\log^5(d/m)} \right)$.
Similarly, as in Remark 1 in \cite{wang2021tight}, we choose $m$, which approximately minimizes the expression above, 
i.e. to maximize:
\begin{align}
    \tm^2 \left( 1 - \sqrt{\frac{2}{\pi}}\frac{n}{\tnu}\frac{2}{m\tm^2} \right) \approx 2\log \left( \frac{d}{m} \right) - 2\sqrt{\frac{2}{\pi}} \frac{n}{\tnu m} 
\end{align}
This gives $m=\sopt(\tnu) := \sqrt{\frac{2}{\pi}} \frac{n}{\tnu}$. We claim that for our choice of $\tnu$ we can set $\sopt$ to be the solution of equation $\sopt = \sqrt{\frac{2}{\pi}}(72\pi)^{1/6}{(n\tsopt\norm{\wgt}_1)^{2/3}}$ which is exactly $\sopt$ given in Equation~\eqref{eq:definition_sopt}. For such $m=\sopt$ we have from Equation~\eqref{eq:loc_noiseless_balpha_midd}:
\begin{align}
    \tbealpha^2 \geq \frac{\sqrt{2}}{3\sqrt{\pi}}\frac{n}{\tnu \tsopt^2} \left( 1 - \frac{2}{\tsopt^2} + O \left( \frac{1}{\tsopt^4}  \right) \right)
\end{align}
So we let:
\begin{align}
    \tbealpha(\tnu) := \sqrt{\frac{\sqrt{2}}{3\sqrt{\pi}}\frac{n}{\tnu \tsopt^2} \left( 1 - \frac{2}{\tsopt^2} + O \left( \frac{1}{\tsopt^4}  \right) \right) }
\end{align}
Now we choose $\tnu$ which minimizes the upper bound on $\PhiCndn$ in Equation~\eqref{eq:loc_phi_upp_bound_one_point} as follows:
\begin{align}
    \tnu := \argmin_{\nu>0} \nu \norm{\wgt}_1 + \tbealpha(\nu) = \argmin_{\nu>0} \nu\norm{\wgt}_1 + \sqrt{\frac{\sqrt{2}}{3\sqrt{\pi}}\frac{n}{\nu t_{\sopt}^2} \left( 1 - \frac{2}{\tsopt^2} + O \left( \frac{1}{\tsopt^4}    \right) \right) }
\end{align}
After minimization, we get that $\tnu$ is given by:
\begin{align}
    \tnu = (72\pi)^{-1/6} \norm{\wgt}_1^{-2/3} \left( \frac{n}{\tsopt^2} \right)^{1/3} \left( 1- \frac{2}{\tsopt^2} + O \left( \frac{1}{\tsopt^4}   \right) \right)>0
\end{align}
Note that indeed $\sopt(\tnu) = \sopt$ for this choice of $\tnu$. Returning to $\tbealpha$, we obtain the following:
\begin{align}
    \tbealpha := \tbealpha(\tnu) = 2(72\pi)^{-1/6} \norm{\wgt}_1^{1/3} \left( \frac{n}{\tsopt^2} \right)^{1/3} \left( 1 + O \left( \frac{1}{\tsopt^4}  \right) \right) 
\end{align}
Summing up the two terms, we obtain a bound from the proposition.
Also, note that for $m=\sopt$ we get:
\begin{align}
    \tb\norm{\gamma(\talpha)}_2 = \tbealpha \frac{\norm{\gamma(\talpha)}_2}{\talpha} = \sqrt{\frac{2}{3}}
    \frac{1}{\tsopt} \left( 1 + O \left( \frac{1}{\tsopt^2} \right) \right)
\end{align}
So, we have $O_t = O\left( \frac{1}{\nu^3}, \frac{\eta^4}{\nu^3} \right) = o\left( \frac{1}{\nu \tsopt^4} \right)$  as we assumed at the beginning of the proof. Thus, the point $(\tnu,\tb,\talpha)$ indeed satisfies inequality~\eqref{eq:loc_phi_upp_bound_one_point} with high probability, and we define the upper bound $\loneboundclasn:=\tnu + \tb\talpha \geq \PhiCndn$.


\subsection{Proof of Uniform Convergence Proposition~\ref{prop:BoundGamma0_class_noiseless}}
\label{sec:proof_unicon_noiseless}
For the sake of completeness, let us recall the definition of set $\setZeroUni$ from Proposition~\ref{prop:parametrization_class_general}: 
\ifaistats
\begin{align}
    \setNoiselessZero = \bigg\{ (\nu, b, \alpha, \etas) \subjto
    \etas\geq 0, b\geq0, \alpha\in[1,\alpham] ~~\andtxt~~
    \frac{1}{n} &  (2\sqrt{\smax}\etas + b\normk{\hgaussian}_{\infty})^2 \geq  \fnclasn(\nu,\sqrt{b^2\|\gamma(\alpha)\|_2^2+\etas^2})\\
    &~~\andtxt~~  \max\left\{ |\nu| \|\wgt\|_1 - \sqrt{s}\etas , 0 \right\} + b\norm{\gamma(\alpha)}_1 \leq \loneboundclasn \bigg\}
    \label{eq:Gamma0_def_class_noiseless}
\end{align}
\else
\begin{align}
    \setNoiselessZero = \bigg\{ (\nu, b, \alpha, \etas) \subjto
    &\etas\geq 0, b\geq0, \alpha\in[1,\alpham] \\ ~~\andtxt~~ & 
    \frac{1}{n} (2\sqrt{\smax}\etas + b\normk{\hgaussian}_{\infty})^2 \geq  \fnclasn(\nu,\sqrt{b^2\|\gamma(\alpha)\|_2^2+\etas^2})\\
    ~~\andtxt~~  &\max\left\{ |\nu| \|\wgt\|_1 - \sqrt{s}\etas , 0 \right\} + b\norm{\gamma(\alpha)}_1 \leq \loneboundclasn \bigg\}
    \label{eq:Gamma0_def_class_noiseless}
\end{align}
\fi
with $\loneboundclasn$ given in Proposition~\ref{prop:localization_noiseless_ssparse} and $\smax =  \Theta( n^{2/3(1-7p)} )$ for $p\in(0,\frac{1}{12})$. We further recall the notation  $\etasc = b \norm{\gamma(\alpha)}_2$ and $\eta = \sqrt{\etasc^2+\etas^2}$ from Section~\ref{subsec:path_param}.  

The proof consists of three steps where we iteratively bound the set $\Gamma$: for every bound step we use different approximation of $\fnclasn$, and based on it we develop tighter bounds for $\nu,\etasc,\etas$. Finally, the statement of the proposition follows from the last, tightest bound. We start with the following bound:
\subsubsection*{Bound 1: \texorpdfstring{$\nu\norm{\wgt}_1 \lesssim \loneboundclasn, \nu \gtrsim n^{(1+11p)/9}$}{nuomega}}
In order to derive the bounds in this section, we first need to simplify the constraints from the definition of the set $\setNoiselessZero$. First, note that we can relax the second constraint to the following two constraints: $b\alpha\leq \loneboundclasn$ and $\nu\norm{\wgt}_1 \leq \loneboundclasn+\sqrt{s}\etas$. Then, the first constraint is simplified by deriving an upper bound on the term from the LHS as follows. By using simple quadratic inequality, we have that for any $(\nu,b,\alpha,\etas)\in \setNoiselessZero$ it  holds that:
\begin{align}
    \frac{1}{n}  (2\sqrt{\smax}\etas + b\normk{H}_{\infty})^2
    \leq
    \frac{2}{n}b^2 \norm{\hgaussian}_{\infty}^2 + \frac{8}{n}\smax \etas^2
    \label{eq:proof_unicon_noiseless_bound1_fnclasn_upper_bsmax}
\end{align}
Now, recall that $d\lesssim \exp(\kappa n^{p})$ as assumed in the Theorem~\ref{thm:mainl1_class_noiseless}, and that $\tsopt^2 \gtrsim \log (d/\sopt)$ and $\alpha\geq 1$, both from Section~\ref{subsubsec:prop_gamma_alpha}. We can further bound the first term from Equation~\eqref{eq:proof_unicon_noiseless_bound1_fnclasn_upper_bsmax} with probability $\geq 1-\frac{1}{d}$ as follows:
\begin{align}
    \frac{2}{n}b^2 \norm{\hgaussian}_{\infty}^2
    \leq
    \max_{(\nu,b,\alpha,\etas)\in\setNoiselessZero} \frac{2}{n}b^2 \norm{\hgaussian}_{\infty}^2
    &\leq
    \max_{\alpha\in \setNoiselessZero} \frac{2}{n} \frac{\loneboundclasn^2}{\alpha^2} \norm{\hgaussian}_{\infty}^2
    =
    \frac{2}{n} \loneboundclasn^2 \norm{\hgaussian}_{\infty}^2\\
    &\lesssim
    \frac{1}{n} \left( \frac{n}{\tsopt^2} \norm{\wgt}_1 \right)^{2/3} \log d 
    \lesssim
    \frac{\norm{\wgt}_1^{2/3}}{n^{(1-p)/3}},
\end{align}
where we used the concentration of the maximum of i.i.d. Gaussian random variables in the second line. 
We can now define the following (larger) set:
\begin{align}
    \Gamma_1
    = \bigg\{ (\nu, b, \alpha, \etas) 
    \subjto
    \etas^2 \frac{\smax}{n} + \left(\frac{\norm{\wgt}_1^{2}}{n^{(1-p)}} \right)^{1/3} &\gtrsim
    \fnclasn (\nu,\sqrt{b^2\norm{\gamma(\alpha)}_2^2  +\etas^2}) \\
    &~~\andtxt~~ 
    b\alpha \leq \loneboundclasn 
    ~~\andtxt~~ 
    \nu \norm{\wgt}_1 \leq \loneboundclasn + \sqrt{s}\etas
    \bigg\}.
    \label{eq:Gamma_1_nu_constant_lower}
\end{align}
From our discussion above it follows that $\setNoiselessZero \subset \Gamma_1$ with high probability. The goal of this first step is to show that the bounds $\nu\norm{\wgt}_1 \lesssim \loneboundclasn$ and $ \nu \gtrsim n^{(1+11p)/9}$ hold uniformly over all $\nu\in \Gamma_1$, implying that they also hold uniformly for all $\nu\in \setNoiselessZero$ with high probability.

\textbf{Step 1.1: Upper bound $\nu\norm{\wgt}_1\lesssim \loneboundclasn$.}
In all of the Step 1.1 we assume that $(\nu,b,\alpha,\etas)\in\Gamma_1$ i.e. we bound these variables only if they are contained in $\Gamma_1$. Since by the last constraint of $\Gamma_1$ it holds that $\nu \norm{\wgt}_1 \leq \loneboundclasn+\sqrt{s}\etas$, showing that $\sqrt{s}\etas \leq c \loneboundclasn$ for some universal constant $c>0$ is sufficient to deduce that $\nu\norm{\wgt}_1 \leq (c+1) \loneboundclasn$.

Recall that $\smax = \Theta( n^{2/3(1-7p)} )$ for some universal constant $p\in(0,\frac{1}{12})$. Assume by contradiction that $\sqrt{s}\etas > c \loneboundclasn$ for any constant $c>0$. Then, we can relax the first constraint of $\Gamma_1$ as follows:
\begin{align}
    \etas^2 \frac{\smax}{n} &+ \left(\frac{\norm{\wgt}_1^{2}}{n^{(1-p)}} \right)^{1/3} \gtrsim \fnclasn(\nu,\sqrt{b^2\norm{\gamma(\alpha)}_2^2+\etas^2}) = \frac{1}{n} \sum_{i=1}^n ( 1 -  \nu |z_i^{(0)}| - z_i^{(1)} \sqrt{b^2\norm{\gamma(\alpha)}_2^2 + \etas^2} )_+^2\\
    &\geq \frac{1}{n} \sum_{i=1}^n ( 1 -  \nu |z_i^{(0)}| - z_i^{(1)} \sqrt{b^2\norm{\gamma(\alpha)}_2^2 + \etas^2} )_+^2 \indicator{\gausswtwo_i\leq - c_1}\\
    &\geq
    \frac{1}{n} \sum_{i=1}^n ( -\nu |z_i^{(0)}| + c_1 \etas )_+^2 \indicator{\gausswtwo_i\leq - c_1} \\
    &\geq
    \frac{1}{n} \sum_{i=1}^n \left( -\frac{\loneboundclasn+\sqrt{s}\etas}{\norm{\wgt}_1} |z_i^{(0)}| + c_1 \etas \right)_+^2 \indicator{\gausswtwo_i\leq - c_1}\\
    &\geq
    \frac{1}{n} \sum_{i=1}^n \left( -(1+c^{-1}) \frac{\sqrt{s}\etas}{\norm{\wgt}_1} |z_i^{(0)}| + c_1 \etas \right)_+^2 1\left\{|\gausswone_i| \leq \frac{c_1\norm{\wgt}_1}{2(1+c^{-1})\sqrt{s}} 
    \right\} \indicator{\gausswtwo_i\leq - c_1}\\
    &\gtrsim
    \etas^2 \frac{1}{n} \sum_{i=1}^n 1\left\{|\gausswone_i| \leq \frac{c_1\norm{\wgt}_1}{2(1+c^{-1})\sqrt{s}}\right\} \indicator{\gausswtwo_i\leq - c_1}
\end{align}
where in the fourth line we used that $\nu \leq \frac{\loneboundclasn+\sqrt{s}\etas}{\norm{\wgt}_1}$, and in fifth that $\loneboundclasn < c^{-1}\sqrt{s}\etas$. Next, we use that 
\begin{equation}
    \PP(|\Gausswone| \leq \frac{c_1\norm{\wgt}_1}{2(1+c^{-1})\sqrt{s}}) \gtrsim \frac{\norm{\wgt}_1}{\sqrt{s}}\gtrsim  \frac{\norm{\wgt}_1}{\sqrt{\smax}} 
\end{equation} and 
$\PP(\Gausswtwo \leq -c_1) \geq c_2$ 
and thus, from Lemma~\ref{lemma:DKW_ineq} with $\epsilon=c \sqrt{\frac{n}{\smax}}$, we obtain the following inequality holds with probability $\geq 1-\exp\left(-c_3\frac{n}{\smax}\right)$:
\begin{align}
    \etas^2 \frac{\smax}{n} + \left(\frac{\norm{\wgt}_1^{2}}{n^{(1-p)}} \right)^{1/3} \geq c_4 \etas^2 \frac{\norm{\wgt}_1}{\sqrt{\smax}}
    \label{eq:ineq_etas_rough_first_bounds_contradiction}
\end{align}
First note that $\frac{\smax}{n} = \Theta( \frac{1}{n^{(1+14p)/3}})$ and $\frac{\norm{\wgt}_1}{\sqrt{\smax}} \gtrsim \frac{1}{\sqrt{\smax}} \gtrsim \frac{1}{n^{(1-7p)/3}}$, implying that $\etas^2 \frac{\smax}{n} < \frac{c_4}{2} \etas^2 \frac{\norm{\wgt}_1}{\sqrt{\smax}} $. Thus in order for inequality~\eqref{eq:ineq_etas_rough_first_bounds_contradiction} to hold, we need that $\frac{\norm{\wgt}_1^{2/3}}{n^{(1-p)/3}} \geq \frac{c_4}{2} \etas^2 \frac{ \norm{\wgt}_1}{\sqrt{\smax}}$ or equivalently $\etas^2 \lesssim \frac{\sqrt{\smax}}{n^{(1-p)/3}\norm{\wgt}_1^{1/3}} \leq \frac{\sqrt{\smax}}{n^{(1-p)/3}} = \Theta(n^{-2p})$. But then $\sqrt{s} \etas \lesssim n^{(1-13p)/3}$, which is in contradiction with our assumption that $\sqrt{s}\etas>c\loneboundclasn$. 

Indeed, note that for any $d\leq \exp(c n^{p})$ with $0<p<\frac{1}{12}$, $\loneboundclasn$ satisfies:
\begin{align}
    n^{(1-p)/3} \lesssim (n/(\log d))^{1/3}  \lesssim \loneboundclasn\lesssim (n\sqrt{\smax})^{1/3} \lesssim n^{(4-7p)/9}
    ~~\andtxt~~
    \frac{\loneboundclasn}{\norm{\wgt}_1} \lesssim n^{1/3}
    \label{eq:loneboundclasn_upp_low_bound}
\end{align}

Hence we conclude that $\sqrt{s}\etas \leq c\loneboundclasn$, and furthermore $\nu\norm{\wgt}_1 \leq \tc \loneboundclasn$ for some universal constants $c,\tc>0$, which is exactly what we wanted to show in this step.

\textbf{Step 1.2: Lower bound $\nu  \gtrsim n^{(1+11p)/9}$.}


In order to show the lower bound $\nu \gtrsim n^{(1+11p)/9}$ we first lower bound the function $\fnclasn$ for any $\nu,\eta$ as follows:
\ifaistats
\begin{align}    
    \fnclasn (\nu,\eta)  
    =
    \frac{1}{n} \sum_{i=1}^n \possq{ 1 - \nu \abs{\gausswone_i} - \eta\gausswtwo_i} 
    \geq
    \frac{1}{n} \sum_{i=1}^n \possq{ 1 - \nu \abs{\gausswone_i} - \eta\gausswtwo_i} \indicator{\gausswtwo_i \leq 0}
    \gtrsim
    \frac{1}{n} \sum_{i=1}^{n} \possq{ 1 - \nu \abs{\gausswone_i}}
    \label{eq_etanu_to_nu_fnclansn}
\end{align}
\else
\begin{align}    
    \fnclasn (\nu,\eta)  
    =
    \frac{1}{n} \sum_{i=1}^n \possq{ 1 - \nu \abs{\gausswone_i} - \eta\gausswtwo_i} 
    &\geq
    \frac{1}{n} \sum_{i=1}^n \possq{ 1 - \nu \abs{\gausswone_i} - \eta\gausswtwo_i} \indicator{\gausswtwo_i \leq 0}\\
    &\gtrsim
    \frac{1}{n} \sum_{i=1}^{n} \possq{ 1 - \nu \abs{\gausswone_i}}
    \label{eq_etanu_to_nu_fnclansn}
\end{align}
\fi
with probability $\geq 1-\exp(-c_1 n)$ for some positive universal constant $c_1$. Combining this inequality with the first constraint of $\Gamma_1$ we have that any $(\nu,b,\alpha,\etas)\in\Gamma_1$ must satisfy with high probability that:
\begin{align}
    \frac{1}{n} \sum_{i=1}^{n} \possq{ 1 - \nu \abs{\gausswone_i}} 
    &\lesssim
    \fnclasn(\nu,\sqrt{b^2\norm{\gamma(\alpha)}_2^2+\etas^2})
    \lesssim
    \etas^2 \frac{\smax}{n} + \left(\frac{\norm{\wgt}_1^{2}}{n^{(1-p)}} \right)^{1/3}\\
    &\lesssim
    \max\left\{ \frac{1}{n^{(1+14p)/9}}, \frac{1}{n^{(1+11p)/9}} \right\}
    \lesssim
    \frac{1}{n^{(1+11p)/9}}
    \label{eq:ineq_nu_first_bound_condition}
\end{align}
where in the second line we used that $\smax\etas^2 \leq c^2\loneboundclasn^2\lesssim n^{(8-14p)/9}$ shown in the previous step and Equation~\eqref{eq:loneboundclasn_upp_low_bound}, and that $\norm{\wgt}_1^{2/3} \leq (\smax)^{1/3} =\Theta(n^{2/9(1-7p)})$. 

Now, define $F_n(\frac{1}{2\nu}) := \frac{1}{n}\sum_{i=1}^n \indicator{|\gausswone_i|\leq \frac{1}{2\nu}}$ and $F(\frac{1}{2\nu}) := \PP( |\Gausswone| \leq \frac{1}{2\nu} )=\erf(\frac{1}{2\sqrt{2}\nu})$ by definition of error function. We can further simplify inequality~\eqref{eq:ineq_nu_first_bound_condition} as follows:
\ifaistats
\begin{align}
    \frac{1}{n^{(1+11p)/9}} 
    \gtrsim 
    \frac{1}{n} \sum_{i=1}^{n} \possq{ 1 - \nu \abs{\gausswone_i}} 
    \geq
    \frac{1}{n} \sum_{i=1}^{n} (1 - \nu \abs{\gausswone_i})^2\indicator{1-2\nu\abs{\gausswone_i}} 
    \geq
    \frac{n F_n(\frac{1}{2\nu})}{n} \left(\frac{1}{2} \right)^2
    \gtrsim
    F_n \left(\frac{1}{2\nu} \right)
\end{align}
\else
\begin{align}
    \frac{1}{n^{(1+11p)/9}} 
    \gtrsim 
    \frac{1}{n} \sum_{i=1}^{n} \possq{ 1 - \nu \abs{\gausswone_i}} 
    &\geq
    \frac{1}{n} \sum_{i=1}^{n} (1 - \nu \abs{\gausswone_i})^2\indicator{1-2\nu\abs{\gausswone_i}} \\
    &\geq
    \frac{n F_n(\frac{1}{2\nu})}{n} \left(\frac{1}{2} \right)^2
    \gtrsim
    F_n \left(\frac{1}{2\nu} \right)
\end{align}
\fi
where we used that the number of activated indicators of the set $\{ \indicator{1-2\nu\abs{\gausswone_i}}\}_{i=1}^n$ is equal to $n F_n(\frac{1}{2\nu})$ and that $(1-\nu\abs{\gausswone_i})_+ \geq \frac{1}{2}$ when $1-2\nu\abs{\gausswone_i}\geq 0$. Then, according to the Dvoretzky-Kiefer-Wolfowitz inequality from Lemma~\ref{lemma:DKW_ineq} we have with probability at least $1-2\exp(-c n^{(7-22p)/9})$ that
\begin{align}
    \sup_{\nu} \left| F_n \left( \frac{1}{2\nu} \right) - F\left( \frac{1}{2\nu} \right) \right| 
    =
    \sup_{\nu} \left| F_n \left( \frac{1}{2\nu} \right) - \erf\left( \frac{1}{2\sqrt{2}\nu} \right) \right| 
    \lesssim
    \frac{1}{n^{(1+11p)/9}}
    \label{eq:dkw_ineq}
\end{align}
implying that $\sup_{\nu\in\Gamma_1} \erf\left( \frac{1}{2\sqrt{2}\nu} \right) \lesssim \frac{1}{n^{(1+11p)/9}}$ with high probability. Thus we can use the Taylor series approximation of $\erf(\cdot)$ around zero to show that $\nu \gtrsim n^{(1+11p)/9}$, as we wanted to show.

\subsubsection*{Bound 2: \texorpdfstring{$\etasc,\etas = O(1),\ \nu\norm{\wgt}_1 \geq \kappamlower \loneboundclasn$}{bound1} }
For this bound we use results from the previous steps. Restricting to the set where  $\nu \gtrsim n^{(1+11p)/9}$, and $\nu \lesssim \frac{\loneboundclasn}{\norm{\wgt}_1} \lesssim n^{1/3}$ from Equation~\eqref{eq:loneboundclasn_upp_low_bound}, we can use the lower bound from Proposition~\ref{prop:fnoiselessquadratic}:\begin{align}
    \fnclasn (\nu,\eta) \geq \kappa_1 \frac{1}{\nu} + \kappa_2 \frac{\eta^2}{\nu}
    \label{eq:proof_unicon_noiseless_fnclasn_lower_kappas}
\end{align}
which holds with probability $\geq 1-2\exp(-c_2 n^{1/3})$.

Now we further simplify the LHS of the first constraint in the definition of set $\Gamma_1$. Combining the upper bound from Equation~\eqref{eq:proof_unicon_noiseless_bound1_fnclasn_upper_bsmax} with the lower bound~\eqref{eq:proof_unicon_noiseless_fnclasn_lower_kappas}, we have
\begin{align}
    \frac{2}{n}b^2 \norm{\hgaussian}_{\infty}^2 + \frac{8}{n}\smax \etas^2
    \geq
    \fnclasn(\nu,\eta)
    \geq
    \kappa_1 \frac{1}{\nu} + \kappa_2 \frac{b^2\norm{\gamma(\alpha)}_2^2+\etas^2}{\nu} 
    \label{eq:bound2_relax_first_constraint}    
\end{align}
As before, we have $\frac{\smax}{n} = \Theta( \frac{1}{n^{(1+14p)/3}})$ from the definition of $\smax$, and, as noted above, we have that $\frac{1}{\nu} \gtrsim \frac{1}{n^{1/3}}$. Thus, for $p>0$ and $n\geq c$ we have that $\frac{8\smax}{n} \leq \frac{\kappa_2}{2\nu}$, and hence:
\begin{align}
    \frac{2}{n}b^2 \norm{\hgaussian}_{\infty}^2
    \geq
    \kappa_1 \frac{1}{\nu} + \kappa_2 \frac{b^2\norm{\gamma(\alpha)}_2^2}{\nu}
    +\etas^2 \left(\frac{\kappa_2}{\nu} - \frac{8\smax}{n} \right)
    \geq
    \kappa_1 \frac{1}{\nu} + \kappa_2 \frac{b^2\norm{\gamma(\alpha)}_2^2}{\nu}
    +\kappa_3\frac{\etas^2}{\nu}
\end{align}
where we set $\kappa_3=\frac{\kappa_2}{2}$. Since the above inequalities hold with high probability, we define a set $\Gamma_2$ as the set of all $(\nu, b, \alpha,\etas) $ that satisfy:
\begin{align}
    \frac{2}{n}b^2 \norm{\hgaussian}_{\infty}^2
    \geq
    \kappa_1 \frac{1}{\nu} + \kappa_2 \frac{b^2\norm{\gamma(\alpha)}_2^2}{\nu} +\kappa_3\frac{\etas^2}{\nu} 
    ~~\andtxt~~ 
    b\alpha \leq \loneboundclasn
    ~~\andtxt~~ 
  n^{(1+11p)/9}   \lesssim \nu  \lesssim \frac{\loneboundclasn}{\norm{\wgt}_1}
    \label{eq:Gamma_2_eta_def_upp}
\end{align}
and by the above discussion we have that with high probability,  $\setNoiselessZero \subset \Gamma_2$. Hence, by bounding variables $\nu,\eta$ from $\Gamma_2$, we will obtain valid upper bounds in $\setNoiselessZero$ as well.

\textbf{Step 2.1: Upper bound $\etasc= O(1)$.}
Recall that we use the parametrization of $\etasc$ such that $\etasc = b\norm{\gamma(\alpha)}_2$. Thus, we bound $\etasc$ as follows:
\begin{align}
    \etasc^2
    &\leq
    \max_{(\nu,b,\alpha,\etas)\in\Gamma_2} b^2\norm{\gamma(\alpha)}_2^2\\
    &\leq
    \max_{\nu,b,\alpha} \left[ b^2\norm{\gamma(\alpha)}_2^2 
    \subjto 
    \norm{\gamma(\alpha)}_2^2 \leq \frac{2}{\kappa_2 n}\nu \norm{\hgaussian}_{\infty}^2
    ~~\andtxt~~ 
    b\alpha \leq \loneboundclasn
    ~~\andtxt~~ 
    \nu \norm{\wgt}_1 \leq c\loneboundclasn
    \right]\\
    &=
    \max_{\alpha} \left[ \loneboundclasn^2\frac{\norm{\gamma(\alpha)}_2^2}{\alpha^2} 
    \subjto 
    \norm{\gamma(\alpha)}_2^2 \leq \frac{2c}{\kappa_2 n} \frac{\loneboundclasn}{\norm{\wgt}_1} \norm{\hgaussian}_{\infty}^2
    \right]
    \label{eq:eta_upper_bound_step21}.
\end{align}
 As we mentioned in Section~\ref{subsubsec:prop_gamma_alpha}, the function $\frac{\norm{\gamma(\alpha)}_2^2}{\alpha^2}$ is monotonically decreasing function in $\alpha$, whereas $\norm{\gamma(\alpha)}_2^2$ is a convex function. Thus, similarly to the proofs  in \cite{wang2021tight}, it is sufficient to find $\alphauls < \alphasopt$, such that
\begin{align}
     \frac{\norm{\gamma(\alphauls)}_2^2}{\norm{\hgaussian}_{\infty}^2} > \frac{2c}{\kappa_2 n} \frac{\loneboundclasn}{\norm{\wgt}_1} 
\end{align}
to obtain an upper bound on $\frac{\norm{\gamma(\alpha)}_2^2}{\alpha^2}$ (where we implicitly make use of the fact that the set $\Gamma$ contains the point $(\tnu,\tb, \talpha = \alphasopt, 0)$ from Proposition~\ref{prop:BoundGamma0_class_noiseless}). Using the concentration results from Section~\ref{subsubsec:prop_gamma_alpha} we can rewrite the above inequality as follows:
\begin{align}
    \frac{2}{\ulm \tulm^2} \left( 1+O \left( \frac{1}{\tulm^2} \right) \right) > \frac{2c}{\kappa_2 n} \frac{\kappam}{\norm{\wgt}_1^{2/3}} \left( \frac{n}{\tsopt^2} \right)^{1/3} \left( 1 + O \left( \frac{1}{\tsopt^2} \right) \right)
\end{align}
After recalling that $\tm^2 = 2\log(d/m) + O(\log\log(d/m))$ from Section~\ref{subsubsec:prop_gamma_alpha}, it is straightforward to show that we can choose $\ulm = \lambda \left( \frac{n\norm{\wgt}_1}{\tsopt^2} \right)^{2/3}$ with sufficiently small universal constant $\lambda>0$. We finish this step by substituting this choice of $\ulm$ into the upper bound from Equation~\eqref{eq:eta_upper_bound_step21} to get:
\begin{align}
    \etasc^2 
    \leq
    \loneboundclasn^2 \frac{\norm{\gamma(\alphauls)}_2^2}{\alphauls^2}
    = \kappam^2 \left( \frac{n\norm{\wgt}_1}{\tsopt^2} \right)^{2/3} \left( 1 + O \left( \frac{1}{\tsopt^2} \right) \right) 
    \frac{2}{\ulm} \left( 1 + O \left( \frac{1}{\tulm^2} \right) \right) =: B_{\etasc}^2 = O(1) 
\end{align}
\noindent\textbf{Step 2.2: Upper bound $\etas=O(1)$.}
Similarly as in the previous step we use the relaxations of the constraints from the set $\Gamma_2$ to bound $\etas$ as follows:
\begin{align}
    \etas^2 
    &\leq
    \max_{(\nu,b,\alpha,\etas)\in \Gamma_2} \etas^2\\
    &\leq 
    \max_{\nu,b,\alpha,\etas} 
    \bigg[\etas^2\subjto 
    \etas^2 \leq \frac{2}{\kappa_3 n}\nu b^2\norm{\hgaussian}_{\infty}^2
    ~~\andtxt~~ 
    \norm{\gamma(\alpha)}_2^2 \leq \frac{2}{\kappa_2 n}\nu \norm{\hgaussian}_{\infty}^2\\
    &
    \hphantom{{} \leq 
    \max_{\nu,b,\alpha,\etas} 
    \etas^2\subjto 
    \etas^2 \leq \frac{2}{\kappa_3 n}\nu b^2\norm{\hgaussian}_{\infty}^2\nu b^2\norm{\hgaussian}_{\infty}^2}
    ~~\andtxt~~ 
    b\alpha \leq \loneboundclasn
    ~~\andtxt~~ 
    \nu \norm{\wgt}_1 \leq c\loneboundclasn\bigg]\\
    &\leq
    \max_{\nu,b,\alpha} \left[ \frac{2}{\kappa_3 n}\nu b^2\norm{\hgaussian}_{\infty}^2 \subjto     \norm{\gamma(\alpha)}_2^2 \leq \frac{2}{\kappa_2 n}\nu \norm{\hgaussian}_{\infty}^2
    ~~\andtxt~~ 
    b \leq \frac{\loneboundclasn}{\alpha}
    ~~\andtxt~~ 
    \nu  \leq c\frac{\loneboundclasn}{\norm{\wgt}_1}
    \right]\\
    &\leq 
    \frac{2}{\kappa_3 n}
    \frac{c\loneboundclasn}{\norm{\wgt}_1}  \loneboundclasn^2 \norm{\hgaussian}_{\infty}^2
    \max_{\alpha}
    \left[ \frac{1}{\alpha^2} \subjto 
     \norm{\gamma(\alpha)}_2^2 \leq \frac{2c}{\kappa_2 n}\frac{\loneboundclasn}{\norm{\wgt}_1} \norm{\hgaussian}_{\infty}^2\right]
\end{align}
Now note that $\frac{1}{\alpha^2}$ is monotonically decreasing function, while the last constraint is identical to constraint from Equation~\eqref{eq:eta_upper_bound_step21}. Thus using exactly the same arguments as in the previous step we upper bound $\etas^2$ as follows:
\begin{align}
    \etas^2 
    \leq
    \frac{2c}{\kappa_3 n}
    \frac{\loneboundclasn^3}{\norm{\wgt}_1} \frac{\norm{\hgaussian}_{\infty}^2}{\alphauls^2}
    =
    \frac{2c}{\kappa_3 n} \kappam^3  \frac{n}{\tsopt^2} \tulm^2  ( 1 + O \left(\frac{1}{\tsopt^2}, \frac{1}{\tulm^2}\right))
    =: B_{\etas}^2 = O(1)
\end{align}
where we again used concentration results from Proposition~\ref{prop:concentration_gammas}, and approximation $\tm^2 = 2\log(d/m) + O(\log\log (d/m))$ from Section~\ref{subsubsec:prop_gamma_alpha}.

\noindent\textbf{Step 2.3: Lower bound $\nu\norm{\wgt}_1 \geq \kappamlower \loneboundclasn$.}
This bound follows the same reasoning as the previous two steps. Namely, we find a lower bound on $\nu$ as follows:
\begin{align}
    \nu
    &\geq
    \min_{(\nu,b,\alpha,\etas)\in \Gamma_2} \nu\\
    &\geq
    \min_{\nu,b,\alpha} \left[\nu \subjto 
    \nu \geq \frac{\kappa_1 }{2} \frac{n}{b^2\norm{\hgaussian}_{\infty}^2}
    ~~\andtxt~~ 
    \norm{\gamma(\alpha)}_2^2 \leq \frac{2}{\kappa_2 n}\nu \norm{\hgaussian}_{\infty}^2
    ~~\andtxt~~ 
    b\alpha \leq \loneboundclasn
    ~~\andtxt~~ 
    \nu \norm{\wgt}_1 \leq c\loneboundclasn\right]\\
    &\geq
    \min_{\nu,b,\alpha} \left[ \frac{\kappa_1 }{2} \frac{n}{b^2\norm{\hgaussian}_{\infty}^2} \subjto     
    \norm{\gamma(\alpha)}_2^2 \leq \frac{2}{\kappa_2 n}\nu \norm{\hgaussian}_{\infty}^2
    ~~\andtxt~~ 
    b\leq \frac{\loneboundclasn}{\alpha}
    ~~\andtxt~~ 
    \nu \norm{\wgt}_1 \leq c\loneboundclasn\right]\\
    &=
    \frac{\kappa_1 n}{2\loneboundclasn^2\norm{\hgaussian}_{\infty}^2}  \min_{\alpha} \left[ \alpha^2 \subjto \norm{\gamma(\alpha)}_2^2 \leq \frac{2c}{\kappa_2 n} \frac{\loneboundclasn}{\norm{\wgt}_1} \norm{\hgaussian}_{\infty}^2 \right]
    \label{eq:nu_lower_bound_step22}
\end{align}
Similarly as in the previous two steps, since $\alpha^2$ is monotonically increasing function, the minimum is lower bounded by $\alpha^2\geq \alphauls^2$ and after substitution of $\ulm$ as defined above, we have:
\begin{align}
    \nu \geq  \frac{\kappa_1 n}{2\loneboundclasn^2} \frac{\alphauls^2}{\norm{\hgaussian}_{\infty}^2} 
    =
    \frac{\kappa_1}{2\kappam^2} \norm{\wgt}_1^{-2/3} \left( \frac{n}{\tsopt^2} \right)^{1/3} \frac{\tsopt^2}{\tulm^2} (1+O \left( \frac{1}{\tsopt^2} \right) ) =:  \kappamlower \frac{\loneboundclasn}{\norm{\wgt}_1}
\end{align}
where once again we applied Proposition~\ref{prop:concentration_gammas}, and used that $\tm^2 = 2\log (d/m) + O (\log \log (d/m) ) $ from Section~\ref{subsubsec:prop_gamma_alpha}. After noting that we have shown $\nu\norm{\wgt}_1 \geq \kappamlower \loneboundclasn$ with high probability, we conclude this part of the proof.

\subsubsection*{Bound 3: Tight bounds}
From Step 2.2 in the previous bound we have that $\etas = O(1)$ and thus $\sqrt{s}\etas \leq \sqrt{\smax}{\etas} = O( n^{(1-7p)/3})$. Combining this bound with the lower bound $\loneboundclasn \gtrsim n^{(1-p)/3}$ from Equation~\eqref{eq:loneboundclasn_upp_low_bound}, we obtain that:
\begin{align}
    \nu\norm{\wgt}_1 \leq \loneboundclasn + \sqrt{s}\etas \leq \loneboundclasn \left(1+\frac{c_1}{n^{2p}} \right)  \leq    \kappam
    \Big( \frac{n}{\tsopt^2} \norm{\wgt}_1 \Big)^{1/3}
    \left( 1 - \frac{2}{3}\frac{1}{\tsopt^2} + \frac{c_2}{\tsopt^4} \right) =: \loneboundclasnt
    \label{eq:def_Mtilde_noiseless}
\end{align}
for some fixed universal constant $c_1,c_2>0$. Moreover, in Step 2.3 of the previous bound we have shown that $\nu\norm{\wgt}_1 \geq \kappamlower \loneboundclasn$, and thus $\nu\norm{\wgt}_1 \geq \tkappamlower \loneboundclasnt$ for some $0<\tkappamlower \leq \kappamlower$. Combining both results, we have that $\nu \norm{\wgt}_1 \in [\tkappamlower,1]\loneboundclasnt$.  


Now we show how we can relax and simplify the first constraint of the set $\setNoiselessZero$. Recall Equation~\eqref{eq:bound2_relax_first_constraint} and note that it implies that $\frac{2}{n}b^2 \norm{\hgaussian}_{\infty}^2 + \frac{8}{n}\smax \etas^2\geq \kappa_1 \frac{1}{\nu}$. Moreover, since $\nu\norm{\wgt}_1\leq \loneboundclasnt$, and $\etas \leq B_{\etas}$ from Step 2.2, we have:
\begin{align}
    \frac{1}{n}b^2\norm{\hgaussian}_{\infty}^2 \geq \frac{\kappa_1}{2}\frac{1}{\nu} - \frac{4\smax}{n}B_{\etas}^2 \geq \frac{\kappa_1}{2} \frac{\norm{\wgt}_1}{\loneboundclasnt} -4B_{\etas}^2 \frac{\smax}{n} \gtrsim  \frac{1}{n^{1/3}} - \frac{1}{n^{1/3(1+14p)}} \gtrsim \frac{1}{n^{1/3}}
\end{align}
for $n$ large enough, since $\smax = \Theta\left(n^{2/3(1-7p)}\right)$ and $\frac{\norm{\wgt}_1}{\loneboundclasnt} \geq \frac{1}{2} \frac{\norm{\wgt}_1}{\loneboundclasn} \gtrsim \frac{1}{n^{1/3}}$ from inequality~\eqref{eq:loneboundclasn_upp_low_bound}. Thus, using this lower bound on $b\norm{\hgaussian}_{\infty}$ and upper bound $\etas\leq B_{\etas}$ we have:
\begin{align}
    \frac{1}{n}  (2\sqrt{\smax}\etas + b\norm{\hgaussian}_{\infty})^2 
    &=
    \frac{1}{n}b^2\norm{\hgaussian}_{\infty}^2 \left( 1 + \frac{2\sqrt{\smax}\etas}{b\norm{\hgaussian}_{\infty}} \right)^2
    \leq
    \frac{1}{n}b^2\norm{\hgaussian}_{\infty}^2 \left( 1 + \OOO_b \right)^2
\end{align}
where we defined $\OOO_b =  \frac{c}{n^{7p/3}} $ for some universal constant $c>0$. This finishes our relaxation of the LHS of the first constraint from definition of $\setNoiselessZero$.

For the RHS of this constraint, we can apply Corollary~\ref{corr:fnoiseless_taylor} with $\epsilon \asymp \frac{1}{n^{1/3}\tsopt^4}$ to obtain that the inequality
\begin{align}
    \fnclasn (\nu,\eta) \geq \frac{\sqrt{2}}{3\sqrt{\pi}} \frac{1}{\nu} + \sqrt{\frac{2}{\pi}} \frac{\eta^2}{\nu} - \epsilon
\end{align}
holds with probability at least $1-c_1\exp\left(-c_2 \frac{n^{1/3}}{\tsopt^8}\right)$.

Now we use the derived relaxations of the first constraint of $\setNoiselessZero$ to define a new set $\Gamma_3$:
\begin{align}
    \Gamma_3 
    = \bigg\{ (\nu, b, \alpha, \etas) 
    \subjto&
    \frac{1}{n}  b^2\norm{\hgaussian}_{\infty}^2 (1+\OOO_b) 
    \geq
    \frac{\sqrt{2}}{3\sqrt{\pi}} \frac{1}{\nu}  +  \sqrt{\frac{2}{\pi}} \frac{b^2\norm{\gamma(\alpha)}_2^2+\etas^2}{\nu} - \epsilon\\
    ~~\andtxt~~ 
    &b\alpha + \nu\norm{\wgt}_1 \leq \loneboundclasnt 
    ~~\andtxt~~ 
    \nu\norm{\wgt}_1 \in [\tkappamlower, 1] \loneboundclasnt~~\andtxt~~ b \|\gamma(\alpha)\|_2 + \etas \lesssim 1
    \bigg\}.
    \label{eq:Gamma_3_eta_lower_def}
\end{align}
Again, we have that with high probability $\setNoiselessZero \subset \Gamma_3$ and in the following four steps we bound variables $\alpha,\nu,\etasc,\etas$ such that $(\nu,b,\alpha,\etasc)\in\Gamma_3$. Furthermore, in the following steps we will use multiple times the fact that:
\begin{align}
    \frac{1}{\tsopt^4}  \gtrsim \frac{1}{\log^2 (d/\sopt) } \gtrsim \frac{1}{\log^2 d} \gtrsim \frac{1}{n^{2p}}
\end{align}
which follows from characterization of $\tm^2$ from Section~\ref{subsubsec:prop_gamma_alpha} and our assumption that $d\leq \exp(c n^{p})$.

In order to derive tight bounds on $\nu,\etasc,\etas$ in Steps 3.3, 3.4. and 3.5, respectively, we first need to show an upper and lower bound on $\alpha$ in Steps 3.1 and 3.2, respectively.
\subsubsection*{Step 3.1: Upper bound $\alpha \leq \alpha_{\lambda \soptn} (\lambda >1) $.}
We upper bound $\alpha$ uniformly over $\Gamma_3$ as follows:
\begin{align}
    &\alpha^2 
    \leq 
    \max_{(\nu,b,\alpha,\etas)\in\Gamma_3}\alpha^2\\
    &\leq
    \max_{\nu, b,\alpha }\left[ \alpha^2 \subjto \frac{1}{n}  b^2\norm{\hgaussian}_{\infty}^2 (1+\OOO_b) 
    \geq
    \frac{\sqrt{2}}{3\sqrt{\pi}} \frac{1}{\nu}  -\epsilon
    ~~\andtxt~~ 
    b\alpha + \nu\norm{\wgt}_1 \leq \loneboundclasnt 
    ~~\andtxt~~ 
    \nu\norm{\wgt}_1 \in [\tkappamlower, 1] \loneboundclasnt
    \right]\\
    &\leq
     \max_{\nu,\alpha}  \left[ \alpha^2  \subjto
    \frac{1}{n} \left( \frac{\loneboundclasnt-\nu\norm{\wgt}_1}{\alpha} \right)^2 \norm{\hgaussian}_{\infty}^2(1+\OOO_b)
    \geq
    \frac{\sqrt{2}}{3\sqrt{\pi}} \frac{1}{\nu}  -\epsilon
    ~~\andtxt~~ 
    \nu\norm{\wgt}_1 \in [\tkappamlower, 1] \loneboundclasnt
    \right]\\
    &\leq 
    \max_{\nu,\alpha}
    \left[
    \alpha^2
    \subjto
    \frac{\alpha^2}{\norm{\hgaussian}_{\infty}^2} \left(1- \frac{3\sqrt{\pi}}{\sqrt{2}}\epsilon\nu \right) \leq \frac{3\sqrt{\pi}}{\sqrt{2}}\frac{1}{n}\nu(\loneboundclasnt-\nu\norm{\wgt}_1)^2(1+\OOO_b) 
    ~~\andtxt~~ 
    \nu \norm{\wgt}_1 \in [\tkappamlower, 1] \loneboundclasnt
    \right]\\
    &\leq
    \max_{\alpha}
    \left[
    \alpha^2
    \subjto
    \frac{\alpha^2}{\norm{\hgaussian}_{\infty}^2} \leq \frac{12\sqrt{\pi}}{27\sqrt{2}} \frac{1}{n} \frac{\loneboundclasnt^3}{\norm{\wgt}_1} (1+O\left( \frac{1}{\tsopt^4} \right))
    \right]
    \label{eq:proof_unicon_lower_bound_etasc}
\end{align}
where in the second line we used the second constraint to upper bound $b$, and in the last line we used that $(1+\OOO_b)(1- \frac{3\sqrt{\pi}}{\sqrt{2}}\epsilon\nu)^{-1} \leq 1+O(\frac{1}{\tsopt^4})$ and that the function $\nu (\loneboundclasnt -\nu\norm{\wgt}_1)^2$ under the constraint $\nu \norm{\wgt}_1 \in [\tkappamlower, 1] \loneboundclasnt$ is maximized for $\nu\norm{\wgt}_1 = \loneboundclasnt/3$. Furthermore, note that $1/3\in[\tkappamlower,1]$ since $\setNoiselessZero\subset\Gamma_3$ and point $\nu = \frac{\loneboundclasn}{3\norm{\wgt}_1}\in \setNoiselessZero$ by arguments from proof of the localization proposition~\ref{prop:localization_noiseless_ssparse}.

Similarly as in the previous bounds, we use that $\alpha^2$ is monotonically increasing, convex function, and thus in order to lower bound $\norm{\gamma(\alpha)}_2^2$, it is sufficient to find a point $\alphaols$ such that $\alphaols \geq \alpha_m$ for which constraint from Equation~\eqref{eq:proof_unicon_lower_bound_etasc} does not hold. Now, using concentration result from Proposition~\ref{prop:concentration_gammas} and definition of $\loneboundclasnt$, we have that $\alpha=\alpha_{\olm}$ does not satisfy the constraint if:
\begin{align}
    \frac{1}{\tolm^2} \left( 1 - \frac{4}{\tolm^2} + O \left( \frac{1}{\tolm^4} \right)\right) 
    >
    \frac{1}{\tsopt^2} \left( 1 - \frac{2}{\tsopt^2} + O \left( \frac{1}{\tsopt^4} \right) \right)
\end{align}
We can choose $\olm = \lambda \sopt$ for some constant $\lambda>1$ since using characterization of $\tm$ from Section~\ref{subsubsec:prop_gamma_alpha} we have:
\begin{align}
    \frac{\tsopt^2}{\tolm^2} = 1 + \frac{2\log\lambda}{\tsopt^2} + O \left( \frac{1}{\tsopt^4} \right)
\end{align}
Thus we finally obtain that $\alpha \leq \alpha_{\olm}$, as we wanted to show.


\subsubsection*{Step 3.2: Lower bound $\alpha \geq \alpha_{\lambda \soptn} (\lambda \in(0,1)) $}
The bound in this step is derived similarly to the bound from Step 3.1. However in this step we cannot neglect the term $\sqrt{\frac{2}{\pi}} \frac{b^2\norm{\gamma(\alpha)}_2^2}{\nu}$ from the first constraint of $\Gamma_3$, as we did in the previous step. For the sake of shorter equations, we will write only relaxations of constraints that $\alpha$ needs to satisfy and skip writing that we minimize over $\alpha^2$ like we did previously.

We start by rewriting and relaxing the first constraint from $\Gamma_3$ as follows:
\begin{align}
    b^2 \left( \frac{\norm{\hgaussian}_{\infty}^2}{n}(1+\OOO_b) - \sqrt{\frac{2}{\pi}} \frac{\norm{\gamma(\alpha)}_2^2}{\nu} \right) 
    &\geq
    \frac{\sqrt{2}}{3\sqrt{\pi}} \frac{1}{\nu} +\sqrt{\frac{2}{\pi}} \frac{\etas^2}{\nu} -\epstight \\
    &\geq
    \frac{\sqrt{2}}{3\sqrt{\pi}} \frac{1}{\nu} -\epstight 
    \geq
    \frac{\sqrt{2}}{3\sqrt{\pi}} \frac{1}{\nu} (1-\OOO\left(\frac{1}{\tsopt^4} \right))
    \label{eq:lower_bound_alpha_rhs_ineq}
\end{align}
where we used that $\epsilon \nu = O(\frac{1}{\tsopt^4})$. Now, using the second constraint of $\Gamma_3$, we can further relax the LHS of the previous inequality as follows:
\begin{align}
    b^2 \left( \frac{\norm{\hgaussian}_{\infty}^2}{n}(1+\OOO_b) - \sqrt{\frac{2}{\pi}} \frac{\norm{\gamma(\alpha)}_2^2}{\nu} \right) 
    \leq
    \frac{(\loneboundclasnt-\nu\norm{\wgt}_1)^2}{\alpha^2} \left( \frac{\norm{\hgaussian}_{\infty}^2}{n}(1+\OOO_b) - \sqrt{\frac{2}{\pi}} \frac{\norm{\gamma(\alpha)}_2^2}{\nu} \right)
    \label{eq:lower_bound_alpha_lhs_ineq}
\end{align}
Combining inequalities~\eqref{eq:lower_bound_alpha_rhs_ineq} and~\eqref{eq:lower_bound_alpha_lhs_ineq}, and plugging in $\nu\norm{\wgt}_1 = \kappa \loneboundclasnt$ for $\kappa \in [\tkappamlower,1]$ yields:
\begin{align}
    \frac{\sqrt{2}}{3\sqrt{\pi}} \frac{\norm{\wgt}_1}{\kappa\loneboundclasnt} (1-O\left(\frac{1}{\tsopt^4} \right)) 
    \leq
    \frac{\loneboundclasnt^2 (1-\kappa)^2}{\alpha^2} \left( \frac{\norm{\hgaussian}_{\infty}^2}{n}(1+\OOO_b) - \sqrt{\frac{2}{\pi}} \frac{\norm{\gamma(\alpha)}_2^2 \norm{\wgt}_1}{\kappa\loneboundclasnt} \right)
\end{align}
After multiplying the previous inequality by $\frac{\kappa\loneboundclasnt}{\norm{\wgt}_1} \frac{\alpha^2}{\loneboundclasnt^2 (1-\kappa)^2}$ and rearranging terms, we obtain: 
\begin{align}
\label{eq:lower_bound_alpha_fcn_kappa}
    \sqrt{\frac{2}{\pi}} \norm{\gamma(\alpha)}_2^2 
    \leq
    \frac{\norm{\hgaussian}_{\infty}^2}{n} \frac{\kappa \loneboundclasnt}{\norm{\wgt}_1}(1+\OOO_b) - \frac{\sqrt{2}}{3\sqrt{\pi}} \frac{\alpha^2}{\loneboundclasnt^2 (1-\kappa)^2} (1-O\left(\frac{1}{\tsopt^4} \right))
\end{align}
Note that only the right hand side depends on $\nu$ (and thus on $\kappa$). Hence maximizing over $\kappa$ the right hand side 
we obtain:
\begin{align}
    \kappa = 1-\left( \frac{2\sqrt{2}}{3\sqrt{\pi}} \frac{n\alpha^2 \norm{\wgt}_1}{\norm{\hgaussian}_{\infty}^2 \loneboundclasnt^3} (1 - O \left( \frac{1}{\tsopt^4} \right) ) \right)^{1/3}
    \geq
    1-\left( \frac{2\sqrt{2}}{3\sqrt{\pi}} \frac{n{\alpha_{\olm}}^2 \norm{\wgt}_1}{\norm{\hgaussian}_{\infty}^2 \loneboundclasnt^3} (1 - O \left( \frac{1}{\tsopt^4} \right) ) \right)^{1/3} > \frac{1}{3}
\end{align}
where we used that $\alpha \geq \alpha_{\olm}$ derived in the previous step. Moreover, note that $\kappa\in[\tkappamlower,1]\loneboundclasnt$, by the proof of our localization proposition. Substituting this $\kappa$ into \eqref{eq:lower_bound_alpha_fcn_kappa} we get the following inequality:
\begin{align}
    \frac{\alpha^{2/3}}{\norm{\hgaussian}_{\infty}^{2/3}} \left( \frac{9\sqrt{2}}{4\sqrt{\pi}} \right)^{1/3}(1-\OOO\left(\frac{1}{\tsopt^4}\right))
    +
    n^{2/3} \sqrt{\frac{2}{\pi}} \frac{\norm{\gamma(\alpha)}_2^2}{\norm{\hgaussian}_{\infty}^2} \norm{\wgt}_1^{2/3} 
    \leq
    \frac{1}{n^{1/3}} \frac{\loneboundclasnt}{\norm{\wgt}_1^{1/3}}(1+\OOO_b)
\end{align}
Now we further relax the constraint by raising the previous inequality to the third power and keeping only the first two terms to get:
\begin{align}
    \frac{\alpha^{2}}{\norm{\hgaussian}_{\infty}^2} \frac{9\sqrt{2}}{4\sqrt{\pi}}  + 3 \sqrt{\frac{2}{\pi}} \frac{\norm{\gamma(\alpha)}_2^2}{\norm{\hgaussian}_{\infty}^2}\norm{\wgt}_1^{2/3} n^{2/3} \frac{\alpha^{4/3}}{\norm{\hgaussian}_{\infty}^{4/3}} \left( \frac{9\sqrt{2}}{4\sqrt{\pi}} \right)^{2/3}
    \leq
    \frac{1}{n} \frac{\loneboundclasnt^3}{\norm{\wgt}_1}   ( 1 + O\left(\frac{1}{\tsopt^4} \right)  )
\end{align}
We can further relax this constraint by using the that $\alpha \geq \alphauls$ with $\ulm = \lambda \left( \frac{n\norm{\wgt}_1}{\tsopt^2} \right)^{2/3}$ as shown in the Bound 2. Then, we have substitute this value of $\alpha$ only in the second term as follows:
\begin{align}
    \frac{\alpha^{2}}{\norm{\hgaussian}_{\infty}^2} \frac{9\sqrt{2}}{4\sqrt{\pi}}  + 3 \sqrt{\frac{2}{\pi}} \frac{\norm{\gamma(\alpha)}_2^2}{\norm{\hgaussian}_{\infty}^2} \norm{\wgt}_1^{2/3} n^{2/3} \frac{\alphauls^{4/3}}{\norm{\hgaussian}_{\infty}^{4/3}} \left( \frac{9\sqrt{2}}{4\sqrt{\pi}} \right)^{2/3}
    \leq
    \frac{1}{n} \frac{\loneboundclasnt^3}{\norm{\wgt}_1}  (1+O\left(\frac{1}{\tsopt^4}\right))  
\end{align}
Now, note that the term on the left hand side is a sum of two convex functions in $\alpha$ and thus is convex. Similarly, as before, we look for $\alphauuls < \alpha_m$ such that the previous inequality is not satisfied. Using concentration results from Proposition~\ref{prop:concentration_gammas}, we get:
\begin{align}
     3&\sqrt{\frac{2}{\pi}} \left( \frac{9\sqrt{2}}{4\sqrt{\pi}} \right)^{2/3} \norm{\wgt}_1^{2/3} \frac{2 n^{2/3}}{\uulm \tuulm^2} \left(1+ O \left( \frac{1}{\tuulm^2} \right) \right) \frac{1}{\tulm^{4/3}} \left( 1- \frac{8}{3}\frac{1}{\tulm^2} + O \left( \frac{1}{\tulm^4} \right) \right) \\
    &+\frac{9\sqrt{2}}{4\sqrt{\pi}} \frac{1}{\tuulm^2} \left( 1 - \frac{4}{\tuulm^2} + O \left( \frac{1}{\tuulm^4} \right) \right) >
    \frac{9\sqrt{2}}{4\sqrt{\pi}} \frac{1}{\tsopt^2} \left( 1- \frac{2}{\tsopt^2} + O \left( \frac{1}{\tsopt^4} \right) \right)
\end{align}
and we can choose $\uulm = \lambda \sopt$ with $\lambda\in(0,1)$. This gives us a lower bound on $\alpha$ which is tight enough for obtaining bounds on $\nu$ with right multiplicative constant.

\subsubsection*{Step 3.3: Tight bounds on $\nu$}
Now consider a set $\Gamma_3^{\nu}:= \Gamma_3 \cap \{(\nu,b,\alpha,\etas)\subjto \alpha\geq\alphauuls\}$ with $\uulm$ given in the previous step.
Furthermore, from the arguments in the previous step it holds that $\setNoiselessZero \subset \Gamma_3^{\nu}$ with high probability. 

Now, similarly to Step 3.1 we can relax first constraint of $\Gamma_3$ to $\frac{1}{n}b^2\norm{\hgaussian}_{\infty}^2 \geq \frac{\sqrt{2}}{3\sqrt{\pi}} \frac{1}{\nu}(1-O\left(\frac{1}{\tsopt^4}\right))$. Combining this lower bound on $b$ with the second constraint of $\Gamma_3$ we have:
\begin{align}
    \loneboundclasnt - \nu\norm{\wgt}_1 \geq b\alpha \geq 
    \sqrt{\frac{\sqrt{2}}{3\sqrt{\pi}}}  \frac{\sqrt{n}}{\norm{\hgaussian}_{\infty}} \frac{\alpha}{\sqrt{\nu}}(1-O\left(\frac{1}{\tsopt^4}\right))
\end{align}
Rearranging the terms we obtain that for any $(\nu,b,\alpha,\etas)\in \Gamma_3^{\nu}$ must hold that:
\begin{align}
    0 &\geq 
    \nu^{3/2}\norm{\wgt}_1 - \loneboundclasnt \nu^{1/2} + \sqrt{\frac{\sqrt{2}}{3\sqrt{\pi}}} \sqrt{n} \frac{\alpha}{\norm{\hgaussian}_{\infty}}(1-O\left(\frac{1}{\tsopt^4}\right)) \\
    &\geq
    \nu^{3/2}\norm{\wgt}_1 - \loneboundclasnt \nu^{1/2} + \sqrt{\frac{\sqrt{2}}{3\sqrt{\pi}}} \sqrt{n} \frac{\alphauuls}{\norm{\hgaussian}_{\infty}}(1-O\left(\frac{1}{\tsopt^4}\right)) 
    \label{eq:constraint_nu_tight_noiseless}
\end{align}
where we used that $\alpha \geq \alphauuls$
Thus, the constraint~\ref{eq:constraint_nu_tight_noiseless} must hold uniformly for all $\nu\in\Gamma_3^{\nu}$. Setting $\nu\norm{
\wgt}_1 = \kappa^2\loneboundclasnt$ with $\kappa^2 \in [\tkappamlower,1]$ we obtain the following constraint on $\kappa$:
\begin{align}
    \kappa^3 - \kappa + \sqrt{\frac{\sqrt{2}}{3\sqrt{\pi}} \frac{n\norm{\wgt}_1}{\loneboundclasnt^{3}}} \frac{\alphauuls}{\norm{\hgaussian}_{\infty}}(1-O\left(\frac{1}{\tsopt^4}\right)) \leq 0
\end{align}
Using definition of $\loneboundclasnt$ from Equation~\eqref{eq:def_Mtilde_noiseless} and concentration inequality from Proposition~\ref{prop:concentration_gammas} we obtain
\begin{align}
    \kappa^3 - \kappa + \frac{2}{3\sqrt{3}} \frac{\tsopt}{\tuulm} \left( 1-\frac{2}{\tuulm^2} + O \left( \frac{1}{\tuulm^4} \right) \right) \left( 1 + \frac{1}{\tsopt^2} + O\left( \frac{1}{\tsopt^4} \right) \right) \leq 0
\end{align}
and after substituting $\uulm = \lambda \sopt$ with $\lambda<1$ we get the following:
\begin{align}
    \kappa^3 - \kappa + \frac{2}{3\sqrt{3}} + \frac{2}{3\sqrt{3}}\frac{\log\lambda -1}{\tsopt^2} + O \left( \frac{1}{\tsopt^4} \right) \leq 0
\end{align}
Thus, we obtain $\kappa^2 \in \left[ \frac{1}{3}-\frac{\tlambda}{\tsopt^{2/3}}, \frac{1}{3}+\frac{\tlambda}{\tsopt^{2/3}} \right]$ for some positive universal constant $\tlambda$, which we can write as $\nu\norm{\wgt}_1 = \frac{\loneboundclasnt}{3} (1+O(\tsopt^{-2/3}))$.

\subsubsection*{Step 3.4: Tight bounds on $\etasc$}
Define $\Gamma_3^{\etasc} := \Gamma_3 \cap \left\{(\nu,b,\alpha,\etas) \subjto \left| \nu\norm{\wgt}_1-\frac{\loneboundclasnt}{3} \right| \leq \frac{\tlambda \loneboundclasnt}{\tsopt^{2/3}}\right\}$.
Since inequality~\eqref{eq:lower_bound_alpha_fcn_kappa} holds on $\Gamma_3$, it also holds on $\Gamma_3^{\etasc}$. Multiplying this inequality by $\frac{n\norm{\wgt}_1}{\loneboundclasnt \norm{\hgaussian}_{\infty}^2} (1-\kappa)^2(1+\OOO_b)^{-1}$, we get:
\begin{align}
    \sqrt{\frac{2}{\pi}} \frac{\norm{\gamma(\alpha)}_2^2}{\norm{\hgaussian}_{\infty}^2} &\frac{n\norm{\wgt}_1}{\loneboundclasnt} (1-\kappa)^2 (1+\OOO_b)^{-1} + \frac{\sqrt{2}}{3\sqrt{\pi}} \frac{\alpha^2}{\norm{\hgaussian}_{\infty}^2} \frac{n\norm{\wgt}_1}{\loneboundclasnt^3} (1+\OOO_b)^{-1}(1-O\left(\frac{1}{\tsopt^4}\right))\\
    &\leq
    \kappa(1-\kappa)^2 \leq \frac{4}{27}
\end{align}
and using our established bound on $\nu\norm{\wgt}_1$ we get $(1-\kappa)^2 \geq (1- \frac{1}{3}-\frac{\tlambda}{\tsopt^{2/3}})^2 = \frac{4}{9} (1 - \frac{3\tlambda}{\tsopt^{2/3}} + O(\frac{1}{\tsopt^{4/3}}) )$ and hence we obtain:
\begin{align}
    3\sqrt{\frac{2}{\pi}} \frac{\norm{\gamma(\alpha)}_2^2}{\norm{\hgaussian}_{\infty}^2}\frac{n\norm{\wgt}_1}{\loneboundclasnt} \left(1 - \frac{3\tlambda}{\tsopt^{2/3}} + O\left(\frac{1}{\tsopt^{4/3}}\right) \right) +  \frac{9\sqrt{2}}{4\sqrt{\pi}} \frac{\alpha^2}{\norm{\hgaussian}_{\infty}^2} \frac{n\norm{\wgt}_1}{\loneboundclasnt^3} (1- O\left(\frac{1}{\tsopt^4}\right))
    \leq 1
    \label{eq:tight_bound_noisy_alpha_ineq_tight}
\end{align}
Note that the function is convex in $\alpha$. Using concentration, we get for $\alpha=\alpha_m$:
\begin{align}
    2 \frac{2\left(\frac{3}{\pi}\right)^{1/3}(n \tsopt \norm{\wgt}_1 )^{2/3}}{m \tm^2} &\left(1+O \left( \frac{1}{\tm^2} \right) \right) \left(1 - \frac{3\tlambda}{\tsopt^{2/3}} + O \left( \frac{1}{\tsopt^{4/3}} \right)  \right) \\
    &+
    \frac{\tsopt^2}{\tm^2} \left(1-\frac{4}{\tm^2}+ O\left( \frac{1}{\tm^4} \right) \right) \left(1+\frac{2}{\tsopt^2} + O\left( \frac{1}{\tsopt^4} \right) \right)
    \leq 1
\end{align}
Now we claim that the $\ulmstar<\sopt,\olmstar>\sopt$ given in Lemma~\ref{lemma_uls_ols_exp}, respectively, with $\kappa=1/3$ and parameter $\mu$ do not satisfy this inequality for the well-chosen universal constant $\mu$ since
\begin{align}
    2& \frac{2\left(\frac{3}{\pi}\right)^{1/3}(n \tsopt \norm{\wgt}_1)^{2/3}}{\ulmstar \tulmstar^2} \left(1 - \frac{3\tlambda}{\tsopt^{2/3}} +  O \left( \frac{1}{\tsopt^{4/3}} \right) \right) 
    +
    \frac{\tsopt^2}{\tulmstar^2} \left( 1- \frac{2}{\tsopt^2} + O \left( \frac{1}{\tsopt^4} \right) \right) 
    \\
    &=
    1 - \frac{\mu}{\tsopt^{7/3}} \left(1-\frac{\tsopt^2}{\tulmstar^2} \right)  + \frac{2\mu^2-6\tlambda}{\tulmstar^2\tsopt^{2/3}} + O \left( \frac{1}{\tsopt^{10/3}} \right) 
    =
    1 +\frac{2\mu^2-6\tlambda}{\tsopt^{8/3}} + O \left( \frac{1}{\tsopt^{10/3}} \right) > 1
\end{align}
for $\mu > \sqrt{3\tlambda}$. Similarly, for $\olmstar$ we get:
\begin{align}
    2& \frac{2\left(\frac{3}{\pi}\right)^{1/3}(n \tsopt \norm{\wgt}_1)^{2/3}}{\olmstar \tolmstar^2} \left(1 - \frac{3\tlambda}{\tsopt^{2/3}} + O \left( \frac{1}{\tsopt^{4/3}} \right) \right) 
    +
    \frac{\tsopt^2}{\tolmstar^2} \left( 1- \frac{2}{\tsopt^2} + O \left( \frac{1}{\tsopt^4} \right) \right) 
    \\
    &=
    1 + \frac{\mu}{\tsopt^{7/3}} \left(1-\frac{\tsopt^2}{\tolmstar^2} \right)  + \frac{2\mu^2-6\tlambda}{\tolmstar^2\tsopt^{2/3}} + O \left( \frac{1}{\tsopt^{10/3}} \right) 
    =
    1 +\frac{2\mu^2-6\tlambda}{\tsopt^{8/3}} + O \left( \frac{1}{\tsopt^{10/3}} \right) > 1
\end{align}
In order to bound $\etasc$ we use that $b\leq \frac{\loneboundclasnt -\nu\norm{\wgt}_1}{\alpha}$, $\alpha \geq \alpha_{\ulmstar}$, and $\nu \geq \loneboundclasnt(\frac{1}{3}- \frac{\tlambda}{\tsopt^{2/3}})$, respectively, to obtain:
\ifaistats
\begin{align}
    \etasc^2 
    &\leq
    \max_{(\nu,b,\alpha,\etas)\in\Gamma_3^{\etasc}} b^2 \norm{\gamma(\alpha)}_2^2 
    \leq 
    \max_{\nu,\alpha} (\loneboundclasnt-\nu\norm{\wgt}_1)^2 \frac{\norm{\gamma(\alpha)}_2^2}{\alpha^2}
    \leq
    \loneboundclasnt^2 \left( 1-\frac{1}{3} + \frac{\tlambda}{\tsopt^{2/3}} \right)^2 \frac{\norm{\gamma(\alpha_{\ulmstar})}_2^2}{\alpha_{\ulmstar^2}}
\end{align}
\else
\begin{align}
    \etasc^2 
    \leq
    \max_{(\nu,b,\alpha,\etas)\in\Gamma_3^{\etasc}} b^2 \norm{\gamma(\alpha)}_2^2 
    &\leq 
    \max_{\nu,\alpha} (\loneboundclasnt-\nu\norm{\wgt}_1)^2 \frac{\norm{\gamma(\alpha)}_2^2}{\alpha^2}\\
    &\leq
    \loneboundclasnt^2 \left( 1-\frac{1}{3} + \frac{\tlambda}{\tsopt^{2/3}} \right)^2 \frac{\norm{\gamma(\alpha_{\ulmstar})}_2^2}{\alpha_{\ulmstar^2}}
\end{align}
\fi
and after application of concentration Proposition~\ref{prop:concentration_gammas} and definition of $\loneboundclasnt$ we obtain:
\begin{align}
    \etasc^2 
    &\leq
    \frac{2}{3} \frac{1}{\tsopt^2} \exp \left( \frac{\mu}{2\tsopt^{1/3}} \right) \left(1+ O \left( \frac{1}{\tsopt^{2/3}} \right) \right)
    = 
    \frac{2}{3} \frac{1}{\tsopt^2} \left( 1 + \frac{\mu}{2\tsopt^{1/3}} + O \left( \frac{1}{\tsopt^{2/3}} \right) \right) 
\end{align}
and
\begin{align}
    \etasc^2 
    &\geq
    \min_{(\nu,b,\alpha,\etas)\in\Gamma_3^{\etasc}} b^2\norm{\gamma(\alpha)}_2^2 
    \geq
    \min_{\nu,\alpha} \frac{\sqrt{2}}{3\sqrt{\pi}} \frac{n}{\nu} \frac{\norm{\gamma(\alpha)}_2^2}{\norm{\hgaussian}_{\infty}^2}(1-O\left(\frac{1}{\tsopt^4}\right))\\
    &\geq
    \frac{\sqrt{2}}{3\sqrt{\pi}} \frac{n\norm{\wgt}_1}{\loneboundclasnt \left( \frac{1}{3}+\frac{\tlambda}{\tsopt^{2/3}} \right)} \frac{2}{\olmstar\tolmstar^2} \left( 1 + O \left( \frac{1}{\tolmstar^2} \right) \right)
    \geq
    \frac{2}{3} \frac{1}{\tsopt^2} \exp \left( -\frac{\mu}{2\tsopt^{1/3}} \right) \left(1 - O \left( \frac{1}{\tsopt^{2/3}} \right) \right)\\
    &\geq
    \frac{2}{3} \frac{1}{\tsopt^2} \left( 1 - \frac{\mu}{2\tsopt^{1/3}} - O \left( \frac{1}{\tsopt^{2/3}} \right) \right),
\end{align}
which are the upper and lower bound claimed in the Proposition~\ref{prop:BoundGamma0_class_noiseless}.

\subsubsection*{Step 3.5: Tight upper bound on $\etas$}
Define $\Gamma_3^{\etas} := \Gamma_3 \cap \left\{(\nu,b,\alpha,\etas) \subjto \left| \nu\norm{\wgt}_1-\frac{\loneboundclasnt}{3} \right| \leq \frac{\tlambda \loneboundclasnt}{\tsopt^{2/3}} ~~~\andtxt~~~ \alpha \leq \alpha_{\olmstar} ~~~\andtxt~~~ \alpha \geq \alpha_{\ulmstar} \right\}$.
In this step we keep the term $\frac{\etas^2}{\nu}$ from the first constraint of $\Gamma_3$, and repeat the same steps leading to Equation~\eqref{eq:lower_bound_alpha_fcn_kappa} to obtain constraint:
\begin{align}
    \frac{\sqrt{2}}{3\sqrt{\pi}} \frac{n\norm{\wgt}_1\alpha^2}{\loneboundclasnt^3 \kappa(1-\kappa)^2\norm{\hgaussian}_{\infty}^2}(1-O\left(\frac{1}{\tsopt^4}\right)) 
    &+
    \sqrt{\frac{2}{\pi}} \etas^2 \frac{\alpha^2 n\norm{\wgt}_1}{\norm{\hgaussian}_{\infty}^2 \loneboundclasnt^3 \kappa(1-\kappa)^2}(1+\OOO_b)^{-1} \\
    &+
    \sqrt{\frac{2}{\pi}} \norm{\gamma(\alpha)}_2^2\frac{n\norm{\wgt}_1}{\norm{\hgaussian}_{\infty}^2\kappa\loneboundclasnt}(1+\OOO_b)^{-1}
    \leq 
    1
\end{align}
As in the Step 3.3 we have that $\kappa(1-\kappa)^2 \leq \frac{4}{27}$ and $\kappa\leq \frac{1}{3}+\frac{\tlambda}{\tsopt^{2/3}}$. Plugging these two bounds into the inequality above, we further relax the constraint to:
\ifaistats
\begin{align}
    \etas^2 \frac{\alpha^2}{\norm{\hgaussian}_{\infty}^2} \frac{n\norm{\wgt}_1}{\loneboundclasnt^3} \lesssim 
    1-3\sqrt{\frac{2}{\pi}} \frac{\norm{\gamma(\alpha)}_2^2}{\norm{\hgaussian}_{\infty}^2}\frac{n\norm{\wgt}_1}{\loneboundclasnt} (1 - \frac{3\tlambda}{\tsopt^{2/3}} +  O\left( \frac{1}{\tsopt^{4/3}} \right) ) -  \frac{9\sqrt{2}}{4\sqrt{\pi}} \frac{\alpha^2}{\norm{\hgaussian}_{\infty}^2} \frac{n\norm{\wgt}_1}{\loneboundclasnt^3} (1-O\left( \frac{1}{\tsopt^4} \right))
\end{align}
\else
\begin{align}
    \etas^2 \frac{\alpha^2}{\norm{\hgaussian}_{\infty}^2} \frac{n\norm{\wgt}_1}{\loneboundclasnt^3} \lesssim 
    1-&3\sqrt{\frac{2}{\pi}} \frac{\norm{\gamma(\alpha)}_2^2}{\norm{\hgaussian}_{\infty}^2}\frac{n\norm{\wgt}_1}{\loneboundclasnt} (1 - \frac{3\tlambda}{\tsopt^{2/3}} +  O\left( \frac{1}{\tsopt^{4/3}} \right) ) \\
    &-  \frac{9\sqrt{2}}{4\sqrt{\pi}} \frac{\alpha^2}{\norm{\hgaussian}_{\infty}^2} \frac{n\norm{\wgt}_1}{\loneboundclasnt^3} (1-O\left( \frac{1}{\tsopt^4} \right))
\end{align}
\fi
At the end we use derived bounds on $\alpha$ to upper bound $\etas$ as follows:
\ifaistats
\begin{align}
    \etas^2 
    &\lesssim 
    \frac{\loneboundclasnt^3 \norm{\hgaussian}_{\infty}^2}{\alpha_{\ulmstar}^2 n\norm{\wgt}_1} \left[
    1-3\sqrt{\frac{2}{\pi}} \frac{\norm{\gamma(\alpha_{\olmstar} )}_2^2}{\norm{\hgaussian}_{\infty}^2}\frac{n\norm{\wgt}_1}{\loneboundclasnt} (1 - \frac{3\tlambda}{\tsopt^{2/3}} +  O \left( \frac{1}{\tsopt^{4/3}} \right) ) -  \frac{9\sqrt{2}}{4\sqrt{\pi}} \frac{\alpha_{\ulmstar}^2}{\norm{\hgaussian}_{\infty}^2} \frac{n\norm{\wgt}_1}{\loneboundclasnt^3} (1-O\left( \frac{1}{\tsopt^4} \right))\right]
\end{align}
\else
\begin{align}
    \etas^2 
    \lesssim 
    \frac{\loneboundclasnt^3 \norm{\hgaussian}_{\infty}^2}{\alpha_{\ulmstar}^2 n\norm{\wgt}_1} \bigg[
    1-&3\sqrt{\frac{2}{\pi}} \frac{\norm{\gamma(\alpha_{\olmstar} )}_2^2}{\norm{\hgaussian}_{\infty}^2}\frac{n\norm{\wgt}_1}{\loneboundclasnt} (1 - \frac{3\tlambda}{\tsopt^{2/3}} +  O \left( \frac{1}{\tsopt^{4/3}} \right) ) \\
    &-  \frac{9\sqrt{2}}{4\sqrt{\pi}} \frac{\alpha_{\ulmstar}^2}{\norm{\hgaussian}_{\infty}^2} \frac{n\norm{\wgt}_1}{\loneboundclasnt^3} (1-O\left( \frac{1}{\tsopt^4} \right))\bigg]
\end{align}
\fi
Finally, after application of concentration Proposition~\ref{prop:concentration_gammas} and definitions of $\alpha_{\ulmstar},\alpha_{\olmstar}$ and $\loneboundclasnt$ we obtain $\etas^2 \lesssim \frac{1}{\tsopt^{7/3}}$, which finishes the proof of this proposition.

\section{Proof of Theorem~\ref{thm:mainl1_class_noisy}} 
\label{subsec:proof_sketch_noisy_main}
In this section we present the proof of  Theorem~\ref{thm:mainl1_class_noisy}. We begin by recalling some definitions: $\fnclas(\nu,\eta) = \frac{1}{n} \sum_{i=1}^n ( 1 - \xi_i \nu |z_i^{(0)}| - z_i^{(1)} \eta )_+^2$ and $\fclas(\nu,\eta) = \EE  \fnclas \left( \nu, \eta \right)  = \EE \left( 1- \xi \nu \vert \Gausswone \vert -  \Gausswtwo \eta \right)_+^2 $ and $\nubar := \arg\min \fclas(\nu, 0)$. Further, define  
$\fstar = \fclas(\nubar, 0)$, ${\feestar = \frac{d^2}{d^2 \eta}\vert_{(\nubar, 0)} \fclas(\nu,\eta)}$, ${\fnnstar = \frac{d^2}{d^2 \nu}\vert_{(\nubar, 0)} \fclas(\nu,\eta)}$. 
which are all non-zero positive constants.  
We define the constant $\kappanoise$ in Theorem~\ref{thm:mainl1_class_noisy} by:
\begin{equation}
\label{eq:kappanoise}
  \kappanoise = \frac{ 2 \fstar }{\feestar \nubar^2 \pi^2}.
\end{equation}
In a first localization step, we bound $\PhiCNuni$. By proposition~\ref{prop:CGMT_application_classification}, it suffices to upper bound $\PhiCdn$, which by Proposition~\ref{prop:parametrization_class_general} can be reduced to a low dimensional stochastic optimization problem. We show:
\begin{proposition}
\label{prop:Rbound_noisy}
     Let the assumptions of Theorem~\ref{thm:mainl1_class_noisy} hold.
     Let $\tsoptn$ (as in Equation~\eqref{eq:deftm} in Appendix~\ref{subsubsec:prop_gamma_alpha}) be such that $2\Phic(\tsoptn) = \soptn/d$ with $\soptn = n\feestar/2$.
     There exist universal positive constants $c_1,c_2,c_3>0$ such that
    \begin{align} \label{PhiCdn_bound}
        (\PhiCdn)^2 
        \leq
        \frac{ n \fstar }{ \tsoptn^2 } \left( 1 - \frac{ 2 }{ \tsoptn^2 } + \frac{c_1}{ \tsoptn^3 } \right) =:\loneboundclas^2
    \end{align}
with probability at least $ 1 - c_2 \exp \left( -c_3\frac{n}{\log^5(d/n)} \right)$
over the draws of $\hgaussian_1,\hgaussian_2$, $\gausswone,\gausswtwo$ and $\xi$. 
\end{proposition}  
The proof of the proposition is is deferred to Appendix~\ref{subsec:proof_loc_noisy}. As described in Section~\ref{subsec:proof_sketch}, in a second uniform convergence step, we bound the constraint set $\setZeroUni$ from Equation~\eqref{eq:Gamma0_def_class_uni}:

\begin{proposition}
\label{prop:BoundGamma0_class_noisy}
Let the assumptions of Theorem~\ref{thm:mainl1_class_noisy} hold and let $\setNoisyZero$ be as in Equation~\eqref{eq:Gamma0_def_class_uni} with $\loneboundclas$ from Proposition~\ref{prop:Rbound_noisy}. 
Define a set $\tsetNoisyZero$ as a set of all $(\nu,b,\alpha,\etas)$ that satisfy:
\ifaistats
\begin{gather} 
    \abs{ \nu-\nubar }^2 \lesssim \frac{1}{ \log(d/\soptn)} ~~\andtxt~~ \etas^2 \lesssim  \frac{1}{\log^{5/4}(d/\soptn)} ~~\andtxt~~ 
    \left| b^2\norm{\gamma(\alpha)}_2^2 - \frac{2\fstar}{\feestar \log(d/\soptn) } \right| \lesssim   \frac{1}{\log^{5/4}(d/\soptn)}  
\end{gather}
\else
\begin{align} 
    &\abs{ \nu-\nubar }^2 \lesssim \frac{1}{ \log(d/\soptn)} ~~\andtxt~~ \etas^2 \lesssim  \frac{1}{\log^{5/4}(d/\soptn)} \\&~~\andtxt~~ 
    \left| b^2\norm{\gamma(\alpha)}_2^2 - \frac{2\fstar}{\feestar \log(d/\soptn) } \right| \lesssim   \frac{1}{\log^{5/4}(d/\soptn)}  
\end{align}
\fi
with $\soptn = n \feestar/2$.
There exist universal constants $ c_1, c_2, c_3,c_4 > 0 $ such that
$\setNoisyZero \subset \tsetNoiselessZero$
with probability at least $1 - c_1\exp \left( -c_2  \frac{n}{\log^5(d/n)}  \right) -  c_3\exp \left( -c_4 \frac{ n }{\log n \log^{3/2}(d/n)} \right)$ over the draws of $\hgaussian_1,\hgaussian_2$, $\gausswone,\gausswtwo$ and $\xi$.
\end{proposition}  
The proof of the proposition is deferred to Appendix~\ref{subsec:proof_unicon_noisy}.   As a consequence, when applying Proposition~\ref{prop:parametrization_class_general} we can upper and lower bound $\PhiCdp$ and $\PhiCdm$:
\begin{align}
    \PhiCdp
    \leq \left[1 + \frac{\min\limits_{(b,\alpha) \in \tsetNoisyZero} ~b^2\norm{\gamma(\alpha)}_2^2 + \min\limits_{\etas \in \tsetNoisyZero} ~\etas^2}{\max\limits_{\nu \in \tsetNoisyZero}~ \nu^2}  \right]^{-1/2}
    \leq 1 - \frac{ \fstar }{ \feestar \nubar^2 } \frac{1}{ \log(d/ \soptn)} \left( 1 - \frac{c}{\log(d/\soptn)^{1/4}} \right) 
\end{align}
\begin{align}
    \PhiCdm \geq \left[1 + \frac{\max\limits_{(b,\alpha) \in \tsetNoisyZero} ~b^2\norm{\gamma(\alpha)}_2^2 + \max\limits_{\etas \in \tsetNoisyZero} ~\etas^2}{\min\limits_{\nu \in \tsetNoisyZero}~ \nu^2}  \right]^{-1/2}
    \geq 1 - \frac{ \fstar }{ \feestar \nubar^2 } \frac{1}{ \log(d/ \soptn)} \left( 1 + \frac{c}{\log(d/\soptn)^{1/4}}\right) 
\end{align}
Where we slightly abuse the notation by writing $(b,\alpha) \in \tsetNoisyZero$ and similar for $\nu \in \tsetNoisyZero$ and $\etas \in \tsetNoisyZero$. Finally, the proof follows when applying Proposition~\ref{prop:CGMT_application_classification} and using 
the exact same series expansion for risk as in Equation~\eqref{eq:risk_equivalence_appendix_noiseless}.
\subsection{Proof of Localization Proposition~\ref{prop:Rbound_noisy}}
\label{subsec:proof_loc_noisy}
Recall the upper bound for $\PhiCdn$ from Proposition ~\ref{prop:parametrization_class_general}. Since $\wgt$ is $s$-sparse vector, we have that $\norm{\wgt}_1 \leq \sqrt{s}$, and we can further upper bound $\PhiCdn$ as follows:
\begin{align}
    \PhiCdn \leq 
    \min_{\nu,b,\alpha} \vert \nu \vert \sqrt{s}   + b \norm{ \gamma \left(\alpha \right) }_1 
    \subjto \frac{ 1 }{n} b^2\norm{\hgaussian}_{\infty}^2 \geq \fnclas \left( \nu, b\norm{ \gamma(\alpha)}_2 \right)
    \label{eq:proof_loc_noisy_phicn_first}
\end{align}

Given that $(\tnu,\tb,\talpha)$ is a feasible point for given upper bound, we have that $\PhiCdn \leq  \vert \tnu \vert \sqrt{s} + \tb \norm{ \gamma( \talpha) }_1$. Thus, in the following discussion our goal is to find a single feasible point of the constraint set from Equation~\eqref{eq:proof_loc_noisy_phicn_first}.

In order to show that a point satisfies the constraint above, it is necessary to evaluate function $\fnclas ( \nu, b \norm{ \gamma(\alpha)}_2 )$ at this point. We do this by using concentration of Lipschitz continuous function from Lemma~\ref{lemma:conc_lipschitz}. Namely, recall that we defined $\fclas = \EE [ \fnclas]$ and thus according to Lemma~\ref{lemma:conc_lipschitz}
for any $\nu,\eta$ holds that:
    \begin{align}
        \prob \left( \abs{ \fnclas \left( \nu, \eta\right) -  \fclas \left( \nu, \eta\right)  }  \geq \epsilon \right) \leq 2\exp \left( - c \frac{ n \epsilon^2 }{ \nu^2 + \eta^2 } \right)
        \label{eq:point_conv_Lipschitz}
    \end{align}
with some universal constant $c>0$. 
Hence, with high probability we can approximate evaluation of function $\fnclas$ at a point by evaluation of function $\fclas$ at the same point. 

From definition of $\gamma ( \alpha )$ we know that $ \norm{ \gamma ( \alpha ) }_1 = \alpha$ and hence we can upper bound $\PhiCdn$ by an optimization problem over $\nu>0$ and $ b_{\alpha} := b \alpha $ as follows:
\begin{align}
    \PhiCdn &\leq \min_{ \nu, b_{\alpha},\alpha } \nu \sqrt{s} + b_{\alpha}
    \subjto \frac{1}{n}\frac{ b_{\alpha}^2 }{\alpha^2}\norm{\hgaussian}_{\infty}^2 \geq \fnclas \left( \nu, b_{\alpha} \frac{ \norm{ \gamma \left( \alpha \right) }_2}{ \alpha } \right)
    \label{eq:loc_proof_noisy_phicn_removeds}
\end{align}

Using Equation~\eqref{eq:point_conv_Lipschitz} with $ \epsilon = \fstar \tsoptn^{-3} $ and for a feasible point $( \nu, b_{\alpha} \frac{ \norm{ \gamma \left( \alpha \right) }_2}{ \alpha } )$ we have that:
\begin{align} \label{eq:constraint_f_conc}
    \frac{ b_{\alpha}^2 \norm{\hgaussian}_{\infty}^2 }{n \alpha^2} 
    \geq
    \fclas \left( \nu, b_{\alpha} \frac{ \norm{ \gamma \left( \alpha \right) }_2}{ \alpha }\right) + \frac{\fstar}{\tsoptn^{3}}
\end{align}
with probability at least $ 1 - 2 \exp \left( -c \frac{ n }{ \tsoptn^6 \left( \nu^2 + b_{\alpha}^2  \norm{ \gamma \left( \alpha \right) }_2^2 / \alpha^2 \right) } \right) $.

Recall that we defined $\nubar := \arg\min \fclas(\nu, 0)$. Now let us choose $ \tnu = \nubar $ and show that there exists a pair $ ( b, \alpha )$  such that $ ( \nubar, b, \alpha) $ is feasible for constraint \eqref{eq:constraint_f_conc}. We propose to search for a point with parameter $ ( b, \alpha ) $ such that $ b \norm{ \gamma ( \alpha )}_2 = b_{\alpha} \frac{ \norm{ \gamma \left( \alpha \right) }_2}{ \alpha } $ is near zero. We show in Lemma~\ref{lemma:fclas_fclasn_inf_diff} that $\fclas$ is infinitely differentiable function and thus, using Taylor series approximation of the function $ \fclas ( \nubar, \cdot) : \eta \mapsto \fclas (\nubar, \eta) $ around  the point $(\nuopt,0)$ we can rewrite the constraint \eqref{eq:constraint_f_conc} as:
\begin{align}
     \frac{ b_{\alpha}^2 \norm{\hgaussian}_{\infty}^2 }{n \alpha^2}   
     \geq
      \fstar + \frac{1}{2}\feestar b_{\alpha}^2 \frac{ \norm{ \gamma \left( \alpha \right) }_2^2}{ \alpha^2 } + O \left( b_{\alpha}^3 \frac{ \norm{ \gamma \left( \alpha \right) }_2^3}{ \alpha^3 } \right) + \frac{\fstar}{\tsoptn^{3}}
      \label{eq:proof_loc_noisy_balpha_midexp}
\end{align}
with $\zeta_{\eta} := \frac{\partial \fclas(\nuopt,\eta)}{\partial \eta}\Big|_{\eta=0} = 0$ and where we recall that by definition $\fstar = \fclas \left( \nubar, 0 \right)$, $\feestar = \frac{\partial^2 \fclas (\nuopt,\eta)}{\partial \eta^2}\Big|_{\eta=0}$ and $\soptn = \frac{1}{2}\feestar n$.

As we mentioned in Section~\ref{subsubsec:prop_gamma_alpha}, $ \gamma ( \alpha ) $ is a piecewise linear function with breakpoints at $ \alpha_m $ for $ m = 2, \dots, d $. Hence instead of optimizing over $\alpha$ we optimize over $m$. Rearranging the terms from Equation~\eqref{eq:proof_loc_noisy_balpha_midexp} we get:
\begin{equation}
    b_\alpha^2 \geq \frac{ n \alpha_{m}^2 }{ \norm{\hgaussian}_{\infty}^2 } \frac{ \fstar \left( 1 + \frac{1}{\tsoptn^{3}} \right) }{ 1 - \frac{1}{2} n \feestar \frac{ \norm{ \gamma \left( \alpha_{m} \right) }_2^2 }{ \norm{\hgaussian}_{\infty}^2 } - O \left( b_\alpha n \frac{\norm{\gamma(\alpha_m)}_2^3}{\alpha_m\norm{\hgaussian}_{\infty}^2} \right) }
    \label{eq:B_ineq}
\end{equation}
Note that we have only one constraint but two free variables $ ( b, \alpha ) $ and so we can set $ \talpha = \alpha_{\soptn} $ with $ \soptn = \frac{1}{2}\feestar n $. To gain an intuition for why this choice is approximately optimal, one can follow a similar argument as in Remark 1 in \cite{wang2021tight} and show that $\soptn$ approximately maximizes expression:
\begin{equation}
   \frac{ \norm{\hgaussian}_{\infty}^2 }{ \alpha_{m}^2 } \left( 1 - \frac{1}{2} n \feestar \frac{ \norm{ \gamma \left( \alpha_{m} \right) }_2^2 }{ \norm{\hgaussian}_{\infty}^2 } - O \left( b_\alpha n \frac{\norm{\gamma(\alpha_m)}_2^3}{\alpha_m\norm{\hgaussian}_{\infty}^2} \right) \right)
\end{equation}
Thus, $\soptn$ approximately minimizes expression on the right hand side of Equation~\eqref{eq:B_ineq} and maximally relaxes this constraint on $b_\alpha^2$. We now claim that
\begin{equation}
    \tb_\alpha^2 = \frac{ n \alpha_{\soptn}^2 }{ \norm{\hgaussian}_{\infty}^2 } \frac{ \fstar \left( 1 + \frac{1}{\tsoptn^{3}} \right) }{ 1 - \frac{1}{2} n \feestar \frac{ \norm{ \gamma \left( \alpha_{\soptn} \right) }_2^2 }{ \norm{\hgaussian}_{\infty}^2 } - O \left( \frac{1}{\tsoptn^3} \right) }
\end{equation}
satisfies inequality~\eqref{eq:B_ineq} with probability at least $ 1 - 6 \exp \left( -\frac{2\soptn}{\log^5(d/\soptn)} \right)$. Using Proposition~\ref{prop:concentration_gammas} we have with high probability that:
\begin{equation}
    1 - \frac{1}{2} n \feestar \frac{ \norm{ \gamma \left( \alpha_{\soptn} \right) }_2^2 }{ \norm{\hgaussian}_{\infty}^2 } - O \left( \frac{1}{\tsoptn^3} \right) > 1 - \frac{1}{2} n \feestar  \frac{2}{\soptn \tsoptn^2} - O \left( \frac{1}{\tsoptn^3} \right) = 1 - \frac{2}{\tsoptn^2} - O \left( \frac{1}{\tsoptn^3} \right)  > 0
\end{equation}
for $d,n$ sufficiently large. Applying Proposition~\ref{prop:concentration_gammas} once again we can upper bound $\tb_\alpha$:
\begin{equation}
    \tb_\alpha^2 
    \leq
    \frac{ n \fstar }{ \tsoptn^2 } \left( 1 + \frac{1}{\tsoptn^3} \right) \left( 1 - \frac{4}{\tsoptn^2} + \frac{c}{\tsoptn^4} \right) \frac{1}{ 1 - \frac{2}{\tsoptn^2} - O \left( \frac{1}{\tsoptn^3} \right)}
    \leq
    \frac{ n \fstar }{ \tsoptn^2 } \left( 1 - \frac{2}{\tsoptn^2} + \frac{c}{\tsoptn^3} \right)
\end{equation}

Now applying Proposition~\ref{prop:concentration_gammas} we have that $ O \left( \tb_\alpha n  \frac{ \norm{ \gamma \left( \alpha_{\soptn} \right) }_2^3}{ \alpha_{\soptn} \norm{\hgaussian}_{\infty}^2 } \right) = O \left( \frac{\sqrt{n}}{\tsoptn}n \frac{1}{\soptn\sqrt{\soptn} \tsoptn^2} \right)= O \left( \frac{1}{\tsoptn^3} \right)$ and $\tb_\alpha^2$ indeed satisfies Equation~\eqref{eq:B_ineq}. From the upper bound on sparsity, we have $\nubar \sqrt{s} \lesssim \frac{\sqrt{n}}{\tsoptn^4}$. Since $(\tnu,\tb,\talpha)$ is a feasible point, from Equation~\eqref{eq:loc_proof_noisy_phicn_removeds} and derived bounds on $\nubar \sqrt{s}$ and $\tb_\alpha$ follows that
\begin{align}
\label{eq:lonebound_def}
    \loneboundclas := \sqrt{ \frac{ n \fstar }{ \tsoptn^2 } \left( 1 - \frac{2}{\tsoptn^2} + \frac{\tc}{\tsoptn^3} \right) }
\end{align}
is an upper bound on $\PhiCdn$ with probability at least $ 1 - 2 \exp \left( -c \frac{ n }{ \log^3(d/n) \left( \nubar^2 + b_\alpha^2  \norm{ \gamma \left( \alpha_{\soptn} \right) }_2^2 / \alpha_{\soptn}^2 \right) } \right) - 6 \exp \left( -c\frac{n}{\log^5(d/n)} \right)
$. The proposition is proved after noting that $ \nubar^2 + \tb_\alpha^2  \norm{ \gamma \left( \alpha_{\soptn} \right) }_2^2 / \alpha_{\soptn}^2 = O(1)$.

\subsection{Proof of Uniform Convergence Proposition~\ref{prop:BoundGamma0_class_noisy}}
\label{subsec:proof_unicon_noisy}
The proof of the proposition follows from several steps where in each step we approximate $\fnclas$ using the bounds on $(\nu, \etasc, \etas)$ from the previous steps to obtain a tighter bound on $(\nu, \etasc, \etas)$ using the tools developed in \cite{wang2021tight}.  
The probability statement in Proposition~\ref{prop:BoundGamma0_class_noisy} follows when taking the union bound over all equations which we condition on throughout the proof. 

Furthermore, we note that the set $\setZeroUni$ from Proposition~\ref{prop:parametrization_class_general} is not empty as clearly the choice $(\tnu,\tb,\talpha,0)$ from Section~\ref{subsec:proof_loc_noisy} leads with high probability to a feasible point due to the choice of $\loneboundclas$. Moreover, we can even relax set $\setZeroUni$ from Proposition~\ref{prop:parametrization_class_general} and bound the variables that are elements of the following set:
\begin{align}
    \left\{ (\nu, b, \alpha, \etas) \subjto\frac{1}{n}  (2\sqrt{\smax}\etas + b\norm{\hgaussian}_{\infty})^2   \geq  \fnclas(\nu,\sqrt{b^2\|\gamma(\alpha)\|_2^2+\etas^2}) ~~\andtxt~~  b\alpha \leq \loneboundclas \right\}
    \supset
    \setNoisyZero.
    \label{eq:Gamma0_def_class_noisy}
\end{align}
where we implicitly assume bounds $\etas\geq 0, b\geq0, \alpha\in[1,\alpham]$ in all of the following discussion. The inclusion of $\setNoisyZero$ into the above set holds, since any point satisfying $\max\left\{ |\nu| \|\wgtp\|_1 - \sqrt{s}\etas , 0 \right\} 
+ b\alpha \leq \loneboundclasn $ satisfies $ b\alpha \leq \loneboundclasn$ as well. In what follows, we bound variables of interest from Proposition~\ref{prop:BoundGamma0_class_noisy} if they are elements of the above given set, which, by the inclusion, implies high probability bounds of the same variables in the set $\setZeroUni$.

\subsubsection*{Bound 1: \texorpdfstring{$\nu^2,\etasc^2, \etas^2 = O(1)$}{}}
\label{sec:l1class_nueta_const_bound}

In order to apply Lemma~\ref{lemma_quadratic_plus_fstar_bound} in the next step,  which gives tight bounds for $\fnclas$, we first need to show that with high probability, $\nu^2,\eta^2, \etas^2 = O(1)$. This is the goal of this first step. More specifically, 
the goal of this first step is to show that there exist universal constants $B_{\nu,1}, B_{\etasc,1}, B_{\etas,1} > 0$ such that for any element $(\nu, b, \alpha, \etas)$ of $\Gamma_0$ we have that $ \nu^2 \leq B^2_{\nu,1} $, $ \etasc = b\norm{ \gamma(\alpha)}_2 \leq B_{\etasc,1}$ and $\etas \leq B_{\etas,1}$ with high probability over the draws of $\hgaussian_1,\hgaussian_2$, $\gausswone,\gausswtwo$ and $\xi$. 

For this first step, we use the fact that in the presence of label noise, $\fnclas$ is lower bounded by a quadratic function as stated in Lemma ~\ref{quadratic_bound_fn} i.e. we have that
\begin{align}
\fnclas(\nu,\sqrt{b^2\|\gamma(\alpha)\|_2^2+\etas^2}) \geq \constnu \nu^2 + \consteta (b^2\norm{\gamma(\alpha)}_2^2+\etas^2) \geq \consteta \etas^2
\end{align}
holds with probability $\geq 1-\exp(-cn)$. As a result, we can relax the first constraint in Definition \eqref{eq:Gamma0_def_class_noisy} of $\setNoisyZero$ to 
\begin{align}
    \frac{1}{n}(2\sqrt{\smax}\etas + b\norm{\hgaussian}_{\infty})^2   \geq \constnu \nu^2  + \consteta  b^2 \norm{\gamma(\alpha)}_2^2  + \consteta \etas^2
    \label{eq:proof_unicon_noisy_bound1_fapprox}
\end{align}
This implies that $\consteta\etas^2 \leq  \frac{1}{n}(2\sqrt{\smax}\etas + b\norm{\hgaussian}_{\infty})^2  \leq \frac{8}{n}\smax\etas^2+\frac{2}{n}b^2\norm{\hgaussian}_{\infty}^2$. Thus for some universal constants $c_1,c_2>0$ we have
\begin{align}
    \etas^2 \leq \frac{2}{\consteta n}b^2\norm{\hgaussian}_{\infty}^2 \left( 1-\frac{8}{\consteta n}\smax \right)^{-1} \leq \frac{2}{\consteta n}b^2\norm{\hgaussian}_{\infty}^2 \left( 1+\frac{c_1}{\tsoptn^8} \right) \leq \frac{c_2}{n}b^2\norm{\hgaussian}_{\infty}^2
    \label{eq:proof_unicon_noisy_step1_bound_etas_prelim}
\end{align}
where we used that $\smax = \Theta\left( \frac{n}{\tsoptn^8} \right)$. Now define universal constant $c>0$ as the smallest constant satisfying
\begin{align}
    \frac{1}{n}(2\sqrt{\smax}\etas + b\norm{\hgaussian}_{\infty})^2 \leq \frac{2}{n}b^2\norm{\hgaussian}_{\infty}^2 \left(1+\frac{4c_2}{n}\smax \right) \leq \frac{c}{n} b^2\norm{\hgaussian}_{\infty}^2
    \label{eq:proof_unicon_noisy_bound1_smax_approx}
\end{align}
Combining Equations~\eqref{eq:proof_unicon_noisy_bound1_fapprox} and \eqref{eq:proof_unicon_noisy_bound1_smax_approx} we can relax the first constraint of $\setNoisyZero$ to
\begin{equation}
    \frac{c}{n} b^2\norm{\hgaussian}_{\infty}^2  \geq  \constnu \nu^2  + \consteta  b^2 \norm{\gamma(\alpha)}_2^2  + \consteta \etas^2.
\end{equation} 
 
This approximation leads to an optimization problem similar to the ones discussed in Lemma 1 in \cite{wang2021tight}. After further relaxations we obtain exactly the same form of the inequality, and hence we can use the arguments from \cite{wang2021tight}. Define the following set:
\begin{align}
    \Gamma_1 
    =
    \left\{ (\nu, b, \alpha, \etas) \subjto \frac{c}{n} b^2\norm{\hgaussian}_{\infty}^2  \geq  \constnu \nu^2  + \consteta  b^2 \norm{\gamma(\alpha)}_2^2  + \consteta \etas^2
    \andtxts
    b \alpha \leq \loneboundclas  \right\}
\end{align}
It is evident from the previous discussion that $\setNoisyZero\subset \Gamma_1$ with high probability. Thus, deriving high probability bounds on $\Gamma_1$ give valid bounds for $\setNoisyZero$ as well. In the following three steps we bound  variables $\etasc,\nu,\etas$ from the set $\Gamma_1$, respectively. 

\paragraph{Step 1.1: Upper bound on $\etasc$.}
In this step, as well as in almost every step that follows, we use that by relaxing of constraints from definition of set $\Gamma_1$ and bounding variables on this larger set, we obtain valid bounds for variables in $\Gamma_1$ and, more specifically, in $\setNoisyZero$. Moreover, recall that by our reparametrization from Section~\ref{subsec:proof_path} we have $\etasc^2 = \normk{\wperpsc}_2^2 = b^2\norm{\gamma(\alpha)}_2^2$. Hence, we relax the first constraint in definition of $\Gamma_1$ to show that: 
\begin{align}
    \etasc^2 &\leq \max_{(\nu,b,\alpha,\etas)\in\Gamma_1} b^2 \norm{\gamma(\alpha)}_2^2 
    \leq
    \max_{b,\alpha} \left[b^2\norm{\gamma(\alpha)}_2^2 \subjto \frac{c}{n}b^2\norm{\hgaussian}_{\infty}^2 \geq \consteta b^2\norm{\gamma(\alpha)}_2^2 \andtxts b\alpha \leq \loneboundclas \right]\\
    &= 
    \max_{1\leq \alpha \leq \alpham} \left[\loneboundclas^2 \frac{ \norm { \gamma ( \alpha ) }_2^2 }{ \alpha^2 } 
    \subjto
    \frac{c}{n}\norm{\hgaussian}_{\infty}^2 \geq \consteta \norm{\gamma(\alpha)}_2^2
    \right]
    \label{eq:l1proofstep1bound}
\end{align}

Now note that as discussed in Section~\ref{subsubsec:prop_gamma_alpha} $\norm{\gamma(\alpha)}_2^2$ is convex. Therefore, the set of feasible $\alpha$ that satisfy the last constraint is an non-empty interval. Indeed, to see that the interval is non empty, recall that we defined $ \loneboundclas $ in such way that $ (b,\alpha_{\soptn}) \in \setNoisyZero $ with high probability for $b \alpha_{\soptn}\leq \loneboundclas$. As $ \setNoisyZero \subset \Gamma_1 \subset \{ \alpha \subjto \frac{c}{n}\norm{\hgaussian}_{\infty}^2 \geq \consteta \norm{\gamma(\alpha)}_2^2 \}$, with high probability $\alpha_{\soptn}$ satisfies constraint in Equation~\eqref{eq:l1proofstep1bound}. Furthermore,  since $\frac{ \norm { \gamma ( \alpha ) }_2^2 }{ \alpha^2 }$ is monotonically decreasing, to upper bound Equation~\eqref{eq:l1proofstep1bound} it is sufficient to find $\ulalpha < \alpha_{\soptn}$ such that the constraint from Equation~\eqref{eq:l1proofstep1bound} does not hold, i.e. we should have: 
\begin{align}
\label{eq:s_condition_one}
    \frac{\norm{\gamma(\ulalpha)}_2^2}{\norm{\hgaussian}_{\infty}^2} > \frac{c}{\consteta n} . 
\end{align}

It is sufficient to only consider the discretized version of $\alpha$, i.e., $\alpha_m$, for which we have access to tight concentration inequalities from Proposition~\ref{prop:concentration_gammas}. We now  claim that $\alpha_{\ulm}$ with $\ulm =\lambda_{\ulm} \frac{ n }{ \log (d/n) }$ satisfies the inequality~\eqref{eq:s_condition_one} for some positive universal constant $ \lambda_{\ulm} >0  $. Using the characterization $\tm^2 = 2\log (d/m) + O (\log \log (d/m) ) $ and concentration inequalities from Section~\ref{subsubsec:prop_gamma_alpha} we show that $\ulm$ satisfies Equation~\eqref{eq:s_condition_one} since
\begin{align}
    \frac{ 2 }{ \ulm \tulm^2 } \left( 1 - O \left( \frac{ 1 }{ \tulm^2 } \right) \right)
    & >
    \frac{ 1 }{ n \lambda_{\ulm} } \left( 1 - O \left( \frac{\log\log (d/n) }{\log (d/n) } \right) \right) > \frac{c}{\consteta n},
\end{align}
where last inequality holds for $d/n$ sufficiently large and $\lambda_{\ulm}$ small enough.

Therefore, from Equation~\eqref{eq:l1proofstep1bound} and the concentration inequality from Proposition \ref{prop:concentration_gammas}, we get:
\begin{align}
    \etasc^2 \leq  \loneboundclas^2 \frac{\norm{\gamma(\alphauls)}_2^2}{\alphauls^2} \leq \frac{ n \fstar }{ \tsoptn^2 } \frac{2}{\ulm}\left( 1 + O \left( \frac{1}{\tsopt^2} \right) \right) 
    =: B_{\etasc,1}^2,
\end{align}
with $B_{\etasc,1}= \Theta (1)$, as desired. 

\paragraph{Step 1.2: Upper bound on $\nu$.} Similarly as in the previous step, we first relax the first constraint from definition of $\Gamma_1$ and use obtained constraints to upper bound $\nu^2$ as follows:
\ifaistats
\begin{align} 
    \nu^2 &\leq 
    \max_{(\nu,b,\alpha,\etas)\in\Gamma_1} \nu^2 
    \leq
    \max_{\nu,b,\alpha,\etas}\left[ \nu^2 \subjto \frac{c}{n}b^2\norm{\hgaussian}_{\infty}^2 \geq \constnu \nu^2 \andtxts \frac{c}{n}b^2\norm{\hgaussian}_{\infty}^2 \geq \consteta b^2\norm{\gamma(\alpha)}_2^2 \andtxts b\alpha \leq \loneboundclas \right]\\
    &= \frac{ c }{ n \constnu }\norm{\hgaussian}_{\infty}^2 \max_{b,\alpha} \left[ b^2 
    \subjto \frac{c}{n}\norm{\hgaussian}_{\infty}^2 \geq \consteta \norm{\gamma(\alpha)}_2^2 \andtxts b\alpha \leq \loneboundclas \right]\\
    &= \frac{ c }{ n \constnu }\loneboundclas^2\norm{\hgaussian}_{\infty}^2  \min_{1\leq \alpha\leq \alpha_{\max}}\left[ \frac{1}{\alpha^2}\subjto \frac{c}{n}\norm{\hgaussian}_{\infty}^2 \geq \consteta \norm{\gamma(\alpha)}_2^2 \right] 
    \label{eq:nu_opt_first_reduction}
\end{align}
\else
\begin{align} 
    \nu^2 &\leq 
    \max_{(\nu,b,\alpha,\etas)\in\Gamma_1} \nu^2 \\
    &\leq
    \max_{\nu,b,\alpha,\etas}\left[ \nu^2 \subjto \frac{c}{n}b^2\norm{\hgaussian}_{\infty}^2 \geq \constnu \nu^2 \andtxts \frac{c}{n}b^2\norm{\hgaussian}_{\infty}^2 \geq \consteta b^2\norm{\gamma(\alpha)}_2^2 \andtxts b\alpha \leq \loneboundclas \right]\\
    &= \frac{ c }{ n \constnu }\norm{\hgaussian}_{\infty}^2 \max_{b,\alpha} \left[ b^2 
    \subjto \frac{c}{n}\norm{\hgaussian}_{\infty}^2 \geq \consteta \norm{\gamma(\alpha)}_2^2 \andtxts b\alpha \leq \loneboundclas \right]\\
    &= \frac{ c }{ n \constnu }\loneboundclas^2\norm{\hgaussian}_{\infty}^2  \min_{1\leq \alpha\leq \alpha_{\max}}\left[ \frac{1}{\alpha^2}\subjto \frac{c}{n}\norm{\hgaussian}_{\infty}^2 \geq \consteta \norm{\gamma(\alpha)}_2^2 \right] 
    \label{eq:nu_opt_first_reduction}
\end{align}
\fi
Since $\frac{1}{\alpha^2}$ is a monotonically decreasing function, we can use exactly the same reasoning as in the Step 1.1 to obtain a high probability upper bound $\frac{1}{\alpha^2} \leq \frac{1}{\alpha_{\ulm}^2}$. Hence, using Equation~\eqref{eq:nu_opt_first_reduction} and the concentration results from Proposition~\ref{prop:concentration_gammas} we upper bound $\nu$ as follows:
\begin{align}
    \nu^2 &\leq  \frac{ c }{ n \constnu } \loneboundclas^2 \frac{\norm{\hgaussian}_{\infty}^2}{\alphauls^2} 
    \leq
    \frac{c\fstar}{\constnu}\frac{\tulm^2}{\tsopt^2} \left( 1 + O \left( \frac{1}{\tsopt^2} \right) \right)
    =: B_{\nu,1}^2,
\end{align}
and in particular, after using the characterization $\tm^2 = 2\log (d/m) + O (\log \log (d/m) ) $ from Section~\ref{subsubsec:prop_gamma_alpha}, we have again that $B_{\nu,1} = \Theta(1)$. 
\paragraph{Step 1.3: Upper bound on $\etas$.}
Replacing $\nu$ by $\etas$ and applying exactly the same procedure as in the Step 1.2, we obtain that with high probability:
\begin{align}
    \etas^2 &\leq 
    \frac{ c }{ n \consteta } \loneboundclas^2 \frac{\norm{\hgaussian}_{\infty}^2}{\alphauls^2} 
    \leq
    \frac{c\fstar}{\consteta}\frac{\tulm^2}{\tsopt^2} \left( 1 + O \left( \frac{1}{\tsopt^2} \right) \right)
    =: B_{\etas,1}^2,
\end{align}
for $B_{\etas,1} = \Theta(1)$, which completes the first part of the proof.

\subsubsection*{Bound 2: \texorpdfstring{$\deltanu^2,\etasc^2,\etas^2 = O\left(\frac{1}{\log(d/n)}\right)$}{}}

Recall that $\nubar := \arg\min \fclas(\nu, 0)$ and define $\deltanu = \nu -\nubar$. Conditioning on the event where the bounds from the first step hold for $\nu,\etasc,\etas$, the goal of this second step is to show that for any element $(\nu, b, \alpha, \etas)$ of $\setNoisyZero$ we have that $\deltanu^2 = O \left( \frac{1}{ \log \left( d/n \right) }  \right) $, $ \etasc^2 = b^2 \norm{\gamma(\alpha)}_2^2 = O \left( \frac{1}{ \log \left( d/n \right) } \right)$ and $\etas^2 = O \left( \frac{1}{ \log \left( d/n \right) } \right) $ with high probability over the draws of $\hgaussian_1,\hgaussian_2$, $\gausswone,\gausswtwo$ and $\xi$.

From the previous step, we know that with high probability, $ \nu^2 \leq \boundnuone^2 $, $ \etasc \leq B_{\etasc,1} $ and $\etas \leq B_{\etas,1}$. Hence we can use Lemma~\ref{lemma_quadratic_plus_fstar_bound} to obtain a tight lower bound for $\fnclas$, which is based on uniform convergence of $\fnclas$ to its expectation in Proposition \ref{prop:unifConv}, and relax the constraint from definition of the set $\setNoisyZero$ as follows:
\begin{align}
\label{eq:second_iteration_relaxed_constraint_f}
    \frac{1}{n}  (2\sqrt{\smax}\etas + b\norm{\hgaussian}_{\infty})^2 
    \geq
    \fnclas(\nu,\sqrt{b^2\|\gamma(\alpha)\|_2^2+\etas^2}) 
    &\geq
    f ( \nu, \sqrt{b^2\norm{\gamma(\alpha)}_2^2+\etas^2}) - \constconc \\
    &\geq
    \fstar  + \tconstnu \deltanu^2 + \tconsteta b^2\norm{\gamma(\alpha)}_2^2 + \tconsteta \etas^2 - \constconc, 
\end{align}
where we choose $\constconc =  O \left( \frac{1}{\tsoptn^3} \right)$ and hence the bound holds uniformly with probability at least $1 - \exp \left( -c_2  \frac{n}{\tsoptn^6}  \right) -  \exp \left( -c_3 \frac{ n  }{\tsoptn^3 \log n} \right)$. 

Now we show how we can relax and simplify the LHS from Equation~\eqref{eq:second_iteration_relaxed_constraint_f}. Since $c_1 \soptn \leq d$, we have according to Equation~\eqref{eq:second_iteration_relaxed_constraint_f} that $\frac{1}{n} (2\sqrt{\smax}\etas + b\norm{\hgaussian}_{\infty})^2 \geq \frac{1}{2}\fstar$. As before, we also have that $\frac{1}{n}(2\sqrt{\smax}\etas + b\norm{\hgaussian}_{\infty})^2 \leq \frac{8\smax}{n}\etas^2 + \frac{2}{n}b^2\norm{\hgaussian}_{\infty}^2 $. Combining last two expressions with the bound $\etas \leq B_{\etas,1}$ from Step 1.3 we have:
\begin{align}
    \frac{1}{n}b^2\norm{\hgaussian}_{\infty}^2 \geq \frac{1}{4}\fstar - \frac{4\smax}{n}B_{\etas,1}^2 \geq \frac{1}{8}\fstar
\end{align}
for $n,d$ large enough since $\smax = \Theta\left(\frac{n}{\tsoptn^8}\right)$. Thus we have:
\ifaistats
\begin{align}
    \frac{1}{n}  (2\sqrt{\smax}\etas + b\norm{\hgaussian}_{\infty})^2 
    &=
    \frac{1}{n}b^2\norm{\hgaussian}_{\infty}^2 \left( 1 + \frac{2\sqrt{\smax}\etas}{b\norm{\hgaussian}_{\infty}} \right)^2
    \leq
    \frac{1}{n}b^2\norm{\hgaussian}_{\infty}^2 \left( 1 + 2B_{\etas,1} \sqrt{\frac{8}{\fstar}} \sqrt{\frac{\smax}{n}} \right)^2
\end{align}
\else
\begin{align}
    \frac{1}{n}  (2\sqrt{\smax}\etas + b\norm{\hgaussian}_{\infty})^2 
    &=
    \frac{1}{n}b^2\norm{\hgaussian}_{\infty}^2 \left( 1 + \frac{2\sqrt{\smax}\etas}{b\norm{\hgaussian}_{\infty}} \right)^2\\
    &\leq
    \frac{1}{n}b^2\norm{\hgaussian}_{\infty}^2 \left( 1 + 2B_{\etas,1} \sqrt{\frac{8}{\fstar}} \sqrt{\frac{\smax}{n}} \right)^2
\end{align}
\fi
and $ \frac{1}{n}  (2\sqrt{\smax}\etas + b\norm{\hgaussian}_{\infty})^2  \leq \frac{1}{n}b^2\norm{\hgaussian}_{\infty}^2 \left( 1 + c\sqrt{\frac{\smax}{n}} \right)$ for a large enough constant $c>0$. Furthermore, define $\OOO_b = c\sqrt{\frac{\smax}{n}} = \Theta \left( \frac{1}{\tsoptn^4} \right)$.

Motivated by Equation~\eqref{eq:second_iteration_relaxed_constraint_f} and discussion after it, we define the following set:
\ifaistats
\begin{align}
    \Gamma_2 = \left\{ (\nu, b, \alpha, \etas) 
    \subjto 
    \frac{ 1 }{ n } b^2 \norm{\hgaussian}_{\infty}^2 \left(1+\OOO_b\right) \geq \fstar  + \tconstnu \deltanu^2 + \tconsteta b^2\norm{\gamma(\alpha)}_2^2 + \tconsteta \etas^2 - \constconc
    \andtxts
    b\alpha \leq \loneboundclas
    \right\} 
\end{align}
\else
\begin{align}
    \Gamma_2 = \bigg\{ (\nu, b, \alpha, \etas) 
    \subjto 
    \frac{ 1 }{ n } b^2 \norm{\hgaussian}_{\infty}^2 \left(1+\OOO_b\right) \geq \fstar  + \tconstnu \deltanu^2 +\tconsteta b^2\norm{\gamma(\alpha)}_2^2 &+ \tconsteta \etas^2 - \constconc\\
    &\andtxts
    b\alpha \leq \loneboundclas
    \bigg\} 
\end{align}
\fi
Again, from the discussion in this section, we have that with high probability $\setNoisyZero \subset \Gamma_2$. Similarly as in the previous bound, we will bound variables of interest i.e. $\etasc,\nu,\etas$ in the set $\Gamma_2$ and use inclusion of set $\setNoisyZero$ into $\Gamma_2$ to claim that these bounds are valid even in $\setNoisyZero$. 

\paragraph{Step 2.1: Upper bound on $\etasc$.} 

Similarly to the Equation~\eqref{eq:l1proofstep1bound} in Step 1.1, we relax constraints of $\Gamma_2$ to obtain:
\begin{align}
    \etasc^2 &\leq \max_{(\nu,b,\alpha,\etas)\in\Gamma_2} b^2 \norm{\gamma(\alpha)}_2^2 \\
    &\leq
    \max_{b,\alpha} \left[b^2\norm{\gamma(\alpha)}_2^2 \subjto \frac{ 1 }{ n } b^2 \norm{\hgaussian}_{\infty}^2 \left(1+\OOO_b\right) \geq \fstar + \tconsteta b^2\norm{\gamma(\alpha)}_2^2  - \constconc
    \andtxts
    b\alpha \leq \loneboundclas \right]\\
    &\leq
    \max_{b,\alpha}\left[  b^2\norm{\gamma(\alpha)}_2^2 \subjto b^2 \geq (\fstar - \constconc) \left( \frac{1}{n}\norm{\hgaussian}_{\infty}^2 \left(1+\OOO_b\right) - \tconsteta \norm{\gamma(\alpha)}_2^2 \right)^{-1}
    \andtxts
    b \leq \frac{\loneboundclas}{\alpha} \right]\\
    &=  \max_{\alpha} \left[ \frac{\loneboundclas^2}{\alpha^2} \norm{\gamma(\alpha)}_2^2 \subjto \frac{ 1 }{ n } \frac{\loneboundclas^2}{\alpha^2} \norm{\hgaussian}_{\infty}^2(1+\OOO_b) \geq \fstar +  \tconsteta \frac{\loneboundclas^2}{\alpha^2} \norm{\gamma(\alpha)}_2^2 - \constconc \right].
    \label{eq:l1proofstep2eta}
\end{align}
Multiplying the constraint on both sides with $\alpha^2$ and using the fact that $\norm{\gamma(\alpha)}_2^2$ is convex shows that the set of feasible $\alpha$ is again a (non-empty) interval. 
Thus, by the monotonicity of $ \frac{\norm{\gamma(\alpha)}_2^2}{\alpha^2} $ the problem reduces again to finding $\alphauls < \alpha_{\soptn}$ (where we use again that $\alpha_{\soptn}$ satisfies the constraints with high probability) such that $\alphauls$ violates the constraint in Equation~\eqref{eq:l1proofstep2eta}, i.e., 
\begin{align}
\label{eq:l1proofstep2eta2}
    \frac{ \fstar - \constconc}{1+\OOO_b} \frac{ n \alphauls^2 }{  \loneboundclas^2 \norm{\hgaussian}_{\infty}^2 } + \frac{\tconsteta}{1+\OOO_b} n \frac{ \norm{ \gamma(\alphauls) }_2^2 }{ \norm{\hgaussian}_{\infty}^2 } > 1
\end{align}

We now show that we can choose $ \ulm = \lambda_{\ulm} \soptn$ with universal constant $\lambda_{\ulm} \in (0,1)$. Indeed, applying  Proposition \ref{prop:concentration_gammas} and using the characterization $t_m^2 =   2 \log(d/m) - \log \log (d/m) -\log(\pi) +\frac{\log\log(d/m)}{2\log(d/m)} + O\left(\frac{1}{\log(d/m)}\right)$ from Section~\ref{subsubsec:prop_gamma_alpha} we get:
\begin{align}
 \frac{\fstar - \constconc}{1+\OOO_b}\frac{ n \alphauls^2 }{  \loneboundclas^2 \norm{\hgaussian}_{\infty}^2 } &= 1 + \frac{2\log \lambda_{\ulm} -2}{\tsoptn^2} + O\left(\frac{1}{\tsoptn^3}\right) \\
 \mathrm{and}~~\frac{\tconsteta}{1+\OOO_b} n \frac{ \norm{ \gamma(\alphauls) }_2^2 }{ \norm{\hgaussian}_{\infty}^2 } &= 
\frac{1}{\tsoptn^2}\frac{4\tconsteta}{\feestar\lambda_{\ulm}}  +  O \left( \frac{1}{ \tsoptn^4} \right)
\end{align}
where $O(.)$ has hidden dependencies on $\lambda_{\ulm}$. Hence, it is straight forward to see that for any $d \geq c n$ with universal constant $c>0$ (and thus $\tsoptn$ lower bounded), we can find a universal constant $\lambda_{\ulm}$ such that Equation~\eqref{eq:l1proofstep2eta2} holds. 

Hence, we can upper bound $\etasc^2$ in Equation \eqref{eq:l1proofstep2eta} as follows:
\begin{align} \label{eq:eta_bound_second_reduction}
    \etasc^2 
    \leq
    \loneboundclas^2 \frac{ \norm { \gamma ( \alphauls ) }_2^2 }{ \alpha_{\ulm}^2 } 
    \leq
    \frac{ n \fstar }{ \tsoptn^2 } \frac{2}{\ulm}\left( 1 + O \left( \frac{1}{\tulm^2} \right) \right) 
    \leq
    \frac{2\fstar}{\feestar \lambda_{\ulm} \log (d/n) } \left( 1 + O \left( \frac{1}{ \log (d/n) } \right) \right) 
    =:
    \frac{B_{\etasc,2}^2}{\tsoptn^2}
\end{align}
with $B_{\etasc,2}^2 = \Theta(1)$.

\paragraph{Step 2.2: Upper bound on $\deltanu$.}
Instead of directly bounding $\nu$, here we upper bound $\deltanu^2$ with $\nu = \nubar +\deltanu$ and thus obtain both an upper and lower bound for $\nu$. 

Similarly as before, we have:
\ifaistats
\begin{align}
    &\deltanu^2 \leq \max_{(\nu,b,\alpha,\etas)\in\Gamma_2} \deltanu^2
    \leq
    \max_{\nu,b,\alpha} \bigg[\deltanu^2
    \subjto \frac{ 1 }{ n } b^2 \norm{\hgaussian}_{\infty}^2 \left(1+\OOO_b\right) \geq \fstar + \tconstnu \deltanu^2 - \constconc\\
    & \hphantom{{} \leq \max_{(\nu,b,\alpha,\etas)\in\Gamma_2} \deltanu^2
    \leq
    \bigg[ \max_{\nu,b,\alpha} \deltanu^2
    \subjto}\andtxts
    \frac{ 1 }{ n } b^2 \norm{\hgaussian}_{\infty}^2 \left(1+\OOO_b\right) \geq \fstar + \tconsteta b^2\norm{\gamma(\alpha)}_2^2  - \constconc 
    \andtxts
    b\alpha \leq \loneboundclas
    \bigg]\\
    &= \max_{b,\alpha} \left[ \frac{1}{\tconstnu} \left( \frac{1}{n}b^2\norm{\hgaussian}_{\infty}^2(1+\OOO_b) - \fstar +\constconc \right) \subjto \frac{ 1 }{ n } b^2 \norm{\hgaussian}_{\infty}^2 \left(1+\OOO_b\right) \geq \fstar + \tconsteta b^2\norm{\gamma(\alpha)}_2^2  - \constconc 
    \andtxts
    b\alpha \leq \loneboundclas \right]\\
    &= \max_{\alpha} \left[ \frac{1}{\tconstnu} \left( \frac{1}{n}\frac{\loneboundclas^2}{\alpha^2}\norm{\hgaussian}_{\infty}^2(1+\OOO_b) - \fstar +\constconc \right) \subjto \frac{ 1 }{ n } \frac{\loneboundclas^2}{\alpha^2} \norm{\hgaussian}_{\infty}^2 \left(1+\OOO_b\right) \geq \fstar + \tconsteta \frac{\loneboundclas^2}{\alpha^2}\norm{\gamma(\alpha)}_2^2  - \constconc  \right]
    \label{eq:unicon_noisy_deltanu_step22}
\end{align}
\else
\begin{align}
    \deltanu^2 &\leq \max_{(\nu,b,\alpha,\etas)\in\Gamma_2} \deltanu^2
    \leq
    \max_{\nu,b,\alpha} \bigg[\deltanu^2
    \subjto \frac{ 1 }{ n } b^2 \norm{\hgaussian}_{\infty}^2 \left(1+\OOO_b\right) \geq \fstar + \tconstnu \deltanu^2 - \constconc\\
    & \hphantom{{} \leq \max_{(\nu,b,\alpha,\etas)\in\Gamma_2} \deltanu^2}\andtxts
    \frac{ 1 }{ n } b^2 \norm{\hgaussian}_{\infty}^2 \left(1+\OOO_b\right) \geq \fstar + \tconsteta b^2\norm{\gamma(\alpha)}_2^2  - \constconc 
    \andtxts
    b\alpha \leq \loneboundclas
    \bigg]\\
    &= \max_{b,\alpha} \bigg[ \frac{1}{\tconstnu} \left( \frac{1}{n}b^2\norm{\hgaussian}_{\infty}^2(1+\OOO_b) - \fstar +\constconc \right) \\
    & \hphantom{{}   \frac{1}{\tconstnu} ( \frac{1}{n}b^2\norm{\hgaussian}_{\infty}^2(1+\OOO_b) }
    \subjto \frac{ 1 }{ n } b^2 \norm{\hgaussian}_{\infty}^2 \left(1+\OOO_b\right) \geq \fstar + \tconsteta b^2\norm{\gamma(\alpha)}_2^2  - \constconc 
    \andtxts
    b\alpha \leq \loneboundclas \bigg]\\
    &= \max_{\alpha} \bigg[ \frac{1}{\tconstnu} \left( \frac{1}{n}\frac{\loneboundclas^2}{\alpha^2}\norm{\hgaussian}_{\infty}^2(1+\OOO_b) - \fstar +\constconc \right) \\
    &\hphantom{{} \frac{1}{\tconstnu} ( \frac{1}{n}b^2\norm{\hgaussian}_{\infty}^2(1+\OOO_b) }
    \subjto \frac{ 1 }{ n } \frac{\loneboundclas^2}{\alpha^2} \norm{\hgaussian}_{\infty}^2 \left(1+\OOO_b\right) \geq \fstar + \tconsteta \frac{\loneboundclas^2}{\alpha^2}\norm{\gamma(\alpha)}_2^2  - \constconc  \bigg]
    \label{eq:unicon_noisy_deltanu_step22}
\end{align}
\fi

As in Step 1.2 we use that $\frac{1}{\alpha^2}$ is monotonically decreasing function and the fact that $\alphauls$ from previous step, with $ \ulm = \lambda_{\ulm} \soptn$ and $\alphauls \leq \alpha_{\soptn}$, does not satisfy constraint in Equation~\eqref{eq:unicon_noisy_deltanu_step22}. Thus we can upper bound $\deltanu^2$ as follows:
\begin{align}
    \deltanu^2 
    &\leq
    \frac{1}{ n \tconstnu }  \frac{ \loneboundclas^2}{\alphauls^2} \norm{\hgaussian}_{\infty}^2(1+\OOO_b) - \frac{\fstar}{\tconstnu} + \frac{\constconc}{\tconstnu}
    \leq 
    \frac{ \fstar t_{\ulm}^2}{\tsoptn^2 \tconstnu } \left( 1 + \frac{2}{\tsoptn^2}  \right)
    - \frac{\fstar}{\tconstnu} + O \left( \frac{1}{\tsoptn^3} \right)\\
    & = \frac{\fstar(2 -2\log(\lambda_{\ulm})) }{\tconstnu 2\log(d/n)}\left( 1 + O \left( \frac{1}{\log(d/n)} \right) \right)
    =: 
    \frac{B_{\deltanu,2}^2}{\tsoptn^2} 
\end{align}
for some $B_{\deltanu,2}^2 = \Theta(1)$.
\paragraph{Step 2.3: Upper bound on $\etas$.}
Following the same steps as in Step 2.2 with $\nu$ replaced by $\etas$ we can show that there exists universal constant $B_{\etas,2}=\Theta(1)$ such that:
\begin{align}
    \etas^2 \leq \frac{1}{ n \tconsteta }  \frac{ \loneboundclas^2}{\alphauls^2} \norm{\hgaussian}_{\infty}^2 (1+\OOO_b) - \frac{\fstar}{\tconsteta} + \frac{\constconc}{\tconsteta}
    \leq 
    \frac{\fstar(2 -2\log(\lambda_{\ulm})) }{\tconsteta 2\log(d/n)}\left( 1 + O \left( \frac{1}{\log(d/n)} \right) \right)
    =: 
    \frac{B_{\etas,2}^2}{\tsoptn^2} 
\end{align}

\subsubsection*{Bound 3: Proof of the proposition} 
We already know that $\nu$ concentrates around $\nubar$. However, in order to get a tight expression for the risk and also a valid lower bound, we need to get tighter bounds for $\etasc^2$ and $\etas^2$ conditioning on the bounds from the previous step, leading to Proposition~\ref{prop:BoundGamma0_class_noisy}.

Note that $\fclas$ is infinitely differentiable function as we prove in Lemma~\ref{lemma:fclas_fclasn_inf_diff}. Thus, in this part of the proof we can use Taylor series approximation of the function $\fclas$ where we use the result from the last step to bound the higher order terms involving $\deltanu,\etasc$ and $\etas$. Similarly as in equation \eqref{eq:second_iteration_relaxed_constraint_f}, we obtain from Proposition~\ref{prop:unifConv} and the second order Taylor series approximation of $\fclas$ around the point $(\nubar, 0)$ that with high probability,
\begin{align}
    \label{eq:f_taylor_approx_ssparse}
    \frac{1}{n} b^2 \norm{\hgaussian}_{\infty}^2 (1+\OOO_b)
    \geq 
    \fstar  + \frac{1}{2}\fnnstar \deltanu^2 + \frac{1}{2}\feestar b^2 \norm{\gamma(\alpha)}_2^2 + \frac{1}{2}\feestar \etas^2 - \constconc - \otaylor
\end{align}
with $\otaylor = O(\deltanu^3 + \etasc^3 + \etas^3) =  O \left( \frac{1}{\tsoptn^3} \right)$ and $\constconc,\OOO_b = O \left( \frac{1}{\tsoptn^3} \right)$. 

\paragraph{Step 3.1: Upper and lower bound on $\etasc$.}
We proceed in the same manner as in the previous two steps. We relax the constraint in definition of $\setNoisyZero$ and define the following set:
\begin{align}
    \Gamma_3^{\etasc} = \left\{ (\nu, b, \alpha, \etas) 
    \subjto 
    \frac{ 1 }{ n } b^2 \norm{\hgaussian}_{\infty}^2 (1+\OOO_b) \geq \fstar +  \frac{1}{2}\feestar  b^2 \norm{\gamma(\alpha)}_2^2  -\constconc - \otaylor
    \andtxts
    b\alpha \leq \loneboundclas
    \right\}
\end{align}
Clearly, we have again with high probability that $\setNoisyZero \subset \Gamma_3^{\etasc}$. The only difference between $\Gamma_3^{\etasc}$ and $\Gamma_2$ lies in the constant $\tconsteta$ which is replaced by the tighter constant $\feestar/2$. However, this makes a big difference, as this allows us to choose $\ulm < \soptn <\olm$ much tighter. Similar to Equation~\eqref{eq:l1proofstep2eta2} we again require that $m=\ulm,\olm$ satisfies
\begin{align}
\label{eq:l1thirdeta_ssparse}
    \frac{ \fstar - \constconc - \otaylor}{1+\OOO_b} \frac{ \alpha_m^2 }{ \norm{\hgaussian}_{\infty}^2 } \frac{ n }{ \loneboundclas^2 } 
    +
    \frac{\feestar}{2(1+\OOO_b)} n \frac{ \norm{ \gamma (\alpha_m) }_2^2 }{ \norm{\hgaussian}_{\infty}^2 } 
    > 1.
\end{align}
However, this expression allows us to choose $\ulm$ and $\olm$ as in Lemma~\ref{lemma_uls_ols_exp}, with $\kappa := 1/2$, $\mstar:=\soptn$ and parameter $\lambda>0$. We only show it for $\ulm$ as the same argument holds for $\olm$. Applying Proposition~\ref{prop:concentration_gammas}, the LHS from Equation~\eqref{eq:l1thirdeta_ssparse} can be bounded by
\begin{align}
    \frac{\fstar - \constconc - \otaylor}{1+\OOO_b} \frac{ \alphauls^2 }{ \norm{\hgaussian}_{\infty}^2 } \frac{ n }{ \loneboundclas^2 } 
    &= \frac{\tsoptn^2}{t_{\ulm}^2}\left(1 -\frac{4}{t_{\ulm}^2}+\frac{2}{\tsoptn^2} + O\left(\frac{1}{\tsoptn^3}\right)\right)  \\
    &= 1- \frac{\lambda}{\tsoptn^{5/2}} -\frac{2}{\tsoptn^2} + O\left(\frac{1}{\tsoptn^3}\right) + O\left(\frac{1}{\tsoptn^2 \soptn}\right)
    \\
    ~~\mathrm{and}\:\: \frac{\feestar}{2(1+\OOO_b)}  n \frac{ \norm{ \gamma (\alphauls) }_2^2 }{ \norm{\hgaussian}_{\infty}^2 } &= 
    \frac{2}{\tsoptn^2} + \frac{\lambda}{\tsoptn^{5/2}} + \frac{\lambda^2}{4\tsoptn^3} + O \left( \frac{1}{\tsoptn^3} \right) + O \left( \frac{1}{\tsoptn^2 \soptn} \right)
\end{align}
with $O(.)$ having hidden dependencies on universal constant $\lambda$. In particular, as a result, we see that we can choose $\lambda$ such that Equation~\eqref{eq:l1thirdeta_ssparse} hold for any $d >c n$ with universal constant $c>0$. 
Hence we can upper bound $\etasc^2$ as follows:
\begin{align} \label{eq:eta_bound_third_reduction_ssparse}
    \etasc^2 
    \leq
    \loneboundclas^2 \frac{ \norm { \gamma ( \alphauls ) }_2^2 }{ \alphauls^2 } 
    & \leq
    \frac{ n \fstar }{ \tsoptn^2 } \frac{2}{\ulm}\left( 1 + O\left( \frac{1}{\tsoptn^2} \right) \right)
    \leq
    \frac{4\fstar}{\feestar} \frac{1}{\tsoptn^2} \left( 1 + \frac{\lambda}{2\sqrt{\tsoptn}} + O \left( \frac{1}{\tsoptn} \right) \right) 
\end{align}
Furthermore, we also obtain a lower bound for $\etasc^2$. Similar as in Lemma~5/6 \cite{wang2021tight}, we can lower bound (using again the monotonicity of $\frac{\norm{\gamma(\alpha)}_2}{\alpha}$ and the fact that any feasible $\alpha \leq \alphaols$)
\begin{align}
    \etasc^2 
    &\geq
    \min_b \left[b^2 \norm{\gamma(\alpha_{\olm})}_2^2 \subjto b^2  
    \geq \frac{\fstar   -\constconc - \otaylor}{\frac{  \norm{\hgaussian}_{\infty}^2}{ n }(1+\OOO_b) -\frac{1}{2}\feestar  \norm{\gamma(\alpha_{\olm})}_2^2} \right] \\
    &=
    \frac{\fstar  -\constconc - \otaylor}{\frac{  \norm{\hgaussian}_{\infty}^2}{ n } (1+\OOO_b)-\frac{1}{2}\feestar  \norm{\gamma(\alpha_{\olm})}_2^2} \norm{\gamma(\alpha_{\olm})}_2^2 
    \geq  \frac{4\fstar}{\feestar} \frac{1}{\tsoptn^2} \left( 1 - \frac{\lambda}{2\sqrt{\tsoptn}} + O \left( \frac{1}{\tsoptn} \right) \right) 
\end{align}
\paragraph{Step 3.2: Upper bound on $\etas$.} In order to upper bound $\etas$ we further constrain $\Gamma_3^{\etasc}$ and define a set:
\ifaistats
\begin{align}
    \Gamma_3^{\etas} = \left\{ (\nu, b, \alpha, \etas) 
    \subjto 
    \frac{ 1 }{ n } b^2 \norm{\hgaussian}_{\infty}^2 (1+\OOO_b) \geq \fstar +  \frac{1}{2}\feestar  b^2 \norm{\gamma(\alpha)}_2^2 + \frac{1}{2} \fssstar \etas^2 -\constconc - \otaylor
    \andtxts
    b\alpha \leq \loneboundclas
    \right\}
\end{align}
\else
\begin{align}
    \Gamma_3^{\etas} = \bigg\{ (\nu, b, \alpha, \etas) 
    \subjto 
    \frac{ 1 }{ n } b^2 \norm{\hgaussian}_{\infty}^2 (1+\OOO_b) \geq \fstar +  \frac{1}{2}\feestar  b^2 \norm{\gamma(\alpha)}_2^2 +& \frac{1}{2} \fssstar \etas^2 -\constconc - \otaylor\\
    &\andtxts
    b\alpha \leq \loneboundclas
    \bigg\}
\end{align}
\fi
Note that $\Gamma_3^{\etas} \subset \Gamma_3^{\etasc}$ and thus we can use bounds $\ulm,\olm$ from the previous part. Upper bounding $\etas$ by other variables from the first constraint of $\Gamma_3^{\etas}$ and using that $\frac{1}{\alpha^2}$ and $-\frac{\norm{\gamma(\alpha)}_2^2}{\alpha^2}$ are monotonically decreasing and increasing in $\alpha$, respectively, we obtain the following high probability bound:
\begin{align}
    \etas^2 &\leq \frac{2}{\feestar} \left( \frac{\loneboundclas^2}{n} \left( \frac{\norm{\hgaussian}_{\infty}^2}{\alphauls^2}(1+\OOO_b) - \frac{1}{2}\feestar n \frac{\norm{\gamma(\alphaols)}_2^2}{\alphaols^2} \right)  - \fstar + \constconc + \otaylor \right)\\
    &=
    \frac{2\fstar}{\feestar} \left[ \frac{1}{\tsoptn^2} \left(1-\frac{2}{\tsoptn^2} + \frac{\tc}{\tsoptn^3} \right) \left( \tulm^2 \left(1+\frac{4}{\tulm^2} + \frac{c_2}{\tsoptn^3} \right) -  \frac{2\soptn}{\olm} \left( 1+\frac{c_3}{\tolm^2} \right) \right) - 1 \right]  + O\left(\frac{1}{\tsoptn^3} \right)
\end{align}
where the second line follows again from concentration results from Proposition~\ref{prop:concentration_gammas}. Multiplying all the terms gives $\etas^2 \lesssim \frac{1}{\tsoptn^{5/2}}$, as we wanted to show.

Note that we could prove in the exact same way that $\deltanu^2 = O\left(\frac{1}{\tsoptn^{5/2}} \right)$, but this does not change tightness of our result in Theorem~\ref{thm:mainl1_class_noisy} and hence we skip this step and conclude the proof of Proposition~\ref{prop:BoundGamma0_class_noisy}.

\section{Technical Lemmas}

\subsection{Application of CGMT: Proof of Proposition~\ref{prop:CGMT_application_classification}}
\label{sub:proof_CGMT_technical}
The proof essentially follows exactly the same steps as in \cite{koehler2021uniform} and \citesmart{donhauser2022fast} except for a few simple modifications which we describe next. First we introduce a more general form of  the (C)GMT:
\begin{lemma}
\label{lemma:CGMT_variant}
Let $X_1 \in \RR^{n\times {d-s}}$ be a matrix with i.i.d. $\NNN(0,1)$ entries and let $\ggaussian \sim \NNN(0,I_n)$ and $\hgaussian \sim \NNN(0,I_{d-s})$ be independent random vectors. Let $\compactw \subset \RR^{s}\times \RR^{d-s}$ and $\compactv \subset \RR^{n}$ be compact sets, and let $\psi: \compactw \times \compactv \to \RR$ be a continuous function. Then for the following two optimization problems:
    \begin{equation}
        \Phi = \min_{(w_1,w_2)\in \compactw}\max_{v\in \compactv} \innerprod{v}{X_1 w_1} + \psifcn((w_1,w_2),v)
    \end{equation}
    \begin{equation}
        \phi = \min_{(w_1,w_2)\in \compactw}\max_{v \in \compactv} \norm{w_1}_2\innerprod{v}{\ggaussian} + \norm{v}_2\innerprod{w_1}{\hgaussian}+\psifcn((w_1,w_2),v)
    \end{equation}
    and any $t\in\RR$ holds that:
    \begin{equation}
        \prob(\Phi < t)\leq 2\prob(\phi \leq t) 
    \end{equation}
    
    If in addition $\psi$ is convex-concave function we also have for any $t\in\RR$:
    \begin{equation}
        \prob(\Phi > t)\leq 2\prob(\phi \geq t) 
    \end{equation}
    In both inequalities the probabilities on the LHS and RHS are over the draws of $X_1$, and of $\ggaussian$, $\hgaussian$, respectively.
\end{lemma}
\begin{proof}
The first part of the lemma is equivalent to Theorem 10 in \cite{koehler2021uniform}. The proof of the second part follows from Theorem 9 in \cite{koehler2021uniform} and proof of Theorem 10 in \cite{koehler2021uniform}.
\end{proof}

In order to apply Lemma \ref{lemma:CGMT_variant} we first rewrite $\PhiCNuni$ using the Lagrange multipliers $\lmult\in\RR^n$ as follows:
\ifaistats
\begin{align}
    \PhiCNuni
    = \min_{w} \max_{\lmult\geq 0} \norm{w}_1 + \innerprod{\lmult}{\onevec - \matDy\matX w}
    = \min_{(\wpar,\wperp)} \max_{\lmult\geq 0} \normk{\wpar+\wperp}_1 + \innerprod{\lmult}{\onevec - \matDy\matXpar \wpar} - \innerprod{\lmult}{\matDy \matXperp \wperp}
\end{align}
\else
\begin{align}
    \PhiCNuni
    &= \min_{w} \max_{\lmult\geq 0} \norm{w}_1 + \innerprod{\lmult}{\onevec - \matDy\matX w}\\
    &= \min_{(\wpar,\wperp)} \max_{\lmult\geq 0} \normk{\wpar^d+\wperp}_1 + \innerprod{\lmult}{\onevec - \matDy\matXpar \wpar} - \innerprod{\lmult}{\matDy \matXperp \wperp}
\end{align}
\fi
where $\matDy = \mathrm{diag}(y_1,y_2,\dots,y_n)$. Since $\matDy$ and $\matXperp$ are independent we note that $\matDy\matXperp \in \RR^{n\times d}$ has i.i.d. entries distributed according to the standard normal distribution and hence $\matDy\matXperp\stackrel{d}{=}\matXperp$ with $\stackrel{d}{=}$ denoting equivalence of random variables in distribution. Comparing obtained expression with definition of $\Phi$ from Lemma~\ref{lemma:CGMT_variant}, it is obvious that we should take $X_1:=\matXperp, w_1:=\wperp, w_2:=\wpar$ and function $\psi(w,\lmult) := \normk{\wpar+\wperp}_1 + \innerprod{\lmult}{ \onevec - \matDy \matX \wpar}$, which is a continuous convex-concave function on the whole domain since every norm is a convex function.
Motivated by expression for $\phi$ from Lemma~\ref{lemma:CGMT_variant}, we further define
\begin{align}
    \tPhiCdnuni &:= \min_{(\wpar,\wperp)} \max_{\lmult\geq 0} \normk{\wpar+\wperp}_1 + \innerprod{\lmult}{\onevec - \matDy\matXpar \wpar} - \normk{\wperp}_2 \innerprod{\lmult}{ \ggaussian} - \normk{\lmult}_2 \innerprod{\wperp}{\hgaussian}\\
    &= \min_{(\wpar,\wperp)} \max_{\lambda\geq 0} \normk{\wpar+\wperp}_1 - \lambda \left(\innerprod{\wperp}{\hgaussian}-\norm{\pos{\onevec - \matDy\matXpar \wpar-  \ggaussian\normk{\wperp}_2}}_2 \right) \\
    & = \min_{(\wpar,\wperp)} \normk{\wpar+\wperp}_1
    \subjto \innerprod{\wperp}{\hgaussian}\geq \norm{\pos{\onevec - \matDy\matXpar \wpar-  \ggaussian\norm{\wperp}_2}}_2
\end{align}
where in the second equality we set $\lambda:= \norm{\lmult}_2$. Define $\wperps = \Pi_{\SSS} \wperp, \wperpsc = \Pi_{\SSS^c} \wperp$ where $\Pi_{\SSS}$ and $\Pi_{\SSS^c}$ are projections on $\supp(\wgt)$ and the other $d-s$ entries, respectively. So we can rewrite $\tPhiCdnuni$ as:
\begin{align}
    \tPhiCdnuni = \min_{(\wpar,\wperps,\wperpsc)} &\|\wpar+\wperps\|_1+\|\wperpsc\|_1\\
    &\subjto \langle\wperps,\hone \rangle + \langle\wperpsc,\htwo \rangle \geq \normk{(\onevec - \matDy\matXpar \wpar-  \ggaussian\sqrt{\|\wperps\|_2^2+\|\wperpsc\|_2^2}))_+}_2
\end{align}
with $\hone\sim\NNN(0,I_s)$ and $\htwo\sim\NNN(0,I_{d-s})$, independent of each other. Under constraint that $\langle\wperps,\hone \rangle + \langle\wperpsc,\htwo \rangle \geq 0$ we can square the last inequality and scale with $\frac{1}{n}$ to obtain the following RHS:
\ifaistats
    \begin{align}
    \frac{1}{n} \|(\onevec - \matDy\matXpar \wpar-  \ggaussian\sqrt{\|\wperps\|_2^2+\|\wperpsc\|_2^2})_+ \|_2^2 = \frac{1}{n} \sumin \possq{1-\xi_i\sgn(\langle (\xpar)_i, \wgtp \rangle) \langle (\xpar)_i, \wpar \rangle -\ggaussian_i \normk{\wperp}_2} , 
    \end{align}
\else
    \begin{align}
    \frac{1}{n} \|(\onevec - \matDy\matXpar \wpar- & \ggaussian\sqrt{\|\wperps\|_2^2+\|\wperpsc\|_2^2})_+ \|_2^2 \\
    &=
    \frac{1}{n} \sumin \possq{1-\xi_i\sgn(\langle (\xpar)_i, \wgtp \rangle) \langle (\xpar)_i, \wpar \rangle -\ggaussian_i \normk{\wperp}_2} , 
    \end{align}
\fi
which is exactly the function $\fnclasuni(\langle \wpar, \wgt \rangle,\norm{\wperp}_2)$, as defined in Equation \eqref{eq:fnclas_def_noisy_par}. Hence, comparing with the expression for $\PhiCdnuni$ from Proposition~\ref{prop:CGMT_application_classification} we note that $\tPhiCdnuni \equiv \PhiCdnuni$.

In order to complete the proof of the proposition, we need to discuss
compactness of the feasible sets in the optimization problem 
so that we can apply Lemma~\ref{lemma:CGMT_variant} to $\PhiCNuni$ and $\PhiCdnuni$.
For this purpose, we define the following truncated optimization problems $\PhiCNuni^r(t)$ and $\PhiCdnuni^r(t)$ for some $r,t \geq 0$:
\ifaistats
\begin{gather}
    \PhiCNuni^r(t) := \min_{\norm{w}_1\leq t} \max_{\substack{\norm{v}\leq r\\ v\geq 0}} \norm{w}_1 + \innerprod{v}{1-D_y X w}
    \\
    \PhiCdnuni^r(t) := \min_{\|\wpar+\wperps\|_1+\|\wperpsc\|_1 \leq t} \max_{0\leq \lambda\leq n r} \|\wpar+\wperps\|_1 + \|\wperpsc\|_1 - \lambda \left(  \frac{1}{n}(\langle\wperps,\hone \rangle + \langle\wperpsc,\htwo \rangle) - \sqrt{\fnclasuni(w)} \right).
\end{gather}
\else
\begin{gather}
    \PhiCNuni^r(t) := \min_{\norm{w}_1\leq t} \max_{\substack{\norm{v}\leq r\\ v\geq 0}} \norm{w}_1 + \innerprod{v}{1-D_y X w}
\end{gather}
\begin{align}
    \PhiCdnuni^r(t) := \min_{\|\wpar+\wperps\|_1+\|\wperpsc\|_1 \leq t} \max_{0\leq \lambda\leq n r} &\|\wpar+\wperps\|_1 + \|\wperpsc\|_1 \\
    &- \lambda \left(  \frac{1}{n}(\langle\wperps,\hone \rangle + \langle\wperpsc,\htwo \rangle) - \sqrt{\fnclasuni(w)} \right).
\end{align}
\fi
By definition it follows that $\PhiCdnuni^{r_1}(t) \geq \PhiCdnuni^{r_2}(t)$ for any $r_1\geq r_2$, and thus we have that  
\begin{equation}
    \PP (\PhiCdnuni \geq t | \xi) \geq \lim_{r\to\infty} \PP (\PhiCdnuni^r(t) \geq t | \xi).
    \label{eq:PhiCdnuni_lim_r_infty}
\end{equation}
Furthermore, by making use of the simple (linear) dependency on $\lambda$ in the optimization objective in the definition of $\PhiCNuni$,  a standard limit argument as in the proof of Lemma 7 in \cite{koehler2021uniform} shows that:
\begin{align}
     \lim_{r\to\infty} \PP (\PhiCNuni^r(t) > t | \xi) = \PP ( \PhiCNuni > t | \xi).
     \label{eq:PhiCNuni_lim_r_infty}
\end{align}
Finally, the proof follows when noting that we can apply Lemma~\ref{lemma:CGMT_variant} directly to $\PhiCNuni^r(t)$ and $\PhiCdnuni^r(t)$ for any $r,t\geq 0$, which gives us $\prob(\PhiCNuni^r>t|\xi) \leq 2\prob(\PhiCdnuni^r\geq t|\xi)$. Combining the last inequality with Equations~\eqref{eq:PhiCdnuni_lim_r_infty} and ~\ref{eq:PhiCNuni_lim_r_infty} completes the proof for $\PhiCNuni$.



The proof for $\PhiCpuni$ and $\PhiCmuni$ uses the same steps as discussed above. We only detail the proof for $\PhiCmuni$ here, as the the proof for $\PhiCpuni$ follows from the exact same reasoning.


Now, let $\loneboundclasuni B_1 = \{w\in\RR^d: \norm{w}_1 \leq \loneboundclasuni \}$ be an $\ell_1$-ball of radius $M$ and note that we optimize over $(\wpar,\wperps,\wperpsc)\in \compactw$ where $\compactw = \{w\subjto \norm{w}_2 \geq \delta\} \cap \loneboundclasuni B_1$ is a compact set. Furthermore, define the function $\psi$ by $\psi(w,v) := \frac{\langle\wpar,\wgtp\rangle }{\norm{w}_2} + \innerprod{v}{\onevec - \matDy \matXpar \wpar}$, which is a continuous function on $\compactw$ since $\norm{w}_2\geq \delta$. Similarly as above, we can overcome the issue with the compactness of the set $\compactv$ by using a truncation argument as proposed in Lemma 4 in \cite{koehler2021uniform}. In particular, we define
\begin{gather}
    \PhiCmuni^r := \min_{w\in \compactw} \max_{\substack{\norm{v}\leq r\\ v\geq 0}} \frac{\langle w, \wgt \rangle}{\norm{w}_2}  + \innerprod{v}{1-D_y X w},
    \\
    \PhiCdmuni^r := \min_{w\in \compactw} \max_{0 \leq \lambda\leq n r} \frac{\langle\wpar,\wgt\rangle }{\norm{w}_2} -  \lambda \left(  \frac{1}{n}(\langle\wperps,\hone \rangle + \langle\wperpsc,\htwo \rangle) - \sqrt{\fnclasuni(w)} \right).
\end{gather}
for which we have
\begin{align}
    \PP(\PhiCmuni < t| \xi) \leq \lim_{r\to \infty} \PP(\PhiCmuni^r < t | \xi)~~~\andtxt~~~ \lim_{r\to\infty} \PP(\PhiCdmuni^r \leq t| \xi) =\PP(\PhiCdmuni \leq t|\xi).
\end{align}
We note that the first statement follows from the definition of $\PhiCmuni$ and the monotonicity of $\PhiCmuni^r$ in $r$,  while the second statement follows from a limit argument as in Lemma 4 in \cite{koehler2021uniform}. Finally, we conclude the proof by applying the first part of Lemma~\ref{lemma:CGMT_variant} to $\PhiCmuni^r$ and $\PhiCdmuni^r$ and defining $\gausswone = \langle X_{\parallel}, \wgt\rangle$ with $X_{\parallel}$ the row-wise projection of $X$ in the subspace spanned by $\wgt$

\subsection{Lower bounds for \texorpdfstring{$\fnclasn$}{fnn} in noiseless setting}
Recall that $\nu= \innerprodk{\wpar}{\wgtp}, \etas = \normk{\wperps}_2, \etasc = \normk{\wperpsc}_2$ and $\eta = \normk{\wperp}_2 = \sqrt{\etas^2+\etasc^2}$. In the noiseless setting we defined the following two functions:
\begin{align}
    \fnclasn (\nu, \eta) &= \frac{1}{n} \sum_{i=1}^n ( 1 - \nu |\gausswone_i| - \gausswtwo_i \eta )_+^2\\
    \fclasn(\nu,\eta) &= \EE  \fnclas \left( \nu, \eta \right) = \EE_{\Gausswone,\Gausswtwo \sim \NNN(0,1)} ( 1 - \nu |\Gausswone| - \Gausswtwo \eta )_+^2 .
\end{align}
In this section we show multiple lower bounds of $\fnclasn$. First we show a bound with non-tight constants, and then we show a tight result based on uniform convergence of $\fnclasn$ to $f$. At the end we give a corollaries of the uniform convergence proposition which is used in the proof of the Proposition~\ref{prop:BoundGamma0_class_noiseless}.
\subsubsection*{Lower bounding \texorpdfstring{$\fnclasn$}{fnn} with non-tight constants }
We show the following proposition:
\begin{proposition}
\label{prop:fnoiselessquadratic}
Assume that $\nu$ satisfies $c_1 \leq \nu \leq \nu_{\max}$ for some universal constant $c_1>0$. There exist universal constants $\kappa_1,\kappa_2, c_2$ such that for any $\nu,\eta$ that satisfy the given assumption, the inequality
\begin{align}
    \fnclasn (\nu,\eta) \geq \kappa_1 \frac{1}{\nu} + \kappa_2 \frac{\eta^2}{\nu}  
\end{align}
holds with probability $\geq 1-2\exp\left(-c_2\frac{n}{(\nu_{\max})^2}\right)$ over the draws of $\gausswone,\gausswtwo$.
\end{proposition}
\begin{proof}

Similarly to the above, we have:
\begin{align}
    \fnclasn (\nu,\eta)  
    &=
    \frac{1}{n}\sum_{i=1}^n ( 1 - \nu |\gausswone_i| - \gausswtwo_i\eta )_+^2
    \geq
    \frac{1}{n}\sum_{i=1}^n ( 1 - \nu |\gausswone_i| + c_1 \eta )_+^2 \indicator{\gausswtwo_i \leq -c_1}\\
    &\gtrsim
    \frac{1}{n} \sum_{i=1}^n ( 1-\nu\abs{\gausswone_i} + c_1\eta )^2 \indicator{1-\nu\abs{\gausswone_i} \geq \frac{1}{2}, \gausswtwo_i \leq -c_1}\\
    &\gtrsim
    (1+\eta^2) \frac{1}{n} \sum_{i=1}^n\indicator{1-\nu\abs{\gausswone_i} \geq \frac{1}{2}, \gausswtwo_i \leq -c_1}
    \label{eq:fn_first_bound_eta_lower}
\end{align}
Moreover, from independence of $\Gausswone$ and $\Gausswtwo$, the fact that $\PP \left( \Gausswtwo \leq -c_1 \right)=\Phic (c_1)\geq c_2$ and concentration of Bernoulli random variables we obtain that $\fnclasn(\nu,\eta) \gtrsim (1+\eta^2) \frac{1}{n} \sum_{i=1}^n\indicator{1-\nu\abs{\gausswone_i} \geq \frac{1}{2}}$ with probability $\geq 1-\exp(-c_3 n)$. Now in order to lower bound the last term we note that:
\begin{align}
    \PP \left( \abs{\Gausswone} \leq \frac{1}{2\nu} \right)  = \erf \left( \frac{1}{2\sqrt{2}\nu} \right) \gtrsim \frac{1}{\nu}
\end{align}
where we used Taylor approximation $\erf \left( \frac{1}{2\sqrt{2}\nu} \right) \gtrsim \frac{1}{\nu} $ for any $\nu \geq c_1$ with $c_1>0$ sufficiently large. From Lemma~\ref{lemma:DKW_ineq} with $\epsilon\asymp\sqrt{n}/\nu_{\max}$ we obtain that uniformly over $\nu,\eta$ $\fnclasn (\nu,\eta) \gtrsim \frac{1}{\nu}+\frac{\eta^2}{\nu}$ with probability at least $1-2\exp(-c_2 n/(\nu_{\max})^2 )$. 


\end{proof}

\subsubsection*{Uniform convergence of \texorpdfstring{$\fnclasn$}{fnclasn} to \texorpdfstring{$\fclasn$}{fclasn}}
Similarly as in Section~\ref{apx:subsec:unif_conv} we define a random variable $X = (\Gausswone, \Gausswtwo)$ and a set of functions $\setGn :=\{ (\Gausswone,\Gausswtwo)\mapsto (1-\nu \abs{\Gausswone} - \Gausswtwo \eta )_+^2 \ |\nu_{\max}\geq \nu \geq \nu_{\min}, \eta \leq \eta_{\max} \}$ with $\nu_{\min} = \Theta(\nu_{\max}),\nu_{\min} = \Omega(n^{1/6})$ and $\eta_{\max}\leq c_2$ for some universal constant $c_2>0$. Using notation of Section~\ref{subsec:uniform_conv_tools} we have that $P\gnueta = \EE \gnueta(\Gausswone,\Gausswtwo) = \fclasn (\nu,\eta)$ and $P_n \gnueta = \fnclasn(\nu,\eta)$,  we show the following result:
\begin{proposition}
\label{prop:unifConv_noiseless_whole}
There exist positive universal constants $c_1,c_2,c_3>0$ such that for any $\epsilon \gtrsim \frac{\log n}{\sqrt{n}}$ holds
    \begin{align}
        \prob \left( \norm{ P_n - P }_{\setGn} \leq c_1\frac{\log n}{\sqrt{n}} + \epsilon \right) 
        \geq
        1 - c_2\exp \left( -c_3  n \epsilon^2  \right).
    \end{align}
\end{proposition}
\begin{proof}
The proof is based on Theorem~\ref{thm:unifconv_adamczak}. We choose $\alpha = 1$ and show that the condition from Theorem~\ref{thm:unifconv_adamczak} requiring finite Orlicz norms is indeed satisfied for this choice of $\alpha$.
We separate the proof into three steps, where in a first step we bound the variable $ \psi_{\setGn}$, then we bound $\rademacher(\setGn)$, and at last, we bound $\sigma_{\setGn}^2$ and apply Theorem~\ref{thm:unifconv_adamczak}. 

\paragraph{Step 1: Bounding $\psi_{\setGn}$}

By the definition of Orlicz norms, $\psi_{\setGn}$ is given by:
\begin{align}\label{eq_psi_setGn_def}
    \psi_{\setGn} 
    = \inf \{ \lambda>0:\ \EE [ \exp ( \frac{1}{\lambda} \max_{1\leq i\leq n} \sup_{\gnueta\in\setGn} \frac{1}{n}
    | \gnueta(\gausswone_i,\gausswtwo_i) - \EE [\gnueta] | - 1 ]) \leq 1 \}
\end{align}
Note that $(1-\nu\abs{\gausswone})_+ \leq 1$ and thus we have $\gnueta(\gausswone,\gausswtwo) = ( 1 - \nu |\gausswone| - \gausswtwo \eta)_+^2 \lesssim 1+ (\gausswtwo)^2 \eta^2$ for any $\gausswone,\gausswtwo,\eta,\nu$, implying that
\begin{align}
    \max_i \sup_{\nu,\eta} |\gnueta(\gausswone_i,\gausswtwo_i)| = \max_i \sup_{\nu,\eta} |(1-\nu \abs{\gausswone_i} - \gausswtwo_i \eta )_+^2 |  \leq c_1\gausswtwo_{\max}
\end{align} 
with vector $\gausswtwo_{\max} = \max_{1\leq i\leq n} |\gausswtwo_i|$. Furthermore, it also holds $\EE[\gnueta] \lesssim 1+\eta^2 \EE (\Gausswtwo)^2 \leq 1+\eta_{\max}^2\leq c_3$ for some universal constant $c_3>0$.

Using these results and applying the triangle inequality, the term inside of expectation in Equation~\eqref{eq_psi_setGn_def} can be bounded as:
\begin{align}
    \EE \bigg[ \exp \Big( \frac{1}{\lambda} \max_i& \sup_{\nu,\eta} \frac{1}{n}
    \Big| ( 1 -  \nu |\gausswone_i| - \gausswtwo_i \eta )_+^2 - \EE [ ( 1 -  \nu |\Gausswone| - \Gausswtwo \eta )_+^2 ] \Big| \Big) \bigg]
    \\
     \leq 
    \EE \bigg[ \exp \Big(  \frac{1}{n\lambda} &\max_i \sup_{\nu,\eta} 
    ( 1 -  \nu |\gausswone_i| - \gausswtwo_i \eta )_+^2 \Big) \bigg]\\
    & \cdot
    \exp \Big( \frac{1}{n\lambda}\sup_{\nu,\eta} \EE [ ( 1 -  \nu |\Gausswone| - \Gausswtwo \eta )_+^2] \Big) 
    \leq
    \EE \left[ \exp \left(  \frac{c_1}{n\lambda}z_{\max}^2 \right) \right] \exp \left( \frac{c_3}{n\lambda} \right)  
    \label{eq:proof_unicon_prop_noiseless_exp_triangle}
\end{align}
for some positive universal constants $c_1,c_3$. Now we split the expectation from the above inequality into two terms:
\begin{align}
   \EE \left[\onevec \Big[z_{\max} < \sqrt{2 \log (n)} \Big] \exp \left(  \frac{c_1}{n\lambda}z^2_{\max}  \right) \right]
   \leq
   \exp \left( \frac{2c_1 \log n}{n\lambda} \right)
\end{align}
and
\ifaistats
\begin{align}
    \EE & \left[\onevec \Big[z_{\max} \geq \sqrt{2 \log n} \Big] \exp \left(  \frac{c_1}{n\lambda}z_{\max}^2  \right) \right]
    = 2n\EE  \left[ \onevec \Big[  z_{\max} = \abs{z_1}, \abs{z_1} \geq \sqrt{2 \log n} \Big] \exp \left(  \frac{c_1}{n\lambda}z_1^2  \right) \right]\\
    &\lesssim
    n \int_{z_1 = \sqrt{2 \log n}}^{\infty} \int_{-z_1}^{z_1} \cdots \int_{-z_1}^{z_1} \exp \left( \frac{c_1}{n\lambda}  z_1^2 \right) \left[ \prod_{i=2}^{2n} \frac{\exp(-\frac{1}{2}z_i^2)}{\sqrt{2\pi}}  dz_i \right] dz_1
    \lesssim
    n \int_{\sqrt{2 \log n}}^{\infty} \exp \left( - z_1^2 \left(\frac{1}{2} - \frac{c_1}{n\lambda} \right) \right) dz_1 \\
    &\lesssim
    \frac{\exp \left( \frac{2c_1n}{n\lambda} \right)}{\sqrt{\log n }(1- \frac{2c_1}{n\lambda})}
    \label{eq:zmax_geq_logs}
\end{align}
\else
\begin{align}
    \EE & \left[\onevec \Big[z_{\max} \geq \sqrt{2 \log n} \Big] \exp \left(  \frac{c_1}{n\lambda}z_{\max}^2  \right) \right]
    = 2n\EE  \left[ \onevec \Big[  z_{\max} = \abs{z_1}, \abs{z_1} \geq \sqrt{2 \log n} \Big] \exp \left(  \frac{c_1}{n\lambda}z_1^2  \right) \right]\\
    &\lesssim
    n \int_{z_1 = \sqrt{2 \log n}}^{\infty} \int_{-z_1}^{z_1} \cdots \int_{-z_1}^{z_1} \exp \left( \frac{c_1}{n\lambda}  z_1^2 \right) \left[ \prod_{i=2}^{2n} \frac{\exp(-\frac{1}{2}z_i^2)}{\sqrt{2\pi}}  dz_i \right] dz_1\\
    &\lesssim
    n \int_{\sqrt{2 \log n}}^{\infty} \exp \left( - z_1^2 \left(\frac{1}{2} - \frac{c_1}{n\lambda} \right) \right) dz_1 
    \lesssim
    \frac{\exp \left( \frac{2c_1n}{n\lambda} \right)}{\sqrt{\log n }(1- \frac{2c_1}{n\lambda})}
    \label{eq:zmax_geq_logs}
\end{align}
\fi
where we assumed that $\lambda > \frac{2c_1}{n}$. Now choosing $\lambda = c_{\lambda} \frac{\log n}{n}$ with a positive constant $c_{\lambda}$ sufficiently large, we find that the condition in Equality \eqref{eq_psi_setGn_def} is satisfied for this $\lambda$, which implies that $\psi_{\setGn} \leq c_{\lambda}\frac{\log n}{n}$.

\paragraph{Step 2: Bounding $\rademacher(\setGn)$}
In order to apply Theorem~\ref{thm:unifconv_adamczak} we need to upper bound $\EE \norm{P_n-P}_{\setGn}$. Since $\EE \norm{P_n-P}_{\setGn} \leq 2 \rademacher(\setGn)$, we can instead upper bound Rademacher complexity $\rademacher(\setGn)$, which we do next. Recall the definition of the Rademacher complexity:
\begin{align}
    \rademacher(\setGn) = \EE \left[ \sup_{\gnueta\in \setGn} \left|\frac{1}{n}\sumin \epsilon_i \gnueta(\gausswone_i,\gausswtwo_i) \right| \right]
\label{eq:proof_unicon_noiseless_rademacher_def}
\end{align}
Define random variable $\tz := \abs{\gausswone} \indicator{ \abs{\gausswone} \leq \frac{1+\eta_{\max}\sqrt{3\log n} }{\nu_{\min}}}$ and note that for all $\nu,\eta$ and $1\leq i\leq n$ holds
\begin{align}
    (1 - \nu |\gausswone_i| - \gausswtwo_i \eta )_+^2 \indicator{\gausswtwo_{\max}\leq \sqrt{3\log n}} 
    =
    ( 1 - \nu \tz_i - \gausswtwo_i \eta )_+^2 \indicator{\gausswtwo_{\max}\leq \sqrt{3\log n}}.    
\end{align}
We now apply the triangle inequality to Equation~\eqref{eq:proof_unicon_noiseless_rademacher_def} to obtain:
\begin{align}
    \EE \sup_{\nu, \eta}\Big| \frac{1}{n} \sumin \epsilon_i ( 1 - \nu |\gausswone_i| &- \gausswtwo_i \eta )_+^2 \Big|
    \leq 
    \EE \sup_{\nu, \eta}\Big| \frac{1}{n}\sumin \epsilon_i ( 1 - \nu \tz_i - \gausswtwo_i\eta )_+^2 \indicator{\gausswtwo_{\max} \leq \sqrt{3\log n}} \Big|\\
    &+
    \EE \sup_{\nu, \eta}\Big| \frac{1}{n}\sumin \epsilon_i ( 1 - \nu |\gausswone_i| - \gausswtwo_i \eta)_+^2 \indicator{\gausswtwo_{\max} > \sqrt{3\log n}}  \Big|
    \label{eq:ineq_rademacher_noiseless_two_terms}
\end{align}
Then, using that $(\cdot)_+$ is $1$-Lipschitz, we can bound expectation of the first term from Equation~\eqref{eq:ineq_rademacher_noiseless_two_terms} as follows:
\ifaistats
\begin{align}
    &\EE \sup_{\nu, \eta}\left| \frac{1}{n}\sumin \epsilon_i ( 1 - \nu \tz_i - \gausswtwo_i \eta )_+^2 \indicator{\gausswtwo_{\max} \leq \sqrt{3\log n}} \right|
    \lesssim
    \EE \sup_{\nu, \eta}\left| \frac{1}{n}\sumin \epsilon_i ( 1 - \nu \tz_i - \gausswtwo_i \eta )^2 \indicator{\gausswtwo_{\max} \leq \sqrt{3\log n}} \right|\\
   &=
    \EE \sup_{\nu, \eta}\left| \frac{1}{n}\sumin \epsilon_i \left[ ( 1 - \nu \tz_i)^2 - 2( 1 - \nu \tz_i) \gausswtwo_i \eta + (\gausswtwo_i)^2 \eta^2 \right] \indicator{\gausswtwo_{\max} \leq \sqrt{3\log n}} \right|
\end{align}
\else
\begin{align}
    \EE \sup_{\nu, \eta}\bigg| \frac{1}{n}\sumin &\epsilon_i ( 1 - \nu \tz_i - \gausswtwo_i \eta )_+^2 \indicator{\gausswtwo_{\max} \leq \sqrt{3\log n}} \bigg|\\
    &\lesssim
    \EE \sup_{\nu, \eta}\left| \frac{1}{n}\sumin \epsilon_i ( 1 - \nu \tz_i - \gausswtwo_i \eta )^2 \indicator{\gausswtwo_{\max} \leq \sqrt{3\log n}} \right|\\
   &=
    \EE \sup_{\nu, \eta}\left| \frac{1}{n}\sumin \epsilon_i \left[ ( 1 - \nu \tz_i)^2 - 2( 1 - \nu \tz_i) \gausswtwo_i \eta + (\gausswtwo_i)^2 \eta^2 \right] \indicator{\gausswtwo_{\max} \leq \sqrt{3\log n}} \right|
\end{align}
\fi
We use again the triangle inequality and consider each of the three terms above:
\begin{itemize}
\item Note that $\abs{\nu\tz_i} \leq \frac{\nu_{\max}}{\nu_{\min}}\eta_{\max}\sqrt{3\log n} \lesssim \sqrt{\log n}$ and using concentration of sub-exponential random variables from Lemma~\ref{lemma:bernstein_subexponential} we obtain:
\begin{align}
    &\EE \sup_{\nu}\left| \frac{1}{n}\sumin \epsilon_i  ( 1 - \nu \tz_i)^2 \indicator{\gausswtwo_{\max} \leq \sqrt{3\log n}} \right|
    \leq
    \EE \sup_{\nu}\left| \frac{1}{n}\sumin \epsilon_i  \nu^2 \tz_i^2 \indicator{\gausswtwo_{\max} \leq \sqrt{3\log n}} \right|\\
    &+
    \EE \sup_{\nu}\left| \frac{1}{n}\sumin \epsilon_i  (-2\nu \tz_i) \indicator{\gausswtwo_{\max} \leq \sqrt{3\log n}} \right|
    +
    \EE \left| \frac{1}{n}\sumin \epsilon_i   \indicator{\gausswtwo_{\max} \leq \sqrt{3\log n}} \right|
    \lesssim
    \frac{\log n}{\sqrt{n}}
\end{align}
\item Similarly as in the previous case, we use triangle inequality to split expectation into two terms and then use that $\abs{\gausswtwo_i \eta} \leq \gausswtwo_{\max}\eta_{\max} \lesssim \sqrt{\log n}$ and $\abs{\nu\tz_i\gausswtwo_i\eta} \leq 3\frac{\nu_{\max}}{\nu_{\min}}\eta_{\max}^2 \log n \lesssim \log n$, and apply concentration from Lemma~\ref{lemma:bernstein_subexponential} to get:
\begin{align}
    \EE \sup_{\nu,\eta}\left| \frac{1}{n}\sumin 2\epsilon_i ( 1 - \nu \tz_i) \gausswtwo_i \eta \indicator{\gausswtwo_{\max} \leq \sqrt{3\log n}} \right|
    \lesssim
    \frac{\log n}{\sqrt{n}}
\end{align}
\item Last, use that $\eta^2 (\gausswtwo_i)^2 \leq \eta_{\max}^2 (\gausswtwo_{\max})^2 \lesssim \log n$, and again concentration of sub-exponential random variables from Lemma~\ref{lemma:bernstein_subexponential} to obtain:
\begin{align}
    \EE \sup_{ \eta}\left| \frac{1}{n}\sumin \epsilon_i (\gausswtwo_i)^2 \eta^2 \indicator{\gausswtwo_{\max} \leq \sqrt{3\log n}} \right|
    \lesssim
    \frac{1}{\sqrt{n}}
\end{align}
\end{itemize}
Thus, we bounded the first term from Equation~\eqref{eq:ineq_rademacher_noiseless_two_terms}. Now, we bound the second term. Since $|\epsilon_i ( 1 - \nu |\gausswone_i| - \gausswtwo_i \eta )_+^2| \leq (1+\gausswtwo_i\eta)^2$ we obtain:
\ifaistats
\begin{align}
    \EE \sup_{\nu, \eta}\bigg| &\frac{1}{n}\sumin \epsilon_i ( 1 - \nu |\gausswone_i| - \gausswtwo_i \eta )_+^2 \indicator{\gausswtwo_{\max} > \sqrt{3\log n}}  \bigg|
    \lesssim
    \EE \sup_{\eta} \frac{1}{n}\sumin ( 1 + \gausswtwo_i \eta)^2 \indicator{\gausswtwo_{\max} > \sqrt{3\log n}} \\
    &\lesssim
    \frac{1}{n}\EE \sumin ( 1 + (\gausswtwo_i)^2 ) \indicator{\gausswtwo_{\max} > \sqrt{3\log n}}  
    \lesssim
    \EE \left[ (\gausswtwo_{\max})^2\indicator{\gausswtwo_{\max} > \sqrt{3\log n}} \right] \\
    &\lesssim
    n \int_{z_1=\sqrt{3\log n}}^{\infty} z_1^2 \exp(-z_1^2/2)dz_1 
    \lesssim
    \frac{\sqrt{\log n}}{\sqrt{n}}
\end{align}
\else
\begin{align}
    \EE \sup_{\nu, \eta}\bigg| &\frac{1}{n}\sumin \epsilon_i ( 1 - \nu |\gausswone_i| - \gausswtwo_i \eta )_+^2 \indicator{\gausswtwo_{\max} > \sqrt{3\log n}}  \bigg|\\
    &\lesssim
    \EE \sup_{\eta} \frac{1}{n}\sumin ( 1 + \gausswtwo_i \eta)^2 \indicator{\gausswtwo_{\max} > \sqrt{3\log n}} \\
    &\lesssim
    \frac{1}{n}\EE \sumin ( 1 + (\gausswtwo_i)^2 ) \indicator{\gausswtwo_{\max} > \sqrt{3\log n}}  
    \lesssim
    \EE \left[ (\gausswtwo_{\max})^2\indicator{\gausswtwo_{\max} > \sqrt{3\log n}} \right] \\
    &\lesssim
    n \int_{z_1=\sqrt{3\log n}}^{\infty} z_1^2 \exp(-z_1^2/2)dz_1 
    \lesssim
    \frac{\sqrt{\log n}}{\sqrt{n}}
\end{align}
\fi
where in the last step we used the same approach as for obtaining Equation~\eqref{eq:zmax_geq_logs}. After addition of all terms we obtain that $\rademacher(\setGn)\lesssim \frac{\log n}{\sqrt{n}}$. 

\paragraph{Step 3: Proof of the statement}
To apply Theorem~\ref{thm:unifconv_adamczak}, we also need to bound the variance $\sigma_{\setGn}^2$. But, it is straightforward that there exists some positive universal constant $c_{\sigma_{\setGn}}>0$  such that the variance is bounded as follows:
\begin{align}
\label{eq:bound_var_setGn}
    \sigma_{\setGn}^2
    \leq 
    \sup_{ \gnueta \in \setGn } \EE \left[ \gnueta^2 \right]
    \leq
    c_{\sigma_{\setGn}} \left( 1 + \eta_{\max}^4 \right)
\end{align}
Substituting all derived bounds into the probability statement from Theorem~\ref{thm:unifconv_adamczak} we obtain for $\epsilon\gtrsim \frac{\log n}{\sqrt{n}}$:
    \begin{align}
        \prob \left( \norm{ P_n - P }_{\GB} \geq 2(1+t) \mathcal{R}_{\GB} + \epsilon \right) 
        \leq
        \exp \left( - c_2 n \epsilon^2 \right) + 3\exp \left( -c_3 \frac{ n \epsilon }{\log n} \right) \leq c_4\exp(-c_2 n\epsilon^2)
    \end{align}
with $c_2^{-1} = 2(1+\delta)c_{\sigma_{\setGn}} \left( 1 + \eta_{\max}^4 \right)$ and $c_3^{-1} = C c_{\lambda}$, which concludes the proof. 
\end{proof}



\begin{corollary}
\label{corr:fnoiseless_taylor}
There exist positive universal constants $c_1,c_2$ such that for any $\nu,\eta$ satisfying constraint in $\setGn$ and $\epsilon \gtrsim \frac{\log n}{\sqrt{n}}$, inequality
\begin{align}
    \fnclasn (\nu,\eta) \geq \frac{\sqrt{2}}{3\sqrt{\pi}} \frac{1}{\nu} + \sqrt{\frac{2}{\pi}} \frac{\eta^2}{\nu} - \epsilon
\end{align}
holds with probability at least $1-c_1\exp(-c_2 n\epsilon^2)$ over the draws of $\gausswone,\gausswtwo$.
\end{corollary}
\begin{proof}
Recall that $\fclasn(\nu,\eta) = \EE [\fnclasn(\nu,\eta)]$. From Proposition~\ref{prop:unifConv_noiseless_whole} we have that $\fnclasn(\nu,\eta)\geq \fclasn(\nu,\eta)-\epsilon$ uniformly over all admissible $(\nu,\eta)$ with probability $\geq 1-c_1\exp(-c_2 n\epsilon^2)$. According to Lemma~\ref{lemma:fclas_fclasn_inf_diff}, $\fclas$ is an infinitely differentiable function and thus we can express it by Taylor series. First, we determine the coefficients of the series of $\fclasn(\nu,\cdot):\eta\mapsto\fclasn(\nu,\eta)$.

The constant coefficient is given by:
\begin{align}
    \fclasn(\nu,0) &= \EE \possq{1-\nu\abs{\Gausswone}} 
    =
    \frac{2}{\sqrt{2\pi}} \int_{0}^{1/\nu} (1-\nu z)^2 \exp \left( -\frac{z^2}{2} \right) dz \\
    &=
    (\nu^2+1)\erf \left( \frac{1}{\sqrt{2}\nu} \right) + \sqrt{\frac{2}{\pi}} \nu \left( \exp \left( -\frac{1}{2\nu^2} \right) -2 \right) 
    = \frac{\sqrt{2}}{3\sqrt{\pi}} \frac{1}{\nu} + O \left( \frac{1}{\nu^3} \right)
    \label{eq_exp_partial_nu_fnclasn}
\end{align}
where we used the Taylor expansion around $0$ for functions $\erf$ and $\exp$. The first derivative coefficient is given by 
\begin{align}
\frac{\partial}{\partial \eta} \fclasn(\nu,\eta)\vert_{\eta=0} = -2\EE [\Gausswtwo(1-\nu\abs{\Gausswone}-\eta\Gausswtwo)_+]\vert_{\eta=0} = 0
\end{align}
since $\Gausswone$ and $\Gausswtwo$ are independent random variables and $\EE[\Gausswtwo]=0$. Now consider the second derivative coefficient:
\begin{align}
    \frac{\partial^2}{\partial \eta^2} \fclasn(\nu,\eta)\vert_{\eta=0} 
    &=
    2\EE \left[\indicator{1-\nu\abs{\Gausswone}-\eta\Gausswtwo} (\Gausswtwo)^2 \right]\vert_{\eta=0} 
    =
    2\PP \left(\abs{\Gausswone}\leq \frac{1}{\nu} \right) \\
    &=
    2\erfc \left( \frac{1}{\sqrt{2}\nu} \right) = 2\sqrt{\frac{2}{\pi}}\frac{1}{\nu} + O\left( \frac{1}{\nu^3} \right)
\end{align}
where in the last step we used the Taylor series approximation of the error function around zero. Now, in order to analyze higher order derivatives, we show using Leibniz integral rule that:
\begin{align}
    \frac{\partial^3}{\partial \eta^3} \fclasn(\nu,\eta) 
    &=
    \frac{2}{\pi} \frac{\partial}{\partial \eta}  \int_{\Gausswtwo=-\infty}^{1/\eta} \int_{\Gausswone = 0}^{(1-\eta\Gausswtwo)/\nu} (\Gausswtwo)^2 \exp\left(-\frac{1}{2}(\Gausswtwo)^2 \right) \exp\left(-\frac{1}{2} (\Gausswone)^2\right) d\Gausswone d\Gausswtwo\\
    &=
    -\frac{2}{\pi\nu} \int_{\Gausswtwo=-\infty}^{1/\eta} (\Gausswtwo)^3\exp\left(-\frac{1}{2}(\Gausswtwo)^2\right) \exp\left(-\frac{1}{2} \left(\frac{1-\eta\Gausswtwo}{\nu}\right)^2\right)  d\Gausswtwo
    \label{eq:proof_taylor_third_differential}
\end{align}
Now, note that for higher order derivatives, the term that comes from differentiating the upper bound $1/\eta$ is equal $0$ for $\eta=0$ since it is of the form poly$(1/\eta)\exp(-1/(2\eta^2))$ which is zero for any polynomial. Thus the main term which we need to consider comes from the term $\exp\left(-\frac{1}{2} \left(\frac{1-\eta\Gausswtwo}{\nu}\right)^2\right)$. Note that after taking differential with respect to this term we obtain additional multiplicative factor $1/\nu^2$. However, we also obtain multiplicative term $(1-\nu\Gausswtwo)$ which can be further differentiated with respect to $\eta$. Taking all this into account one can show that for $k=2,3,...$
\begin{align}
    &\frac{\partial^{2k}}{\partial \eta^{2k}} \fclasn(\nu,\eta) \Big|_{\eta=0}=
    \\
    &O \left(
    \frac{1}{\nu^{2k-1}} \int_{\Gausswtwo=-\infty}^{1/\eta} (\Gausswtwo)^{2k}(1-\eta\Gausswtwo)\exp\left(-\frac{1}{2}(\Gausswtwo)^2\right)  \exp\left(-\frac{1}{2} \left(\frac{1-\eta\Gausswtwo}{\nu}\right)^2\right)  d\Gausswtwo \bigg|_{\eta=0} \right)
\end{align}
with all other terms either vanishing at $\eta=0$ or having in front of the integral multiplicative constant $\frac{1}{\nu^p}$ with $p>2k-1$. Thus for $\eta=0$, using that Gaussian moments are bounded, we obtain that $\frac{\partial^{2k}}{\partial \eta^{2k}} \fclasn(\nu,\eta)\vert_{\eta=0} = O\left(\frac{1}{\nu^{2k-1}} \right)$. Similarly to Equation~\eqref{eq:proof_taylor_third_differential}, one can show that every odd differential at $\eta=0$ is equal to scaled odd moments of standard Gaussian random variable, implying that $\frac{\partial^{2k+1}}{\partial \eta^{2k+1}} \fclasn(\nu,\eta)\vert_{\eta=0}=0$. 

Taking all derived coefficients into consideration, we can express $\fclasn$ using the following Taylor series:
\begin{align}
    \fclasn(\nu,\eta) = \frac{\sqrt{2}}{3\sqrt{\pi}} \frac{1}{\nu} + \sqrt{\frac{2}{\pi}} \frac{\eta^2}{\nu} + O\left( \frac{1}{\nu^3}, \frac{\eta^4}{\nu^3} \right)
    \label{eq:taylor_fclasn}
\end{align}
At the end, since $\eta =O(1)$ and $\nu=\Omega(n^{1/6})$ we have $O\left( \frac{1}{\nu^3}, \frac{\eta^4}{\nu^3} \right) = o(\epsilon)$, which finishes the proof.
\end{proof}

\subsection{Lower bounds for \texorpdfstring{$\fnclas$}{fn} in noisy setting}
\label{subsec:appendix_lower_bounds_fnclas}
Recall that we have defined $\nu= \innerprodk{\wpar}{\wgt}, \etas = \normk{\wperps}_2, \etasc = \normk{\wperpsc}_2$ and $\eta = \normk{\wperp}_2 = \sqrt{\etas^2+\etasc^2}$, and also the following two functions:
\begin{align}
    \fnclas (\nu, \eta) &= \frac{1}{n} \sum_{i=1}^n ( 1 - \xi_i \nu |\gausswone_i| - \gausswtwo_i \eta )_+^2\\
    \fclas(\nu,\eta) &= \EE  \fnclas \left( \nu, \eta \right)  = \EE_{\Gausswone,\Gausswtwo \sim \NNN(0,1)} \EE_{\xirv \sim \prob(\cdot\vert \Gausswone)} ( 1 - \xirv \nu |\Gausswone| - \Gausswtwo \eta )_+^2 .
    \label{eq:defoff}
\end{align}
In this section we show three lower bounds for $\fnclas$ of increasing tightness. First, we show a lower bound by a quadratic form in $\nu$ and $\eta$, after that we bound $\fnclas$ by a sum of a quadratic form and a constant, and the last bound is based on the uniform convergence of $\fnclas$ to $\fclas$ which we prove at the end of this subsection.
\subsubsection*{Lower bounding \texorpdfstring{$\fnclas$}{fn} by a quadratic form}
\label{apx:subsec:quadratic_bound_fn}
We show the following lemma.
\begin{lemma} \label{quadratic_bound_fn}
    There exist universal positive constants $\constnu, \consteta$ only depending on $\probsigma$ and $c$ such that for any $\nu,\eta$ we have that: 
        \begin{align}
        \label{eq:quadratic_bound_fn}
            \fnclas(\nu,\eta) \geq \constnu \nu^2  + \consteta \eta^2
        \end{align}
    with probability at least $1 - \exp \left( -cn \right)$ over the draws of $\gausswone,\gausswtwo,\xi$.
\end{lemma}

\begin{proof}
 We can assume that $\nu\geq 0$ as the other cases follow from exactly the same argument. 
 First we show an auxiliary statement which we use later in the proof. Namely, we claim that there exists some positive constant $c_1$ such that for all $z \in [z_1,z_2]$,  $\probsigma \left( \xi = -1; z \right) > c_1$ for some $z_1, z_2 \in \RR$ and $z_1\neq z_2$. Let's prove this statement by contradiction and assume that there exists no $z \in [z_1,z_2]$ which satisfy the previous equation. Then, for almost any $z \sim \NNN(0,1)$, we have $\probsigma (\xi; z) = +1$ and hence the minimum of the function $\fclas(\nu, \eta) = \EE \fnclas (\nu, \eta)$ is obtained for $\nu = \infty$. However, this is in contradiction with the Assumption 1 in Section \ref{subsec:main_results_noisy}. Hence there exists some $z$ for which $\prob \left( \xi = -1;  z \right) > c_1$. By the assumption on $\probsigma$ in Section \ref{subsec:main_results_noisy} we assume piece-wise continuity of $z \to \probsigma(\xi=-1;z)$ and hence there exists some interval $[z-\delta, z+\delta] =: [z_1,z_2]$ in which the given probability is bounded away from zero. 

We can assume without loss of generality that this interval does not contain zero, since in that case we can always define a new interval of the form $[\epsilon, z_2]$ or $[z_1, -\epsilon]$ for $\epsilon>0$ small enough, which does not contain zero. Let's define $\tilde{z} = \min \{ \abs{z_1}, \abs{z_2} \}$.

 We can now bound $\fnclas(\nu,\eta)$ as follows:
\begin{align}
    \fnclas(\nu,\eta) 
    & =
    \frac{1}{n} \sum_{i=1}^n ( 1 - \xi_i \nu |\gausswone_i| - \gausswtwo_i \eta )_+^2\\
    &\geq
    \frac{1}{n}\sumin \indicator{ \xi_i = -1, \gausswone_i \in [z_1,z_2], \gausswtwo_i < - c_2 }  ( 1 - \xi_i \nu |\gausswone_i| - \gausswtwo_i \eta )_+^2
    \\
    & \geq \left( 1 + \tilde{z} \nu + c_2 \eta \right)^2 \frac{1}{n} \sum_{i=1}^n \idvec [ \xi_i = -1, {z_i}^{(0)} \in [z_1,z_2], {z_i}^{(1)} < - c_2 ]
\end{align}
From Section~\ref{subsec:path_param} we have that $\Gausswtwo$ is independent of $\xirv$ and $\Gausswone$. Hence:
\ifaistats
\begin{align}
    \prob ( \xirv = -1,  \Gausswone \in [z_1,z_2], \Gausswtwo < - c_2 ) 
    &=
    \prob ( \xirv = -1 \vert \Gausswone \in [z_1,z_2] ) \prob ( \Gausswone \in [z_1,z_2] ) \prob ( \Gausswtwo < - c_2 )
    \\
    &\geq
    c_1 \left( \Phic(z_1) - \Phic(z_2) \right) \Phic(c_2) 
    \geq
    c
\end{align}
\else
\begin{align}
    \prob& ( \xirv = -1,  \Gausswone \in [z_1,z_2], \Gausswtwo < - c_2 ) \\
    &=
    \prob ( \xirv = -1 \vert \Gausswone \in [z_1,z_2] ) \prob ( \Gausswone \in [z_1,z_2] ) \prob ( \Gausswtwo < - c_2 )
    \geq
    c_1 \left( \Phic(z_1) - \Phic(z_2) \right) \Phic(c_2) 
    \geq
    c
\end{align}
\fi
for some positive universal constant $c$. Now using concentration of i.i.d. Bernoulli random variables we obtain:
\begin{align}
    \fnclas (\nu,\eta) \geq \left( 1 + \tilde{z} \nu + c_2 \eta \right)^2 \frac{c}{2} 
    \gtrsim
    \nu^2 + \eta^2 
\end{align}
with probability at least $1 - \exp \left( - cn \right)$.
\end{proof}


\subsubsection*{Lower bounding \texorpdfstring{$\fnclas$}{fn} by a quadratic form with constant}
\label{apx:subsec:quadratic_plus_fstar_bound}
Recall that $\deltanu = \nu -\nubar$. We show the following lemma. 
\begin{lemma} \label{lemma_quadratic_plus_fstar_bound}
    Let $\boundnu, \boundeta >0$ be universal positive constants. Then, there exist positive constants $\tconstnu,\tconsteta >0$ and $c_1, c_2,c_3 >0$ only depending on $\probsigma$, such that for any $\epsilon \geq  \frac{c_1}{\sqrt{n}}$ and any $\nu^2 \leq \boundnuone^2, \eta \leq \boundetaone$ we have that: 
        \begin{align}
        \label{eq:quadratic_plus_fstar_bound}
            \fnclas(\nu,\eta) \geq \fstar + \tconstnu \left( \deltanu \right)^2  + \tconsteta \eta^2   - \epsilon
        \end{align}
    with probability at least $1 - \exp \left( -c_2  n \epsilon^2  \right) -  \exp \left( -c_3 \frac{ n \epsilon }{\log n} \right)$ over the draws of $\gausswone,\gausswtwo,\xi$.
\end{lemma}
\begin{proof}
First note that from the uniform convergence result in Proposition \ref{prop:unifConv} we have that $\fclas(\nu,\eta) \geq \fnclas(\nu,\eta) - \epsilon$, with $\fclas$ from Equation~\eqref{eq:defoff}, with high probability. Thus, it is sufficient to study $\fclas$.  Clearly, by the convexity of $\fclas$ we have that $\fclas \geq \fstar$ with $\fstar = \fclas(\nubar, 0)$ where we use the simple fact that $(\nubar, 0)$ is the global minimizer of $\fclas$, which follows from the  assumption on $\probsigma$ in Section \ref{subsec:main_results_noisy}. 
Furthermore, it is not difficult to check that for for any $\nu,\eta$, $\nabla^2 \fclas(\nu,\eta) \succ 0$ and therefore, $\fclas$ is strictly convex on every compact set. Hence, the proof follows. 
\end{proof}

\subsubsection*{Uniform convergence of \texorpdfstring{$\fnclas$}{fnclas} to \texorpdfstring{$\fclas$}{fclas}}
\label{apx:subsec:unif_conv}

Recall that $\Gausswone,\Gausswtwo \sim \Normal(0,1)$ are independent Gaussian random variables and $\xirv$ a random variable with $\xirv \vert \Gausswone \sim \probsigma(.; \Gausswone)$. Using notation introduced in Section~\ref{subsec:uniform_conv_tools} with random variable $X=(\Gausswone,\Gausswtwo,\xirv)$, and $\GB = \left\{ \gnueta\ |\ \abs{\nu} \leq \boundnuone, \eta \leq \boundetaone \right\}$, we note that
\begin{align}
    P\gnueta =\linebreak[1] \EE \gnueta(\Gausswone,\Gausswtwo,\xirv) = \fclas (\nu,\eta)
    ~~\andtxt~~
    P_n \gnueta = \fnclas(\nu,\eta).   
\end{align}
We show the following result:

\begin{proposition}
\label{prop:unifConv}
There exist positive universal constants $c_1,c_2,c_3>0$ such that
    \begin{align}
        \prob \left( \norm{ P_n - P }_{\GB} \leq \frac{c_1}{\sqrt{n}} + \epsilon \right) 
        \geq
        1 - \exp \left( -c_2  n \epsilon^2  \right) -  \exp \left( -c_3 \frac{ n \epsilon }{\log n} \right)
    \end{align}
\end{proposition}

\begin{proof}
The proof of the proposition is based on the application of Theorem~\ref{thm:unifconv_adamczak} and follows exactly the same steps as proof of Proposition~\ref{prop:unifConv_noiseless_whole}. In order to apply Theorem~\ref{thm:unifconv_adamczak} we need to upper bound three terms - $\psi_{\GB}, \sigma^2_{\GB}$ and $\rademacher(\GB)$. Similarly as in proof of Proposition~\ref{prop:unifConv_noiseless_whole} we split proof into three steps:

\paragraph{Step 1: Bounding $\psi_{\GB}$}

Recall the definition of $\psi_{\GB}$ from Theorem~\ref{thm:unifconv_adamczak}:
\begin{align}\label{eq_psi_GB_def}
    \psi_{\GB} 
    = \inf \{ \lambda>0:\ \EE [ \exp ( \frac{1}{\lambda} \max_i \sup_{\nu,\eta} \frac{1}{n}
    | \gnueta(\gausswone_i,\gausswtwo_i,\xi_i) - \EE [\gnueta] | - 1 ]) \leq 1 \}
\end{align}
Since $\abs{\nu},\eta$ are bounded by constants, we have that 
\begin{align}
    \EE[\gnueta] = \EE [ ( 1 - \xirv \nu |\Gausswone| - \Gausswtwo \eta )_+^2 ] \leq c(1 + \boundnu^2 + \boundeta^2) \leq c_2
    \label{eq:proof_unicon_noisy_expect_upper_bound}
\end{align}
for some positive universal constants $c_2$ that may depend on $\boundnu,\boundeta$. Furthermore, we have:
\begin{align}
    ( 1 - \xi_i \nu |\gausswone_i| - \gausswtwo_i \eta )_+^2 \leq c(1+(\boundnu^2+\boundeta^2)z_{\max}^2) \leq c_1 z_{\max}^2
    \label{eq:proof_unicon_noisy_sample_upper_bound}
\end{align}
where $z_{\max} = \max_{1\leq i\leq 2n} \{|\gausswone_i|,|\gausswtwo_i|\}$.
Similarly to inequality~\eqref{eq:proof_unicon_prop_noiseless_exp_triangle}, we apply the triangle inequality and bound the two terms using Equations~\eqref{eq:proof_unicon_noisy_expect_upper_bound} and ~\eqref{eq:proof_unicon_noisy_sample_upper_bound} to obtain:
\ifaistats
\begin{align}
    \EE \bigg[ \exp \Big( \frac{1}{\lambda} \max_i& \sup_{\nu,\eta} \frac{1}{n}
    \Big| ( 1 - \xi_i \nu |\gausswone_i| - \gausswtwo_i \eta )_+^2 - \EE [ ( 1 - \xirv \nu |\Gausswone| - \Gausswtwo \eta )_+^2 ] \Big| \Big) \bigg]
    \leq
    \EE \left[ \exp \left(  \frac{c_1}{n\lambda}z_{\max}^2 \right) \right] \exp \left( \frac{c_2}{n\lambda} \right)  
\end{align}
\else
\begin{align}
    \EE \bigg[ \exp \Big( \frac{1}{\lambda} \max_i \sup_{\nu,\eta} \frac{1}{n}
    \Big| ( 1 - \xi_i \nu |\gausswone_i| - \gausswtwo_i \eta )_+^2 -& \EE [ ( 1 - \xirv \nu |\Gausswone| - \Gausswtwo \eta )_+^2 ] \Big| \Big) \bigg]\\
    &\leq
    \EE \left[ \exp \left(  \frac{c_1}{n\lambda}z_{\max}^2 \right) \right] \exp \left( \frac{c_2}{n\lambda} \right)  
\end{align}
\fi
Thus we obtain that $\psi_{\GB} \leq \inf \{ \lambda>0:\ \EE [ \exp ( \frac{c_1}{n\lambda} z_{\max}^2 ) \exp(\frac{c_2}{n\lambda}) - 1 ] \leq 1 \}$, which is similar to expression~\eqref{eq:proof_unicon_prop_noiseless_exp_triangle} in the proof of Proposition~\ref{prop:unifConv_noiseless_whole}. Hence following the same argument we conclude that $\psi_{\GB} \leq c_{\lambda} \frac{\log n}{n}$ for some universal constant $c_{\lambda}>0$.

\paragraph{Step 2: Bounding $\rademacher(\GB)$}
The upper bound on the Rademacher complexity is derived as follows. First use the fact that $(\cdot)_+$ is $1$-Lipschitz to obtain:
\ifaistats
\begin{align}
     \rademacher(\GB) = \EE \left[ \sup_{\gnueta\in \GB} \left|\frac{1}{n}\sumin \epsilon_i \gnueta(\gausswone_i,\gausswtwo_i,\xi_i) \right| \right]
     \leq 
     2\EE \left[\sup_{\abs{\nu}\leq \boundnuone,\eta\leq \boundetaone}\left| \frac{1}{n}\sumin \epsilon_i ( 1 - \xi_i \nu |\gausswone_i| - \gausswtwo_i \eta )^2 \right| \right],
    \label{eq:proof_unicon_noisy_rademacher_lipschitz}
\end{align}
\else
\begin{align}
     \rademacher(\GB) &= \EE \left[ \sup_{\gnueta\in \GB} \left|\frac{1}{n}\sumin \epsilon_i \gnueta(\gausswone_i,\gausswtwo_i,\xi_i) \right| \right]\\
     &\leq 
     2\EE \left[\sup_{\abs{\nu}\leq \boundnuone,\eta\leq \boundetaone}\left| \frac{1}{n}\sumin \epsilon_i ( 1 - \xi_i \nu |\gausswone_i| - \gausswtwo_i \eta )^2 \right| \right],
    \label{eq:proof_unicon_noisy_rademacher_lipschitz}
\end{align}
\fi
then expand quadratic form and apply triangle inequality for every term to obtain that \eqref{eq:proof_unicon_noisy_rademacher_lipschitz} is upper bounded by:
\ifaistats
\begin{align}
    &2\EE \left[\left| \frac{1}{n}\sumin \epsilon_i  \right| \right]
    +
    2\EE \left[\sup_{\abs{\nu}\leq \boundnuone,\eta\leq \boundetaone}\left| \frac{1}{n}\sumin \epsilon_i 2\xi_i \nu |\gausswone_i| \gausswtwo_i \eta ) \right| \right]
    +
    2\EE \left[\sup_{\eta\leq \boundetaone}\left| \frac{1}{n}\sumin \epsilon_i ( -2 \gausswtwo_i \eta ) \right| \right]
    \\
    +
    2\EE& \left[\sup_{\eta\leq \boundetaone}\left| \frac{1}{n}\sumin \epsilon_i  (\gausswtwo_i)^2 \eta^2 \right| \right]
    +
    2\EE \left[\sup_{\abs{\nu}\leq \boundnuone}\left| \frac{1}{n}\sumin \epsilon_i ( -2 \xi_i \nu |\gausswone_i|) \right| \right]
    +
    2\EE \left[\sup_{\abs{\nu}\leq \boundnuone}\left| \frac{1}{n}\sumin \epsilon_i  \nu^2 (\gausswone_i)^2 \right| \right]
\end{align}
\else
\begin{align}
    &2\EE \left[\left| \frac{1}{n}\sumin \epsilon_i  \right| \right]
    +
    2\EE \left[\sup_{\abs{\nu}\leq \boundnuone,\eta\leq \boundetaone}\left| \frac{1}{n}\sumin \epsilon_i 2\xi_i \nu |\gausswone_i| \gausswtwo_i \eta ) \right| \right]\\
    +&
    2\EE \left[\sup_{\eta\leq \boundetaone}\left| \frac{1}{n}\sumin \epsilon_i ( -2 \gausswtwo_i \eta ) \right| \right]
    +
    2\EE \left[\sup_{\eta\leq \boundetaone}\left| \frac{1}{n}\sumin \epsilon_i  (\gausswtwo_i)^2 \eta^2 \right| \right]\\
    +&
    2\EE \left[\sup_{\abs{\nu}\leq \boundnuone}\left| \frac{1}{n}\sumin \epsilon_i ( -2 \xi_i \nu |\gausswone_i|) \right| \right]
    +
    2\EE \left[\sup_{\abs{\nu}\leq \boundnuone}\left| \frac{1}{n}\sumin \epsilon_i  \nu^2 (\gausswone_i)^2 \right| \right]
\end{align}
\fi
Finally, since sums above do not depend on $\nu$ and $\eta$ any more, we can use standard concentration results for sub-exponential random variables to obtain that $\rademacher(\GB)\lesssim \frac{1}{\sqrt{n}}$. 

\paragraph{Step 3: Proof of the statement}
Similarly to Equation~\eqref{eq:proof_unicon_noisy_expect_upper_bound}, we can bound the variance straightforwardly as follows:
\begin{align}
\label{eq:bound_var_GB}
    \sigma_{\GB}^2
    \leq 
    \sup_{ \gnueta \in \GB } \EE \left[ \gnueta^2 \right]
    \leq
    c_{\sigma_{\GB}} \left( 1 + \boundnuone^4 + \boundetaone^4 \right)
\end{align}
for some positive universal constant $c_{\sigma_{\GB}}>0$. 

Combining all derived bounds and using that $\EE \norm{P_n-P}_{\GB} \leq 2 \rademacher(\GB)$ we obtain from Theorem~\ref{thm:unifconv_adamczak}:
    \begin{align}
        \prob \left( \norm{ P_n - P }_{\GB} \geq 2(1+t) \mathcal{R}_{\GB} + \epsilon \right) 
        \leq
        \exp \left( - c_2 n \epsilon^2 \right) + 3\exp \left( -c_3 \frac{ n \epsilon }{\log n} \right)
    \end{align}
with $c_2^{-1} = 2(1+\delta)c_{\sigma_{\GB}} \left( 1 + \boundnuone^4 + \boundetaone^4 \right)$ and $c_3^{-1} = C c_{\lambda}$, which concludes the proof. 

\end{proof}

\subsection{Additional lemmas}
\begin{lemma}
\label{lemma:fclas_fclasn_inf_diff}
The function  $(\nu,\eta)\mapsto \EE_{\Gausswone,\Gausswtwo \sim \NNN(0,1)}  ( 1 -  \nu |\Gausswone| - \Gausswtwo \eta )_+^2$ is an infinitely differentiable function. Moreover, under Assumption 1 from Section~\ref{sec:setting}, the function $(\nu,\eta)\mapsto \EE_{\Gausswone,\Gausswtwo \sim \NNN(0,1)} \allowbreak \EE_{\xirv \sim \prob(\cdot\vert \Gausswone)} ( 1 - \xirv \nu |\Gausswone| - \Gausswtwo \eta )_+^2 $ is also an infinitely differentiable function.
\end{lemma}
\begin{proof}
Note that the conditional expectation of the first function is given by:
\ifaistats
\begin{align}
    \EE_{\Gausswtwo|\Gausswone=\gausswone} [ ( 1&-\nu\abs{\gausswone} - \eta\Gausswtwo )_+^2 ] = 
    \int_{-\infty}^{\frac{1}{\eta}(1-\nu\abs{\gausswone})} \frac{1}{\sqrt{2\pi}} \exp \left( -\frac{1}{2}(\gausswtwo)^2 \right) (1-\nu\abs{\gausswone} - \eta\gausswtwo)^2 d\gausswtwo \\
    &=
    \eta(1-\nu\abs{\gausswone}) \exp\left( -\frac{1}{2\eta^2}(1-\nu\abs{\gausswone})^2 \right) + 
    ((1-\nu\abs{\gausswone})^2+\eta^2) \Phi\left(\frac{1}{\eta}(1-\nu\abs{\gausswone}) \right),
\end{align}
\else
\begin{align}
    \EE_{\Gausswtwo|\Gausswone=\gausswone} &[ ( 1-\nu\abs{\gausswone} - \eta\Gausswtwo )_+^2 ] \\
    &=
    \int_{-\infty}^{\frac{1}{\eta}(1-\nu\abs{\gausswone})} \frac{1}{\sqrt{2\pi}} \exp \left( -\frac{1}{2}(\gausswtwo)^2 \right) (1-\nu\abs{\gausswone} - \eta\gausswtwo)^2 d\gausswtwo \\
    &=
    \eta(1-\nu\abs{\gausswone}) \exp\left( -\frac{1}{2\eta^2}(1-\nu\abs{\gausswone})^2 \right) + 
    ((1-\nu\abs{\gausswone})^2+\eta^2) \Phi\left(\frac{1}{\eta}(1-\nu\abs{\gausswone}) \right),
\end{align}
\fi
which is an infinitely differentiable function in $\nu$ and $\eta$. Since the function given in the lemma is an expectation of an infinitely differentiable function, it is as well infinitely differentiable, which finishes the first part of the proof.

Now, note that using Assumption 1 we can rewrite the second function as: 
\ifaistats
\begin{align}
    \EE_{\Gausswone}\left[ \PP(\xirv=1|\Gausswone) \EE_{\Gausswtwo|\Gausswone}[(1-\nu\abs{\Gausswone} - \eta\Gausswtwo)_+^2] + \PP(\xirv=-1|\Gausswone) \EE_{\Gausswtwo|\Gausswone}[(1+\nu\abs{\Gausswone} - \eta\Gausswtwo)_+^2] \right].
\end{align}
\else
\begin{align}
    \EE_{\Gausswone}\bigg[ \PP(\xirv=1|\Gausswone) \EE_{\Gausswtwo|\Gausswone}[(1-\nu\abs{\Gausswone} &- \eta\Gausswtwo)_+^2] \\
    &+ \PP(\xirv=-1|\Gausswone) \EE_{\Gausswtwo|\Gausswone}[(1+\nu\abs{\Gausswone} - \eta\Gausswtwo)_+^2] \bigg].
\end{align}
\fi
But, similarly to above, we can show that $\EE_{\Gausswtwo|\Gausswone}[(1+\nu\abs{\Gausswone} - \eta\Gausswtwo)_+^2]$ is infinitely differentiable, implying that the whole function is also infinitely differentiable.
\end{proof}

\end{document}